\documentclass{article} 
\usepackage{iclr2026_conference,times}

\usepackage{amsmath,amsfonts,bm}

\def\eqref#1{equation~\ref{#1}}

\def\1{\bm{1}}

\DeclareMathAlphabet{\mathsfit}{\encodingdefault}{\sfdefault}{m}{sl}
\SetMathAlphabet{\mathsfit}{bold}{\encodingdefault}{\sfdefault}{bx}{n}

\usepackage{hyperref}
\usepackage{url}
\usepackage{subfigure}
\usepackage{colortbl}
\usepackage{multirow}
\usepackage{makecell}
\usepackage{graphicx}
\usepackage{booktabs} 
\usepackage{tabularx}
\usepackage{xcolor}
\usepackage{longtable}
\usepackage{adjustbox}

\title{MindVL: Towards Efficient and Effective Training of Multimodal Large Language Models on Ascend NPUs}

\iclrfinalcopy

\author{Feilong Chen$^\ast$$^\dag$, Yijiang Liu$^\ast$, Yi Huang$^\ast$,\\ \textbf{Hao Wang, Miren Tian, Ya-Qi Yu, Minghui Liao, Jihao Wu} \\[1.2mm]
Huawei Technologies Co., Ltd. \\[1.2mm]
\{feilong.chen, liuyijiang7, huangyi98\}@huawei.com\\[1.2mm]
{\fontsize{9.4pt}{9.84pt}\selectfont  \textsuperscript{$\ast$}Equal contribution ~~~\textsuperscript{$\dag$}Project Lead}
\vspace{-20pt}
}

\begin{document}

\maketitle

\begin{abstract}

\vspace{-2mm}
\begin{figure}[ht]
	\centering
    \includegraphics[width=0.99\linewidth]{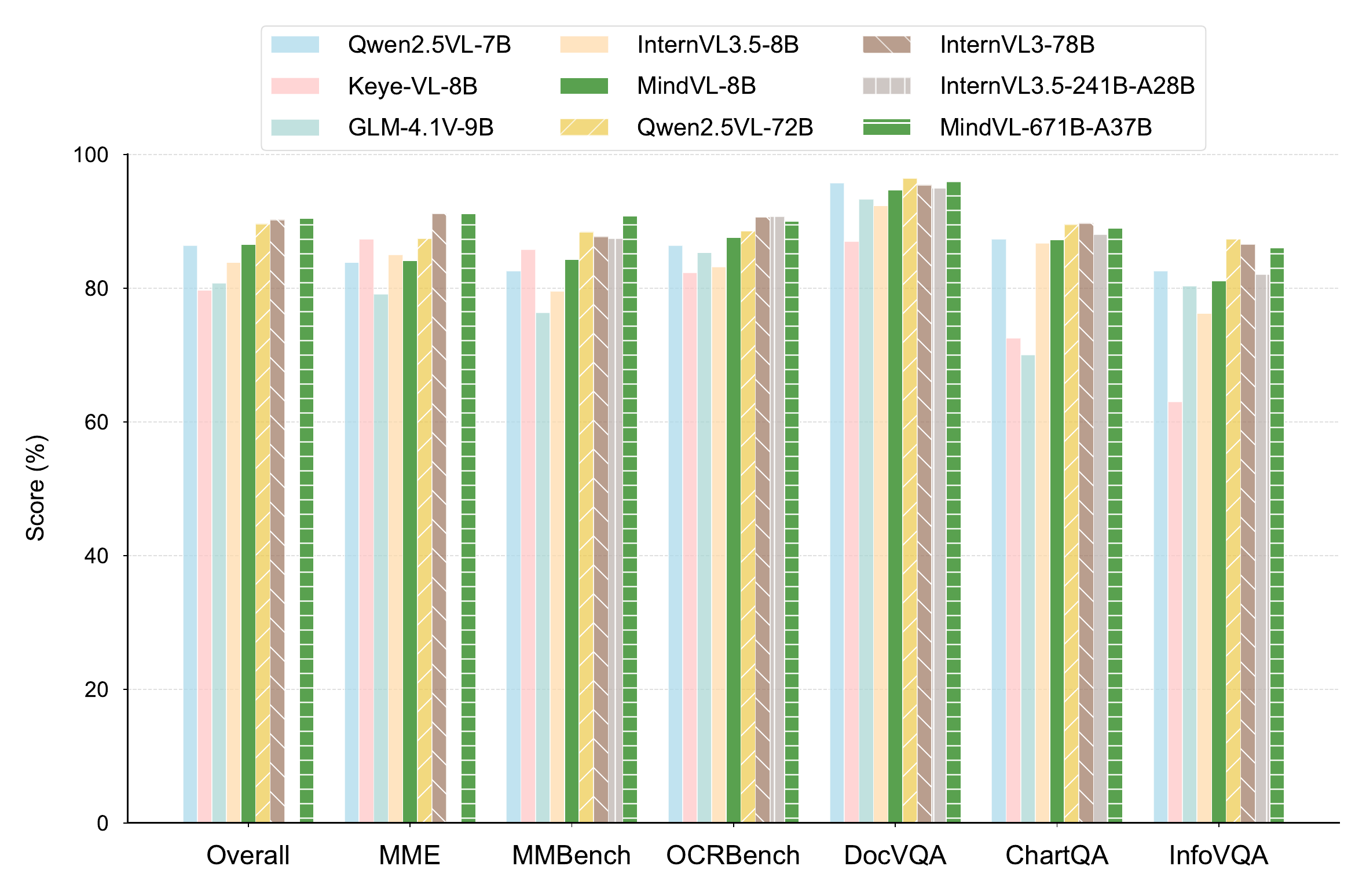}
    \vspace{-4mm}
	\caption{Benchmark performance of MindVL and its counterparts.}
	\label{fig:model_benchmark}
\end{figure}
\vspace{-4mm}
\end{abstract}

We propose \textbf{MindVL}, a multimodal large language model (MLLMs) trained on Ascend NPUs. The training of state-of-the-art MLLMs is often confined to a limited set of hardware platforms and relies heavily on massive, undisclosed data recipes, which hinders reproducibility and open research. To change the common perception that Ascend hardware is unsuitable for efficient full-stage MLLM training, we introduce \textbf{MindSpeed-MLLM}, a highly efficient training framework that supports stable and high-performance training of large-scale Dense and Mixture-of-Experts (MoE) models on Ascend hardware. Based on this, we provide a systematic and open description of the data production methods and mixing strategies for all training stages. Furthermore, we present MindVL, a data-efficient multimodal large language model trained end-to-end on Ascend NPUs. In addition, we find that averaging weights from checkpoints trained with different sequence lengths is particularly effective and yields further gains when combined with test-time resolution search. Our experiments demonstrate superior data efficiency: \textbf{MindVL-8B} matches the performance of Qwen2.5VL-7B using only 10\% of its training data, while our MoE model, \textbf{MindVL-671B-A37B}, matches Qwen2.5VL-72B using only 3\% of the Qwen2.5VL training data, and achieves comparable performance with other leading multimodal MoE models. Our work provides the community with a valuable hardware alternative, open data recipes, and effective performance-enhancing techniques.

\section{Introduction}
\vspace{-8pt}

Multimodal Large Language Models (MLLMs)~\citep{chen2023vlp,zhang2024mm} represent a significant advancement in artificial intelligence, demonstrating remarkable capabilities in understanding and generating content across vision and language modalities. Despite rapid progress, the field faces two major challenges that limit open research. First, the training of top-tier models~\citep{chen2023x,guo2025seed1,chen2024expanding,Qwen2.5-VL,vteam2025glm45vglm41vthinkingversatilemultimodal} is predominantly dependent on a specific hardware ecosystem (e.g., NVIDIA GPUs), creating a perception that alternative platforms like Ascend are incapable of efficient full-stage MLLM training. This perception restricts hardware choices for researchers. Second, while the composition of training data is widely acknowledged as a critical factor for performance, most leading models~\citep{Qwen2.5-VL,guo2025seed1} only offer high-level descriptions of their data; the exact recipes, cleaning pipelines, and mixing strategies are often treated as proprietary secrets. This lack of transparency severely impedes reproducibility and hinders community progress.


To address these challenges, this work advocates for a more open and efficient research paradigm for MLLMs. Our primary contribution is \textbf{MindSpeed-MLLM}, an optimized training framework that demonstrates the full capability of Ascend hardware for stable and efficient training of both dense and large-scale Mixture-of-Experts (MoE) models from pre-training to supervised fine-tuning (SFT). This provides researchers with a crucial and performant hardware alternative.


Second, we aim to demystify the "black box" of leading MLLM data. We provide a comprehensive and open description of our data production methodology, including detailed data collection, cleaning, processing pipelines, and—most importantly—the mixing ratios used for each training stage. We believe this detailed data recipe offers a valuable blueprint for the community.


Third, we present \textbf{MindVL}, a data-efficient multimodal large language model trained end-to-end on Ascend NPUs. MindVL undergoes a three-phase training pipeline: warm-up, multitask training, and supervised instruction tuning, to incrementally enhance its multimodal capabilities. Starting with basic visual and cross-modal pre-training, the pipeline progresses to large-scale instruction adjustment, aligning the model with real-world use cases. Additionally, we integrate multimodal data packaging and hybrid parallelism techniques to significantly boost end-to-end training speed. To further optimize performance, we introduce two key strategies: test-time resolution search (to dynamically select optimal image resolutions for inference) and model weight averaging (to stabilize and improve final performance).


Extensive experiments validate the effectiveness of our approach. Our models achieve performance comparable to state-of-the-art models (e.g., the Qwen2.5VL series~\citep{Qwen2.5-VL}) while utilizing orders of magnitude less training data. This result underscores the high quality of our data recipe and the robust training capability of the MindSpeed-MLLM framework on Ascend hardware. The contributions of this paper are as follows:
\vspace{-4pt}
\begin{itemize}
    \item We introduce \textbf{MindSpeed-MLLM}, a framework that enables efficient full-stage training of both Dense and MoE MLLMs on Ascend hardware, challenging existing perceptions.
    \item We provide a detailed and open data recipe for all training stages, promoting transparency and reproducibility in MLLM research.
    \item We introduce two enhancement techniques: multimodal model weight averaging and test-time resolution search, both of which contribute to improved pure text and multimodal performance.
    \item Experimental results show that MindVL achieves performance comparable to state-of-the-art models. Specifically, \textbf{MindVL-8B} matches the performance of Qwen2.5VL-7B using only 10\% of its training data, while \textbf{MindVL-671B-A37B}, matches Qwen2.5VL-72B using only  3\% of the Qwen2.5VL training data and and achieves comparable performance with other leading multimodal MoE models.. 
\end{itemize}


\vspace{-12pt}
\section{Related Work}

\subsection{Training of Multimodal Large Language Models}
The training of Multimodal Large Language Models (MLLMs) faces significant challenges due to model heterogeneity (e.g., integrating vision encoders with LLMs) and data heterogeneity (e.g., processing images, videos, and text). These challenges necessitate specialized training frameworks and hardware optimizations to achieve efficiency and scalability. MLLM training relied on adapting text-centric frameworks like Megatron-LM~\citep{megatron-lm}  and DeepSpeed~\citep{rasley2020deepspeed} to handle multimodal data by treating visual modules as additional layers within a unified parallelism strategy (e.g., combining Tensor, Pipeline, and Data Parallelism). Although numerous studies have successfully trained state-of-the-art MLLMs using these frameworks, their development has primarily focused on optimization for NVIDIA GPUs. In contrast, exploration of multimodal large-scale model training on Ascend NPUs remains limited, and a comprehensive, full-stage methodology for such environments has yet to be established.

\vspace{-8pt}
\subsection{Data Curation of Multimodal Large Language Models}
Data curation is a cornerstone of multimodal large language model (MLLM) performance, as high-quality, well-structured multimodal data directly enables effective vision-language alignment and task adaptability. Existing literature and MLLM technical reports generally acknowledge the significance of data curation, outlining broad frameworks that typically categorize data by task and emphasizing core curation goals. Leading MLLMs, including Qwen2.5-VL \citep {Qwen2.5-VL} and Seed-VL 1.5 \citep {guo2025seed1}, provide only generalized descriptions of their data curation pipelines, lacking critical granular details that are essential for reproducibility and comparative analysis. This gap hinders the research community from fully dissecting how data curation choices impact MLLM capabilities and limits the development of future MLLMs.

\vspace{-8pt}
\section{MindSpeed-MLLM: Training Infrastructure on Ascend NPUs}\label{sec:mindspeed}

\vspace{-8pt}
Due to substantial hardware and software discrepancies, training frameworks widely used on NVIDIA GPUs—such as Megatron-LM \citep{megatron-lm}—and common acceleration libraries (e.g., FlashAttention \citep{DBLP:conf/nips/DaoFERR22,DBLP:conf/iclr/Dao24}, Transformer-Engine \citep{transformer-engine}) cannot be directly deployed on Ascend devices. \footnote{A detailed analysis of NVIDIA-Ascend hardware and software differences is provided in Appendix \ref{app:mindspeed-mllm}.} 
Thus, developing a robust distributed training framework for the Ascend ecosystem is essential. 
To this end, we introduce MindSpeed-MLLM: a distributed multimodal training library tailored for Ascend NPUs.

\vspace{-8pt}
\subsection{MindSpeed-MLLM}

\subsubsection{MindSpeed Series Libraries}

MindSpeed is a high-performance acceleration library tailored for the Ascend platform, encompassing three core components to support large-model training: MindSpeed-Core \citep{mindspeed-core}, MindSpeed-LLM \citep{mindspeed-llm} MindSpeed-MM \citep{mindspeed-mm} and MindSpeed-RL \citep{feng2025mindspeedrldistributeddataflow}.

MindSpeed-Core, built on and optimized for Ascend hardware based on Megatron-LM, delivers multi-dimensional optimizations in computing, memory, communication, and parallelism, enabling accelerated training for scenarios like long sequences and MoE. 
MindSpeed-LLM provides a rich set of LLM with extensive training optimization features, while MindSpeed-MM realizes mainstream Vision-Language Models. 

Despite their individual strengths, the existing components present integration and functionality gaps for end-to-end multi-modal large language model (MLLM) training. 
Specifically, MindSpeed-MM lacks robust support for critical multi-modal data processing functionalities, including distributed data loading, data packing, and training resumption. 
Furthermore, its optimizations for the language backbone are not as comprehensive or mature as those provided by MindSpeed-LLM.
To address these gaps and leverage the strengths of existing components, MindSpeed-MLLM is developed with targeted optimizations.

\vspace{-8pt}
\subsubsection{MindSpeed-MLLM Framework}

\vspace{-2mm}
\begin{figure}[ht]
	\centering
    \includegraphics[width=0.7\linewidth]{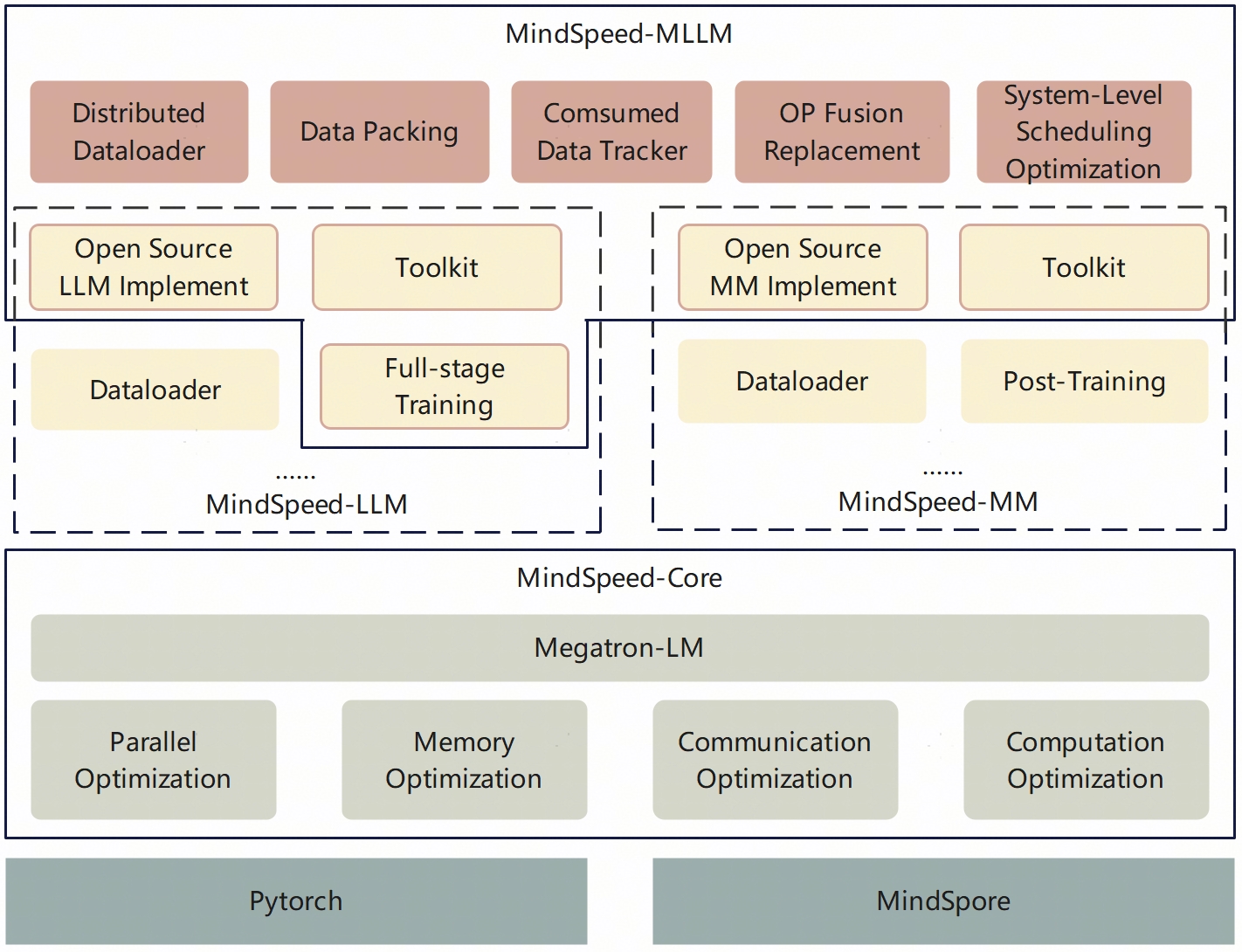}
    \vspace{-4mm}
	\caption{The Overall Architecture of MindSpeed-MLLM and Its Relationship with Other MindSpeed Frameworks.}
	\label{fig:mindspeed_mllm}
\vspace{-12pt}
\end{figure}


As depicted in Figure \ref{fig:mindspeed_mllm}, the MindSpeed-MLLM framework is constructed with a hierarchical architecture. It builds upon the foundational optimizations from MindSpeed-Core and integrates partial modules from MindSpeed-LLM and MindSpeed-MM.
Beyond this integration, the core efforts of MindSpeed-MLLM lie in enhancing multi-modal data processing, operator fusion replacement, and system-level scheduling optimization, which will be introduced separately in the following sections.

\vspace{-8pt}
\subsubsection{Multi-Modal Data Loader}
We have developed a multi-modal data loader with the following features:

\textbf{Distributed Multi-Modal Data Loader: }
It supports distributed data loading, where each data parallel group only reads the data within that group, effectively avoiding bottlenecks caused by reading the same data during large-scale training redundantly.

\textbf{Online Packing:}
It enables online packing of multi-modal data, which combines data of different lengths into a specified length and fills in valid data content as much as possible \cite{DBLP:conf/iclr/MaWL25,DBLP:conf/acl/Wang0W0HG25}.
Thus, each Pack dataset has almost the same length, reducing the number of samples during training and improving training efficiency. 
Meanwhile, it controls the number of visual tokens to avoid load imbalance between pipeline parallel stages caused by uneven quantities of different modal data.

\textbf{Consumed Data Tracker:}
It supports tracking of consumed data, which facilitates the location of data breakpoints during checkpoint-based recovery training, eliminates redundant data retraining, and ensures accurate resumption of training tasks.

\vspace{-8pt}
\subsubsection{Operator Fusion Replacement}

\vspace{-6pt}
Choosing hardware-friendly operators effectively boosts training efficiency in model development. 
MindSpeed already supports fusion operator replacements, such as RMSNorm, SoftMax, MoE Token Permute, Unpermute, and Adamw, etc. 
Beyond these pre-provided fusion operators, MindSpeed-MLLM further optimizes as follows:

\textbf{Common Attention Patch:}
MindSpeed-LLM and MindSpeed-MM replaced attention operators for language and visual components respectively, swapping the flash attention interface with npu fusion attention. 
Since both components require this operator, we patched attention in the shared transformer block to a unified fusion operator call. 
Different mask types are passed to specific components depending on the module ids.

\textbf{Mask Compression:}
When training native resolutions ViTs, we first concatenate the hidden states of distinct images along the sequence dimension, followed by invoking the attention operator through a variable-length way.
This design simultaneously reduces memory overhead and enhances computational efficiency.
Combined with the sparse mode parameter of the npu fusion attention operator, passing the compressed attention mask further reduces the memory usage.

\textbf{Operation Replacement:}
Profiling showed low computational efficiency of Conv operators in CANN 8.0. 
Thus, we replaced Conv2d and Conv3d operations with Matmul operation equivalently. 
The supporting checkpoint convert toolkit also added corresponding support for this replacement.

\subsubsection{System-Level Scheduling Optimization}



Within the Ascend ecosystem, system-level scheduling optimizations deliver performance enhancements through multiple mechanisms. 
Fine-grained core binding minimizes cross-NUMA node memory accesses, reducing both task scheduling overhead and inter-core switching costs. 
Concurrently, operator deployment queue optimizations partition deployment tasks across multi-stage pipelines operating in parallel. This approach enables partial overlap between execution and submission processes, reducing overall latency and improving end-to-end performance.

\vspace{-8pt}
\section{Data Curation}\label{sec:data}
The MindVL training corpus contains 447 billion diverse and high-quality tokens used for three training stages. The data is categorized according to target capabilities, and the curation process for each category is detailed in the following subsections. Due to space limitations, more detailed processing steps and data ratios are provided in the Appendix~\ref{appendix:data}.

As shown in Figure~\ref{fig:data}, the training data of MindVL is open-sourced into two main categories: image-text pairs and visual instructions. The image-text pair data is further divided into eight subcategories. Figure~\ref{fig:data} illustrates the core processing methods for each category, along with fundamental filtering techniques and some models used during data annotation.

\begin{figure}[ht]
	\centering
    \includegraphics[width=0.99\linewidth]{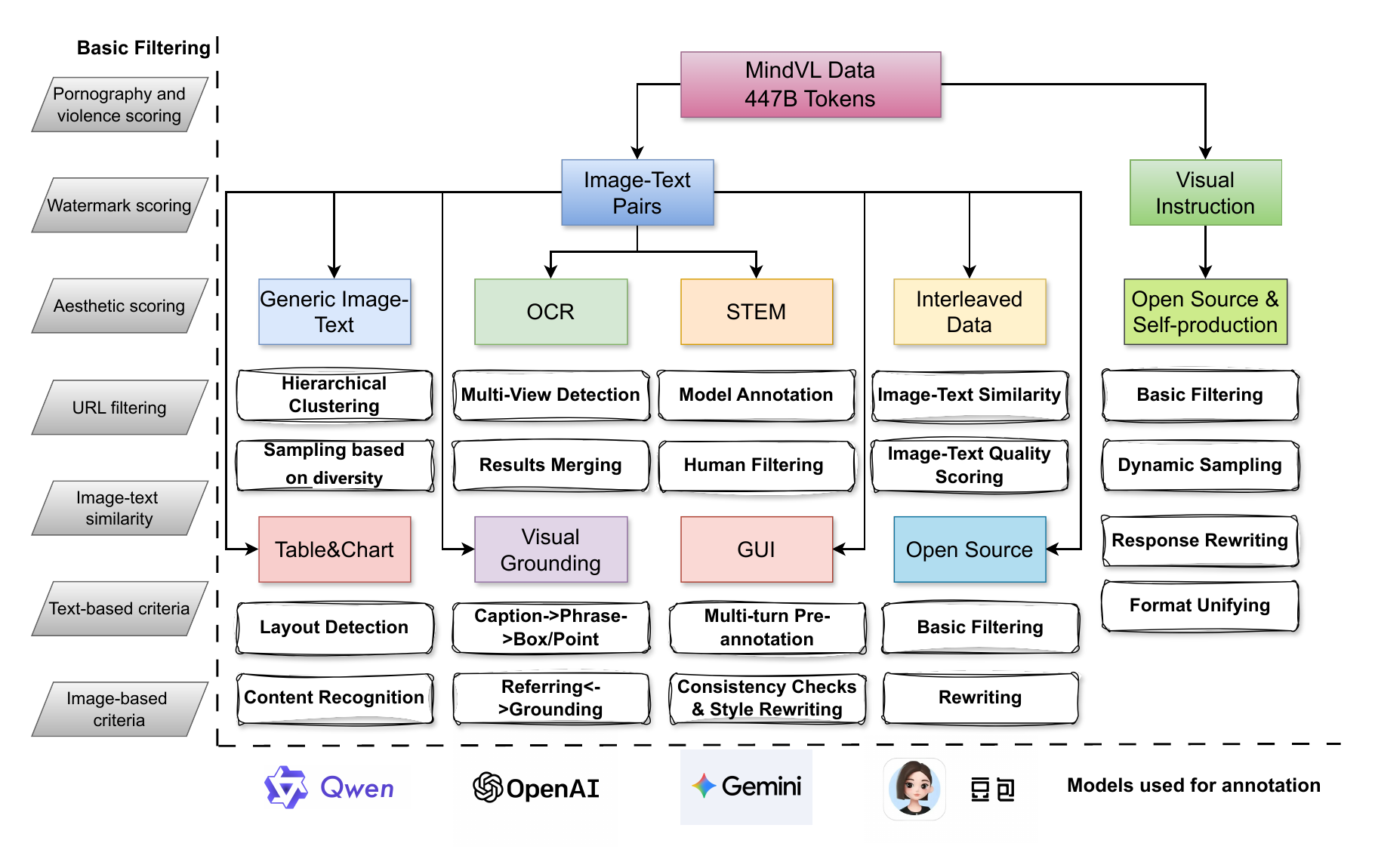}
    \vspace{-8mm}
	\caption{Data curation process of MindVL training data.}
	\label{fig:data}
\end{figure}

\vspace{-15pt}
\subsection{Warm-up Data}\label{sec:warmup}

The MindVL warm-up corpus consists of 256 billion diverse, high-quality tokens. Warm-up data span six categories, including image caption, OCR, Visual Grounding, Table\&Chart, GUI, STEM, with key processing steps described in Appendix~\ref{sec:warm-up}.

\vspace{-8pt}
\subsection{Multitask Training Data}\label{sec:multitask_training_data}

 Multitask training data consists of interleaved image-text (Table \ref{tab:interleaved}), visual instruction (Table \ref{tab:multitask}, about 80B tokens), web2code~\citep{yun2024web2code} and text instruction (Table \ref{tab:language-statistics}), totally 179B tokens. We add textual data to maintain the MindVL's linguistic capabilities. Details are described in Appendix~\ref{sec:appendix_multitask-training-data}



\vspace{-8pt}
\subsection{Supervised Fine-tuning Data}
High-quality instruction data is sampled from open-source multitask datasets (classified by model), with low-quality answers re-annotated (via model and human verification), totally 12B tokens. Textual data is incorporated at a multimodal-to-language ratio of 1:1 to preserve the MLLM’s linguistic performance. Details are described in Appendix~\ref{sec:sft_data}

\vspace{-8pt}
\section{MindVL}\label{sec:mindvl}

\vspace{-4pt}
\subsection{Architecture}
The architecture of MindVL bears resemblance to that of Qwen2.5-VL, comprising three core components: a vision encoder, an MLP projector, and a large language model. The vision encoder natively supports dynamic image resolutions and adopts 2D RoPE~\citep{wang2024qwen2} for positional encoding, enabling flexible adaptation to images of arbitrary dimensions. 

For model initialization, we utilize Qwen2.5ViT~\footnote{The weights of Qwen2.5ViT are derived from the Qwen2.5-VL 72B model.} as visual encoder and Qwen3 LLM~\citep{yang2025qwen3} / DeepSeek-V3-0324~\citep{liu2024deepseek} as language backbones, choices that ensure robust baseline performance. In contrast, the MLP projector is initialized randomly.

\vspace{-8pt}

\subsection{Training Recipe}\label{sec:train_recipe}
As shown in Table~\ref{tab:training_recipe}, the training process of MindVL is divided into three distinct phases, each employing different data
configurations and training strategies to progressively enhance the model’s capabilities.
\vspace{-15pt}

\begin{table}[htbp]
\scriptsize
  \centering
  \caption{Training setup and hyper-parameters in three training stages.}
  \renewcommand{\arraystretch}{1.2} 
    \begin{tabular}{l|ccc}
    \toprule
    \textbf{Stages} & \makecell{\textbf{Warm-up}\\8B / 671B} & \makecell{\textbf{Multitask training}\\8B / 671B} & \makecell{\textbf{SFT}\\8B / 671B} \\
    \midrule
    Training budget (tokens)  & 256B/16B & 179B/80B & 12B \\
    Sequence length & 8192 & 8192 & 2048/4096/8192 \\
    Trainable components & MLP adaptor & All & All \\
    Batch sizes  & 1024 & 1024 & 512 \\
    LR warm-up ratio & 0.1 & 0.1 & 0.1 \\
    Maximum LR & 1e-3/2e-4 & 2e-5/5e-6 & 1e-5/2e-6 \\
    Minimum LR  & 0/1e-5 & 1e-5/5e-7  & 0/2e-7 \\
    \bottomrule
    \end{tabular}%
  \label{tab:training_recipe}%
\end{table}%

During the warm-up phase, only the MLP adapter is trained to align the vision transformer and the language model. This process relies on carefully curated data spanning image captions, visual grounding, OCR, and STEM content to establish foundational multimodal alignment.

In the multitask training phase, all parameters are optimized with diverse multimodal data such as interleaved image-text, VQA, math reasoning, agent tasks, video, and pure text. This enhances complex visual-language reasoning while preserving linguistic abilities.

Finally, supervised fine-tuning unlocks all parameters and focuses on instruction-aware optimization to adapt pretrained representations to downstream tasks.






\vspace{-8pt}
\subsection{Model Merging  with Different Training Sequence Lengths}
Merging weights of deep models~\citep{li2023deepmodelfusion} has been verified to be an efficient way in various applications, including multi-task learning, federated learning, model compression, and continual learning.
We also explore the weight averaging strategies with the models trained under different settings of input sequence length and image resolution.
Specifically, for the SFT phase, we train the model with sequence length of 2K, 4K and 8K with max\_pixels of 1280*28*28, 3072*28*28 and 4096*28*28 respectively to enhance the model’s adaptability to different context inputs and image resolutions.
Finally, the MindVL model is created by averaging weights of models with different training sequence length.

\vspace{-8pt}
\subsection{Test-Time Resolution Search}\label{sec:test_time_resolution_search}

The evaluation model is merged by model weights trained with different sequence lengths as well as different max\_pixels threshold of training images.
Such search strategy is generally effective as the original resolution of the test images maybe be out of distribution relative to that of the training images.
Therefore, we conduct a grid search about up-scaling small images to surpass a specified min\_pixels threshold and down-scaling high-resolution image to be lower than a specified max\_pixels threshold.
Specifically, we set the grid search space of the min\_pixels as \{4, 16, 32, 64\}*28*28 and max\_pixels as \{1280, 2048, 2560,3072,4096,8192\}*28 *28.
The analysis results are presented in Section~\ref{sec:Effectiveness of Model Merging}.


\vspace{-8pt}
\section{Experiments}

\vspace{-8pt}

\subsection{Performance of MindVL-8B}
As shown in Table~\ref{tab:main_results}, we evaluate the overall performance of MindVL-8B on MMBench~\citep{liu2024mmbench}, MME~\citep{chaoyou2023mme}, OCRBench~\citep{liu2024ocrbench}, DocVQA~\citep{mathew2021docvqa}, ChartQA~\citep{masry2022chartqa}, InfoVQA~\citep{mathew2022infographicvqa}. Overall, MindVL-8B outperforms several leading models, including Qwen2.5-VL-7B, GLM-4.1V-9B, Keya-VL-8B, and InternVL3.5-8B. Notably, MindVL-8B achieves this superior performance using only 447B tokens of training data—roughly one-tenth of the data used by Qwen2.5-VL-7B. Furthermore, compared to models trained with trillion-scale tokens such as GLM-4.1V-9B and Keye-VL-8B, MindVL delivers significantly stronger results, outperforming them by margins of 6.6 and 5.8 points, respectively. When compared to InternVL3.5-8B, which has a similar pre-training scale, MindVL-8B maintains a lead of 2.7 points. 
These outcomes highlight the effectiveness of our training methodology and data efficiency, demonstrating the capability to develop high-performing multimodal large language models on Ascend NPUs.

\begin{table}[htbp]

\vspace{-12pt}
  \centering
  \caption{Performance of MindVL-8B and comparison models on multimodal benchmarks. "+" indicates that there is a portion of data with unlabeled quantities. The results of the comparative models are referenced from~\citep{wang2025internvl3_5}.}
  \renewcommand{\arraystretch}{1.2} 
  \resizebox{0.99\linewidth}{!}{
    \begin{tabular}{l|c|ccccccc}
    \toprule
    \textbf{Model} & \#Tokens & \textbf{MME} & \textbf{MMBench} & \textbf{OCRBench} & \textbf{DocVQA} & \textbf{ChartQA} & \textbf{InfoVQA} & \textbf{Overall} \\
    \midrule
    Qwen2.5-VL-7B~\citep{Qwen2.5-VL}  & 4.1T+ & 83.8 & 82.6 & \underline{86.4} & \textbf{95.7} & \textbf{87.3} & 82.6 & \underline{86.4}\\
    Keye-VL-8B~\citep{kwaikeyeteam2025kwaikeyevltechnicalreport}& 1T+ & \textbf{87.3} & \textbf{85.8} & 82.3 & 87.0 & 72.5 & 63.0 & 79.9\\
    GLM-4.1V-9B~\citep{vteam2025glm45vglm41vthinkingversatilemultimodal}  & 2T+ & 79.1 & 76.3 & 85.3 & 93.3 & 70.0 & 80.3 & 80.7\\
    InternVL3.5-8B~\citep{wang2025internvl3_5} & 380B+ & \underline{85.0} & 79.5 & 83.2 & 92.3 & 86.7 & 76.2 & 83.8\\
    \midrule
    MindVL-8B & 447B & 84.1 & \underline{84.3} & \textbf{87.6} & \underline{94.7} & \underline{87.2} & \underline{81.1} & \textbf{86.5}\\
    \bottomrule
    \end{tabular}
    }
  \label{tab:main_results}%
\end{table}%

\vspace{-8pt}
\subsection{Performance of MindVL-671B-A37B}
\begin{table}[t]

\vspace{-8pt}
\centering
\scriptsize
\setlength{\tabcolsep}{3pt}
\caption{Comparison of OCR, chart, document and general understanding performance.
}
\resizebox{\linewidth}{!}{
\begin{tabular}{l|c|cccccc|c}
\toprule
\textbf{Model} & \textbf{\#Tokens} & \makecell{\textbf{MME}\\} & \makecell{\textbf{MMBench}\\} & \makecell{\textbf{OCR}\\\textbf{Bench}} & \makecell{\textbf{DocVQA}} & \makecell{\textbf{ChartQA}} & \makecell{\textbf{InfoVQA}} & \textbf{Overall} \\
\midrule
GPT-4V~\citep{hurst2023gpt} & -- & 68.8 & 80.0 & 64.5 & 88.4 & 78.5 & 75.1 & 70.0 \\
GPT-4o-20240513~\citep{hurst2024gpt} & -- & -- & 83.1 & 73.6 & 92.8 & 85.7 & 79.2 & --\\
Claude-3-Opus~\citep{Claude3} & -- & 56.7 & 60.1 & 69.4 & 89.3 & 80.8 & 55.6 & 67.3 \\
Claude-3.5-Sonnet~\citep{Claude3_5} & -- & -- & 80.9 & 78.8 & 95.2 & 90.8 & 74.3 & -- \\
Gemini-1.5-Pro~\citep{team2024gemini} & -- & -- & 74.6 & 75.4 & 93.1 & 87.2 & 81.0 & -- \\
Step3V~\citep{ste3v} & -- & -- & 81.1 & 83.7 & -- & -- & -- & -- \\
GLM-4.5V~\citep{vteam2025glm45vglm41vthinkingversatilemultimodal} & 2T+ & -- & 86.7 & 87.2 & 94.5 & 86.6 & 84.1 & 75.8 \\
Qwen2-VL-72B~\citep{wang2024qwen2} & 1.4T+ & 88.7 & 85.9 & 87.7 & \textbf{96.5} & 88.3 & 84.5 & 88.6 \\
Qwen2.5-VL-72B~\citep{Qwen2.5-VL} & 4.1T+ & 87.4 & \underline{88.4} & 88.5 & \underline{96.4} & \underline{89.5} & \underline{87.3} & 89.6 \\
InternVL3-78B~\citep{zhu2025internvl3} & 200B+ & \textbf{91.1} & 87.7 & \underline{90.6} & 95.4 & \textbf{89.7} & 86.5 & \underline{90.2} \\
InternVL3.5-241B-A28B~\citep{wang2025internvl3_5} & 380B+ & -- & 87.4 & \textbf{90.7} & 94.9 & 88.0 &  82.0  & -- \\
\midrule
MindVL-671B-A37B  & 106B & \textbf{91.1} & \textbf{90.8} & 90.0 & 96.0 & 89.0 & \textbf{88.9} & \textbf{91.0} \\
\bottomrule
\end{tabular}
}
\label{tab:exp-ocr}
\vspace{-3mm}
\end{table}

\begin{table}[t]

\vspace{-8pt}
\centering
\scriptsize
\setlength{\tabcolsep}{3pt}
\caption{Comparison of multi-modal Hallusion/STEM performance.
}
\resizebox{\linewidth}{!}{
\begin{tabular}{l|c|cccccc|c}
\toprule
\textbf{Model} & \textbf{\#Tokens} & \makecell{\textbf{Hallusion}\\\textbf{Bench}} & \makecell{\textbf{AI2D}} & \makecell{\textbf{MMVet}\\} & \makecell{\textbf{MathVista}\\} & \makecell{\textbf{MathVision}\\} & \makecell{\textbf{MMMU pro}\\} & \textbf{Overall} \\
\midrule
GLM-4.5V~\citep{vteam2025glm45vglm41vthinkingversatilemultimodal} & 2T+ & 65.4 & 86.6 & 75.2 & 78.2 & 52.5 & \textbf{59.8} & \textbf{69.6} \\
Qwen2-VL-72B~\citep{wang2024qwen2} & 1.4T+ & 58.1 & 88.1 & 74.0 & 70.5 & 25.9 & 46.2 & 60.5 \\
Qwen2.5-VL-72B~\citep{Qwen2.5-VL} & 4.1T+ & 55.2 & 88.7 & 76.2 & 74.2 &  38.1 & 51.1 & 63.9 \\
InternVL3-78B~\citep{zhu2025internvl3} & 200B+ & 59.1 & \textbf{89.7} & \textbf{81.3} & 79.0 & 43.1 & -- & -- \\
InternVL3.5-241B-A28B~\citep{wang2025internvl3_5} & 380B+ & 57.3 & 87.3 & 81.2 & \textbf{82.7}  & \textbf{63.9} & --  & -- \\
\midrule
MindVL-671B-A37B & 106B & \textbf{68.6} & 85.2 & 67.9 & 72.9 & 34.1 & 49.5 & 63.0 \\
\bottomrule
\end{tabular}
}
\label{tab:exp-multimodal}
\vspace{-8pt}
\end{table}

\begin{table}[t]
\centering
\scriptsize

\vspace{-8pt}
\setlength{\tabcolsep}{3pt}
\caption{\textbf{Comparison of model performance across diverse language benchmarks.} All results are uniformly evaluated by the internal testing platform. 
}
\resizebox{\linewidth}{!}{
\begin{tabular}{l|ccccccc|c}
\toprule
\textbf{Model} & \textbf{AIME2024} & \textbf{AIME2025} & \textbf{GPQA-D} & \textbf{IFEval} & \textbf{ArenaHard} & \textbf{C-SimpleQA} & \textbf{C-Eval} & \textbf{Overall} \\
\midrule
Qwen2.5VL-32B~\citep{Qwen2.5-VL} &  30.0 & 16.7  &  51.5 & 64.3  &  92.2 & 44.9 & 81.3  & 54.1  \\
Qwen2.5VL-72B~\citep{Qwen2.5-VL}  &  26.7 &  16.7 & 50.0  &  85.6 &  71.2 & 49.8 &  86.1 & 55.2  \\
\midrule
DeepSeek-V3-0324~\citep{liu2024deepseek} & 60.0 & 43.3 & \textbf{69.2} & 81.9 & 94.8 & \textbf{73.3 }& \textbf{89.6} & 73.2 \\
MindVL-671B-A37B & \textbf{63.3} & \textbf{46.7} & 68.7 & \textbf{84.7 }&\textbf{ 97.0} & 72.9 & 88.2 & \textbf{74.5 }  \\
\bottomrule
\end{tabular}
}
\label{tab:exp-diverse-benchmarks}
\vspace{-3mm}
\end{table}

As shown in Table \ref{tab:exp-ocr},
we compare our MindVL-671B-A37B with open source models of approximate parameter size and closed source excellent models on OCR, chart, and document understanding benchmarks.
Overall, our MindVL-671B-A37B, trained with 106B multimodal tokens, achieves best average score across above benchmarks.
These results show that our data and training recipes are effective in Vision-Language Alignment for large language models with different parameter scales, i.e., Qwen3-8B and DeepSeek-V3. 

In addition, the comparison results of MindVL-671B-A37B and other models on multi-modal Hallusion and STEM benchmarks, including HallusionBench~\citep{guan2024hallusionbench}, AI2D~\citep{kembhavi2016diagram}, MMVet~\citep{yu2023mm}, MathVista~\citep{lu2023mathvista}, MathVision~\citep{wang2024measuring} and MMMU pro~\citep{yue2024mmmu}, are shown in Table \ref{tab:exp-multimodal}.
MindVL-671B-A37B gets the best evaluation results on HallusionBench. 
Although MindVL-671B-A37B achieves competitive performance against Qwen2-VL-72B and Qwen2.5-VL-72B on overall score,
some of the model's multimodal reasoning capabilities are not sufficiently activated. 
For example, there is still a certain gap between our model and the state-of-the-art models on MMVet, MathVision and MMMU pro.
The reason is that MindVL-671B-A37B is trained with only 106B multimodal tokens.
In the future, we will add rich interleaved image-text data to boost model reasoning ability with broad domain knowledge.

Moreover, as shown in Table \ref{tab:exp-diverse-benchmarks}, we report the pure text evaluation results of MindVL-671B-A37B cross diverse language benchmarks, including AIME2024~\citep{MAA2024}, AIME2025~\citep{MAA2025}, GPQA-D~\citep{rein2024gpqa}, IFEval~\citep{zhou2023instruction}, ArenaHard~\citep{li2024live}, C-SimpleQA~\citep{he2024chinese} and C-Eval~\citep{huang2023c}.
MindVL-671B-A37B outperforms the original DeepSeek-V3 in Overall evaluation result.
Specifically, MindVL-671B-A37B achieves significant improvements on AIME2024, AIME2025, IFEval and AreanaHard datasets, and maintained the pure text capabilities of DeepSeek-V3 on GPQA-D, C-SimpleQA and C-Eval datasets.
In addition, MindVL-671B-A37B also obtains better evaluation results than Qwen2.5VL-32B and very competitive results compared to Qwen2.5VL-72B across these pure text benchmarks.

\begin{table}[!h]
\vspace{-2pt}
  \centering
  \scriptsize
  \setlength{\tabcolsep}{4pt}
  \caption{Studies of model merging strategy of MindVL-8B on multimodal benchmarks.}
    \renewcommand{\arraystretch}{1.2} 
    
\resizebox{\linewidth}{!}{
    \begin{tabular}{l|cc|cccccc|c}
    \toprule
    \textbf{Model} & \makecell{\textbf{Sequence}\\\textbf{Length}} & \makecell{\textbf{Maximum}\\\textbf{Pixels}} & \textbf{MME} & \textbf{MMBench} & \textbf{OCRBench} & \textbf{DocVQA} & \textbf{ChartQA} & \textbf{InfoVQA} & \textbf{Overall} \\
    \midrule
   MindVL-8B-2K & 2K & 1280 & 81.6 & 82.4 & 87.0 & 94.2 & 86.8 & 79.6 & 85.3 \\
   MindVL-8B-4K & 4K & 3072 & \textbf{84.3} & 82.3 & 87.3 & 94.5 & 86.6 & 79.8 & 85.8 \\
   MindVL-8B-8K & 8K & 4096 & 82.5 & 82.5 & 86.0 & 94.6 & 86.4 & 79.9 & 85.3 \\
   
\midrule
   MindVL-8B & - & - & 84.1 & \textbf{84.3} & \textbf{87.6} & \textbf{94.7} & \textbf{87.2} & \textbf{81.1} & \textbf{86.5} \\
    \bottomrule
    
    \end{tabular}%
    }
  \label{tab:sft_merge}%
\end{table}%
\begin{figure}[!h]

    \centering
    \subfigure{
    \includegraphics[width=0.45\textwidth]{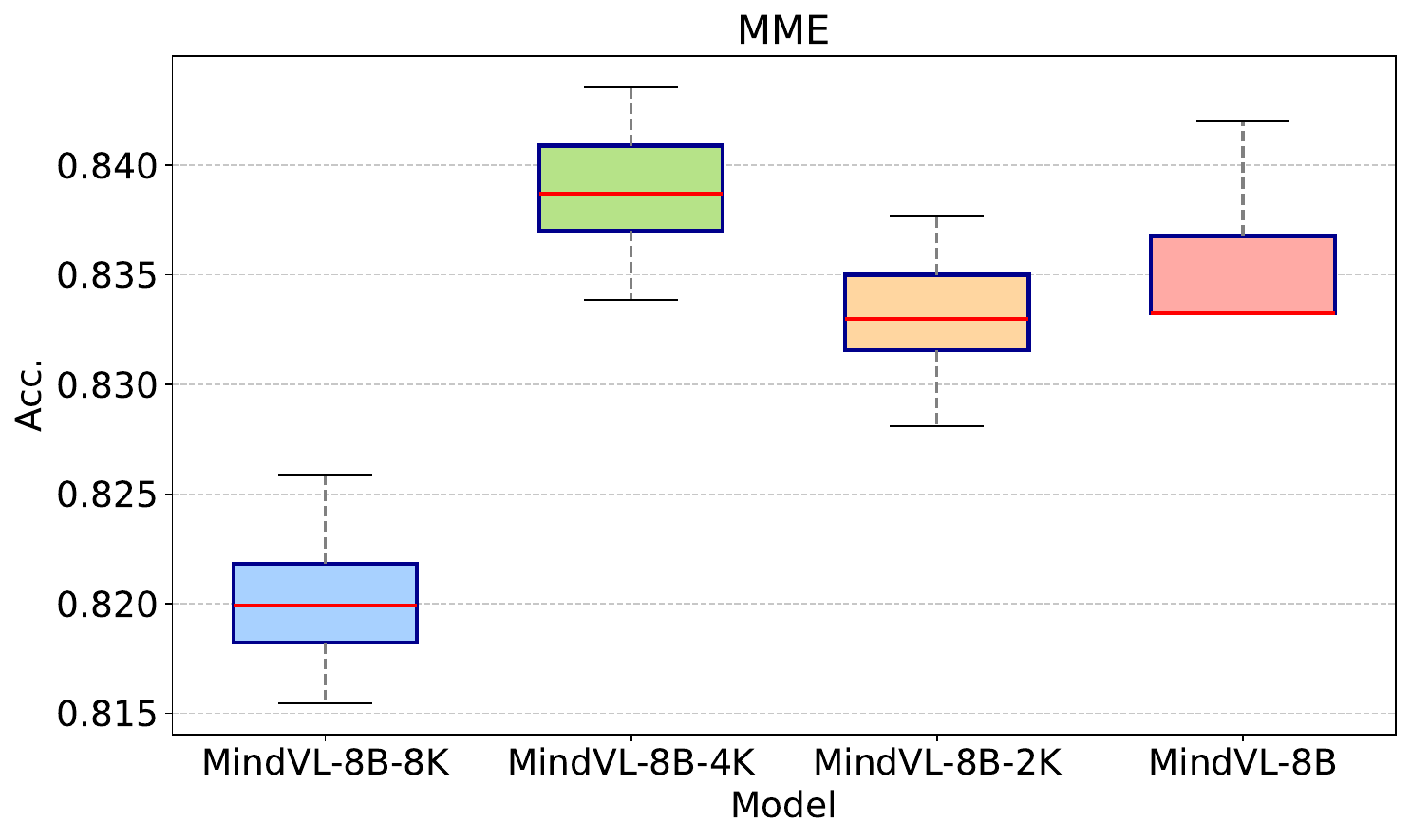}
    \includegraphics[width=0.45\textwidth]{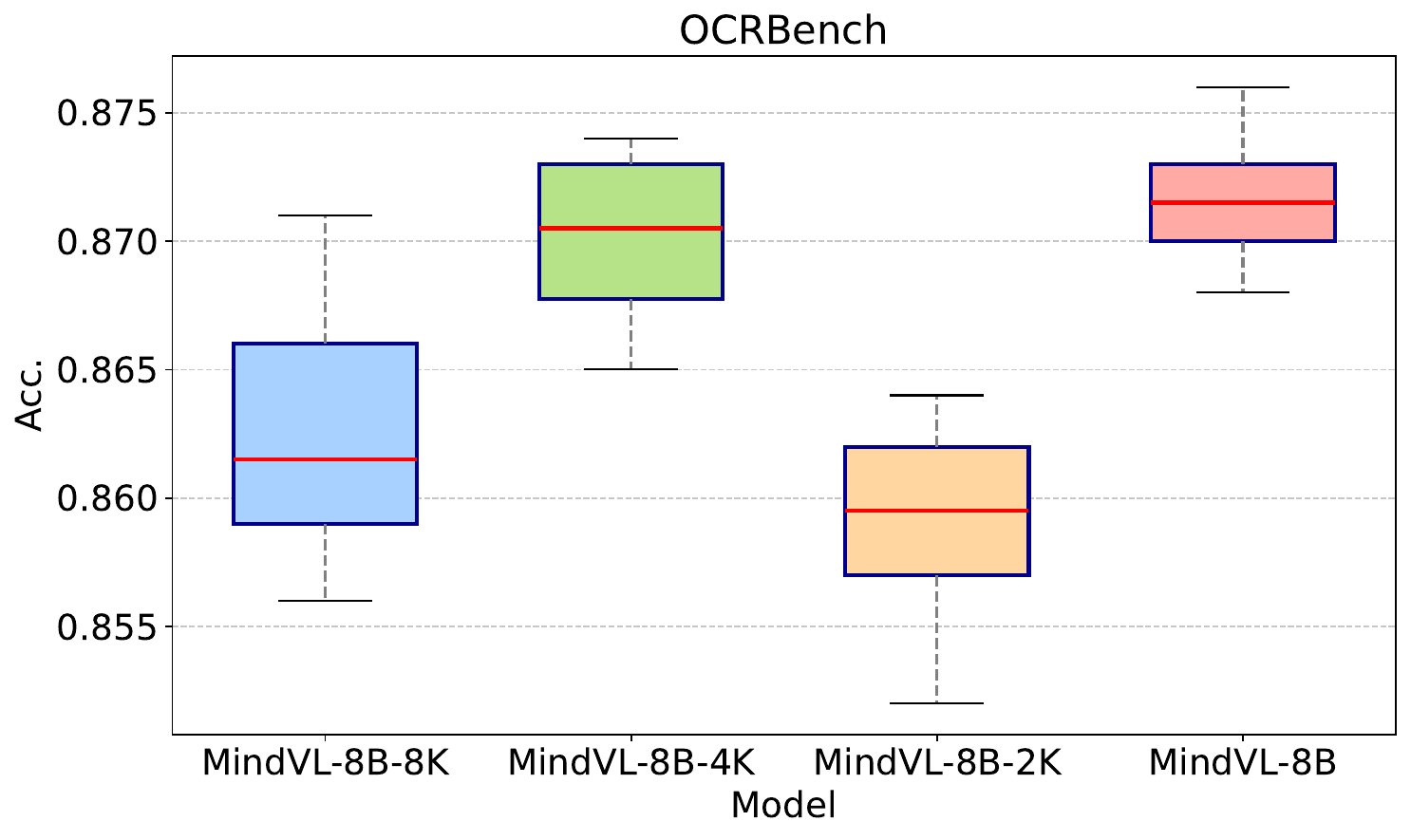}}

    \subfigure{
    \includegraphics[width=0.45\textwidth]{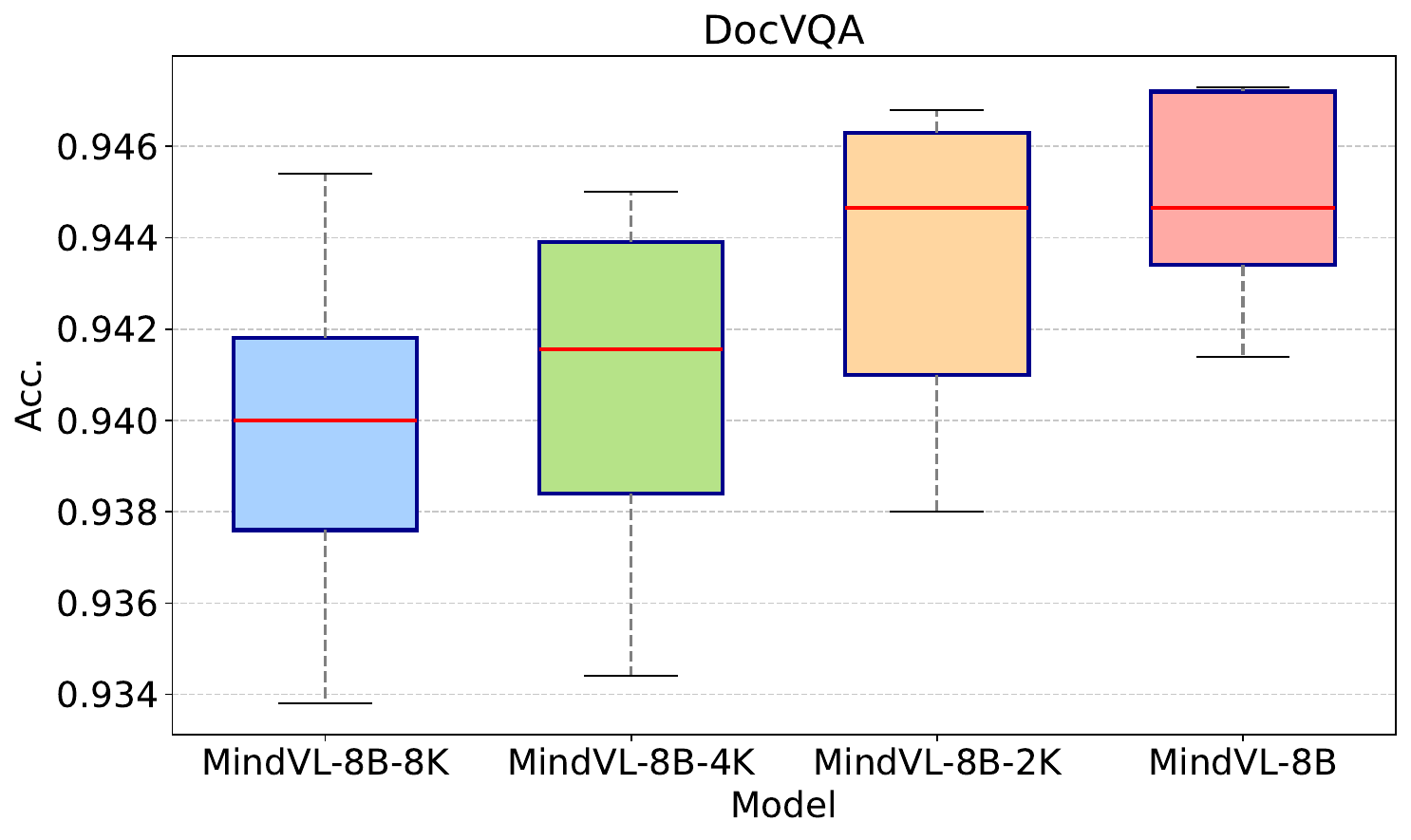}
    \includegraphics[width=0.45\textwidth]{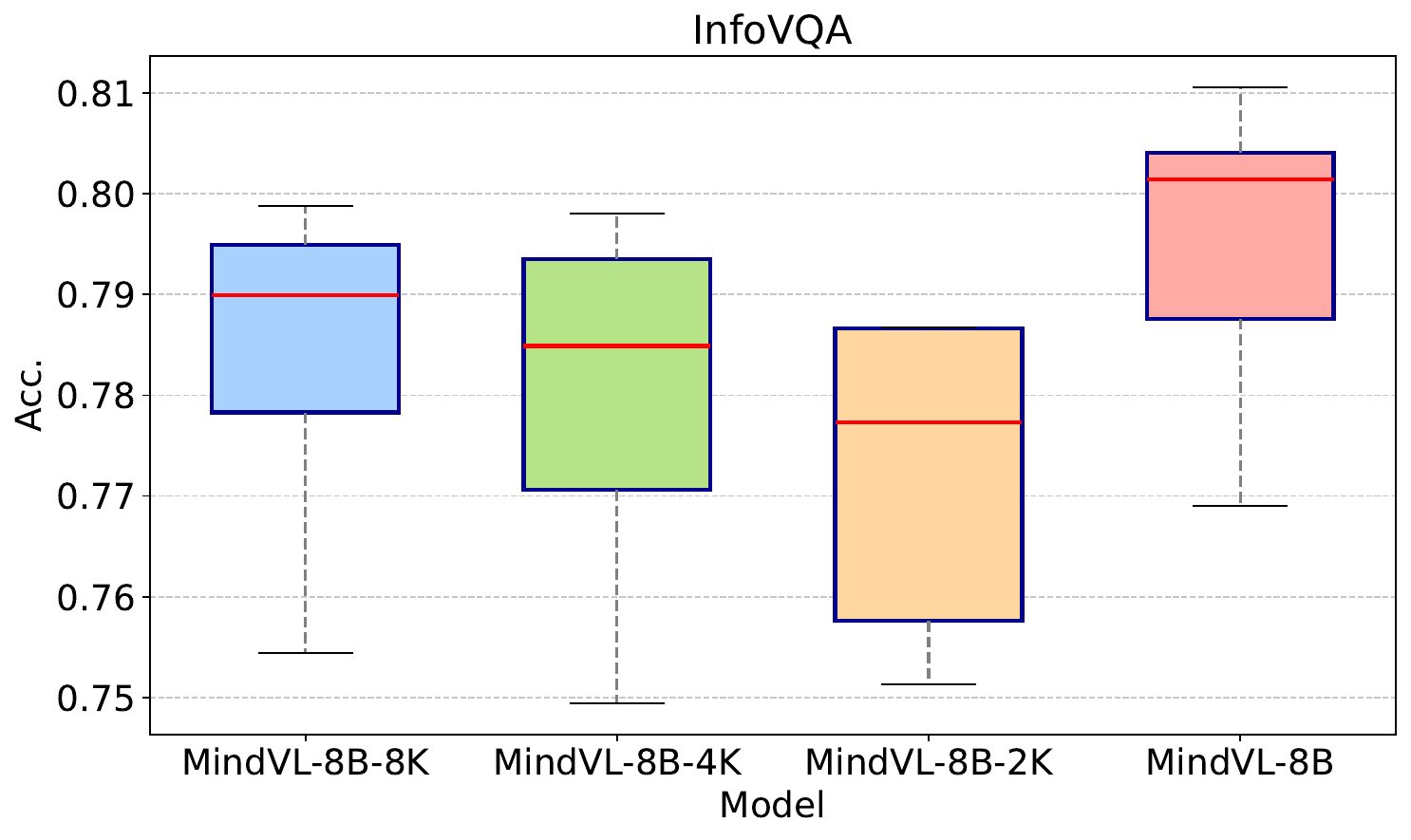}}
\vspace{-8pt}
	\caption{Box plots of accuracies with varying input images resolutions for different models. Maximum value, minimum value, median and quartiles are plotted.}
\vspace{-6pt}
	\label{fig:eval_res_boxes}
\end{figure}

\subsection{Effectiveness of Model Merging with Different Training Sequence Lengths}\label{sec:Effectiveness of Model Merging}

\vspace{-6pt}
Table~\ref{tab:sft_merge} validate the effectiveness of our model merging strategy, which is implemented by averaging the weights of models trained on sequence lengths of 2K, 4K, and 8K—denoted as MindVL-8B-2K, MindVL-8B-4K, and MindVL-8B-8K, respectively. The merged model is denoted as MindVL-8B. Our MindVL-8B model achieves an average benchmark performance of 86.5\%, which are better than the scores of MindVL-8B-2K, MindVL-8B-4K, and MindVL-8B-8K. The proposed strategy significantly surpasses the perceptual capabilities of MindVL-8B, especially on MMBench, ChartQA and InfoVQA datasets.

Figures~\ref{fig:eval_res_boxes} shows the box plots of the evaluation results using test-time image resolution search strategy as mentioned in Section~\ref{sec:test_time_resolution_search}.
On OCRBench, DocVQA and InfoVQA, MindVL-8B outperforms other models not only in terms of the highest accuracy scores across different image resolutions, but also shows improvements in both the lowest scores and median scores.  On MME dataset, MindVL-8B outerperforms MindVL-8B-2K and MindVL-8B-8K.
The above results demonstrate that the model weight merging strategy can enhance the model's robustness against variations in input images of different resolutions.
For specific numerical results, please refer to Appendix~\ref{sec:Test-Time Resolution Search}.

\begin{table}[!t]

\vspace{-4pt}
\centering
\scriptsize
\setlength{\tabcolsep}{3pt}
\caption{Studies of model merging strategy of MindVL-671B-A37B on language benchmarks.}
\resizebox{\linewidth}{!}{
\begin{tabular}{l|ccccccc|c}
\toprule
\textbf{Model} & \textbf{AIME2024} & \textbf{AIME2025} & \textbf{GPQA-D} & \textbf{IFEval} & \textbf{ArenaHard} & \textbf{C-SimpleQA} & \textbf{C-Eval} & \textbf{Overall} \\

    \midrule
DeepSeek-V3-0324~\citep{liu2024deepseek} & 60.0 & 43.3 & \textbf{69.2} & 81.9 & 94.8 &\textbf{ 73.3} & \textbf{89.6} & 73.2 \\
MindVL-671B-A37B-4K & 20.0 & 16.7 & 58.6 & 81.5 & 91.0 & 70.8 & 84.4 & 60.4 \\

\midrule
MindVL-671B-A37B & \textbf{63.3} & \textbf{46.7} & 68.7 &\textbf{ 84.7} & \textbf{97.0} & 72.9 & 88.2 & \textbf{74.5}   \\
\bottomrule
\end{tabular}
}
\label{tab:exp-dslv-merge}
\vspace{-3mm}
\end{table}

In addition, we verified the effectiveness of the model merging strategy for large-scale language model, i.e. DeepSeek-V3, in maintaining the pure text ability after Vision-Language alignment. As shown in Table~\ref{tab:exp-dslv-merge},
MindVL-671B-A37B is obtained by merging the language models of DeepSeek-V3 and MindVL-671B-A37B-4K, i.e., the SFT model trained with 4K sequence length.
MindVL-671B-A37B not only outperforms MindVL-671B-A37B-4K on each benchmark, but also outperforms the original language model DeepSeek-V3 on AIME2024, AIME2025, IFEval, ArenaHard and overall average score.
For multimodal performance, MindVL-671B-A37B achieves the evaluation results similar to MindVL-671B-A37B-4K, as shown in Table~\ref{tab:exp-dslv-merge-mm}.

\begin{table}[!t]
\centering
\scriptsize
\setlength{\tabcolsep}{3pt}
\caption{Studies of model merging strategy of MindVL-671B-A37B on multimodal benchmarks. All results
are uniformly evaluated by the internal testing platform (Correctness Evaluation).}
\begin{tabular}{l|cccccccccc|c}
\toprule
\textbf{Model} & \makecell{\textbf{MM}-\\\textbf{Bench}\\} & \makecell{\textbf{OCR}-\\\textbf{Bench}} & \makecell{\textbf{DocVQA}\\\textbf{Val}} & \makecell{\textbf{ChartQA}} & \makecell{\textbf{InfoVQA}\\\textbf{Val}} & \textbf{AI2D} & \textbf{MMVet} & \makecell{\textbf{Math}-\\\textbf{Vista}} & \makecell{\textbf{Math}-\\\textbf{Vision}} & \makecell{\textbf{MMMU}\\\textbf{pro}} &\textbf{Overall} \\
\midrule
MindVL-671B-A37B-4K  & \textbf{90.8} & \textbf{90.0} & \textbf{96.0} & \textbf{79.6} & 83.0 & \textbf{85.2} & 67.9 & \textbf{72.9} & 33.5&45.0& 74.4\\

\midrule
MindVL-671B-A37B  & 90.7 & 89.4 & 95.8 & 78.7 & \textbf{84.2} & 83.2 & \textbf{70.2} & 70.7 & \textbf{43.0}&\textbf{49.1}& \textbf{75.5} \\
\bottomrule
\end{tabular}
\label{tab:exp-dslv-merge-mm}
\vspace{-1mm}
\end{table}

\vspace{-4pt}
\section{Conclusion}



In this paper, our proposed MindVL, trained on MindSpeed-MLLM framework with Ascend NPUs. We highlight the effectiveness of our data recipe and the robust training capability of the framework on Ascend hardware. Key contributions include the introduction of an efficient full-stage training framework for both Dense and MoE MLLMs, a transparent data recipe, and novel enhancement techniques. Notably, MindVL-8B and MindVL-671B-A37B achieve competitive results on both multimodal benchmark and language benchmark using only a fraction of the training data required by comparable models. Moreover, our proposed model merging method has effectively enhanced the model's performance. We find that model merging is an effective and low-cost approach to improve model performance. In future work, we will further explore the methodologies of model merging and the underlying principles behind it.

\section{Reproducibility Statement}
We have provided a detailed description in the main text and appendix of the training framework MindSpeed-MLLM (Section~\ref{sec:mindspeed} and Appendix~\ref{app:mindspeed-mllm}) used by MindVL, the data and data ratios for each stage (Section~\ref{sec:data} and Appendix~\ref{appendix:data}), and the training hyper-parameters of the model at each stage (Section~\ref{sec:mindvl}), all to facilitate researchers in referencing and reproducing our work. Furthermore, following internal review, we will make the MindSpeed-MLLM code open-source to advance research on multimodal large language models on Ascend NPUs.
\bibliography{iclr2026_conference}
\bibliographystyle{iclr2026_conference}

\newpage
\appendix
\section{MindSpeed-MLLM} \label{app:mindspeed-mllm}

\subsection{Ascend v.s. Nvidia}

\begin{table}[htbp]
  \centering
  \caption{Comparison Between NVIDIA CUDA and Ascend CANN.}
  \renewcommand{\arraystretch}{1.2} 
    \begin{tabular}{lll}
    \toprule
    \textbf{Component} & \textbf{CUDA} & \textbf{CANN} \\
    \midrule
    GE Graph Engine & TensorRT plugins \& parser & Graph Engine \\
    Collective Communication Library & NV NCCL & HCCL \\
    Library/Template & NV CUTLASS & Ascend C High-Level API \\
    General Programming & NV CUDA-C & Ascend C Low-Level API \\
    Operator Acceleration Library & NV cuDNN & Ascend aclNN \\
    Runtime & NV Runtime & Ascend Runtime \\
    Driver & NV Driver & Ascend Driver \\
    \bottomrule
    \end{tabular}%
  \label{infra:env}%
\end{table}%

The ecosystem serves as the cornerstone of distributed pre-training frameworks, with NVIDIA’s CUDA and Huawei’s Ascend CANN being the two mainstream ecosystems.
While both possess dedicated driver and runtime layers for OS-AI accelerator communication as illustrated in Table \ref{infra:env}, they differ significantly in middle-level acceleration libraries, upper-level tools, and ecological maturity. 
NVIDIA’s cuDNN and NCCL are highly optimized, and with CUTLASS and TensorRT providing flexible programming templates and graph optimization, it has formed a mature "hardware-software-tools-community" loop, supported by in-depth adaptation of open source communities like those behind PyTorch and Megatron-LM. 

In contrast, Ascend’s aclNN, HCCL, and two-tier Ascend C API system have realized core functions but lack coverage of long-tail operators, third-party tools, and community resources. 
This gap limits Ascend’s training framework selection—mainstream CUDA-based frameworks cannot directly leverage Ascend’s hardware capabilities, and simple adaptation incurs high costs and performance losses, highlighting the necessity of developing a dedicated framework for Ascend to integrate with CANN components seamlessly.

\begin{table}[htbp]
  \centering
  \caption{Comparison of Differences Between Huawei Ascend and NVIDIA Computing Cards.}
    \renewcommand{\arraystretch}{1.2}
    \begin{tabular}{llllll}
    \toprule
    \textbf{XPU} & \textbf{A100} & \textbf{A800} & \textbf{H800} & \textbf{910B1} & \textbf{910B2} \\
    \midrule
    TF32 (Tensor Core) & 156   & 156   & 495   & 200   & 188 \\
    FP32 (Tensor Core) & NA    & NA    & NA    & 100   & 94 \\
    BF16 (Tensor Core) & 312   & 312   & 989   & 400   & 376 \\
    FP16 (Tensor Core) & 312   & 312   & 989   & 400   & 376 \\
    Int8 (Tensor Core) & 624   & 624   & 1979  & 800   & 752 \\
    HBM (GB) & 80    & 80    & 80    & 64    & 64 \\
    HBM Bandwidth (GB/s) & 2039  & 2039  & 3350  & 1800  & 1800 \\
    NVLink Bandwidth (GB/s) & 600   & 400   & 600   & NA    & NA \\
    PCIe Bandwidth (GB/s) & 64    & 64    & 128   & 64    & 64 \\
    \bottomrule
    \end{tabular}%
  \label{infra:hardware}%
\end{table}%

Beyond underlying chip architecture differences, Ascend and NVIDIA computing cards vary distinctly in core specifications that impact large-model pre-training efficiency and stability as illustrated in \ref{infra:hardware}. 
In terms of computing power, Ascend 910B shows competitiveness at mainstream precisions and unique advantages in FP32 Tensor Core support. 
However, Ascend has obvious shortcomings: 64 GB HBM risks training interruptions for ultra-large models, 1800 GB/s HBM bandwidth is lower than NVIDIA’s, and without NVLink, it relies on 64 GB/s PCIe for multi-card communication, increasing latency. 

To mitigate these, hardware-friendly strategies are needed, which further drives Ascend-specific framework optimizations—such as developing fused operators, dynamic scheduling, and communication optimization to offset shortcomings and leverage its computing power.

\subsection{Precision Alignment}

To verify the forward and backward precision of the developed framework, we conducted comparisons not only with the original MindSpeed-MM but also with the following mainstream training frameworks within the NVIDIA ecosystem:

Llama-Factory\cite{zheng2024llamafactory}: an easy-to-use open-source training framework. 
It is mainly built on native PyTorch and the HuggingFace ecosystem, and supports distributed engines such as DeepSpeed and FSDP. 
Currently, its support for the Ascend ecosystem is becoming increasingly comprehensive.

Pai-Megatron-Patch\cite{pai-megatron}: an open-source training tool developed by Alibaba. 
Similar to MindSpeed-LLM/MM, it includes adaptation and optimization based on Megatron for mainstream open-source LLMs.

\begin{figure}[ht]
	\centering
    \includegraphics[width=1\linewidth]{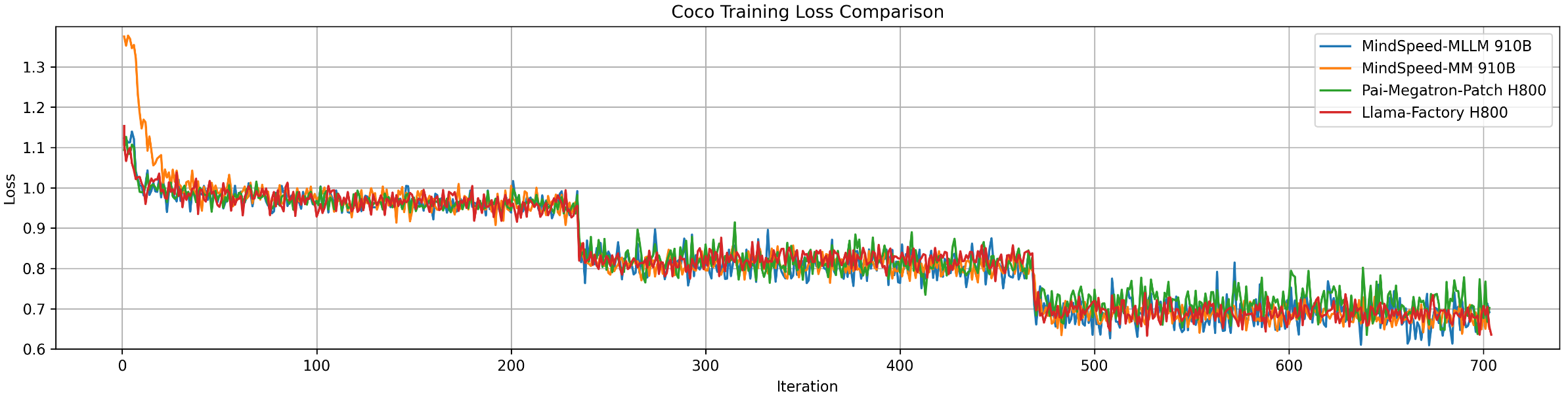}
    \vspace{-4mm}
	\caption{Comparison of loss values among various training tools and platforms on the COCO dataset.}
	\label{fig:coco_loss}
\end{figure}


Two datasets were employed to validate the accuracy of the training framework: the public COCO dataset\citep{DBLP:conf/eccv/LinMBHPRDZ14} and an in-house Slow Thinking dataset.
In all training experiments, consistent training hyper-parameters were maintained. 
Specifically, the training length was set to 4K for the COCO dataset, whereas a 12K training length was adopted for the Slow Thinking dataset.

Firstly, we checked the forward precision of the MindSpeed-MLLM. 
By manually feeding hundreds of identical inputs into MindSpeed-MLLM and the transformer-based forward code, the Mean Absolute Error (MAE) and Mean Relative Error (MRE) are both within five thousandths.

Next, we conducted precision checks through training tasks.
In the COCO data set, as illustrated in Figure \ref{fig:coco_loss}, the loss curves of all frameworks exhibited very similar trends, although with minor discrepancies. These differences are deemed acceptable, as the frameworks utilize distinct data loading modules and workflows.

\begin{figure}[ht]
    \centering
    \includegraphics[width=1\linewidth]{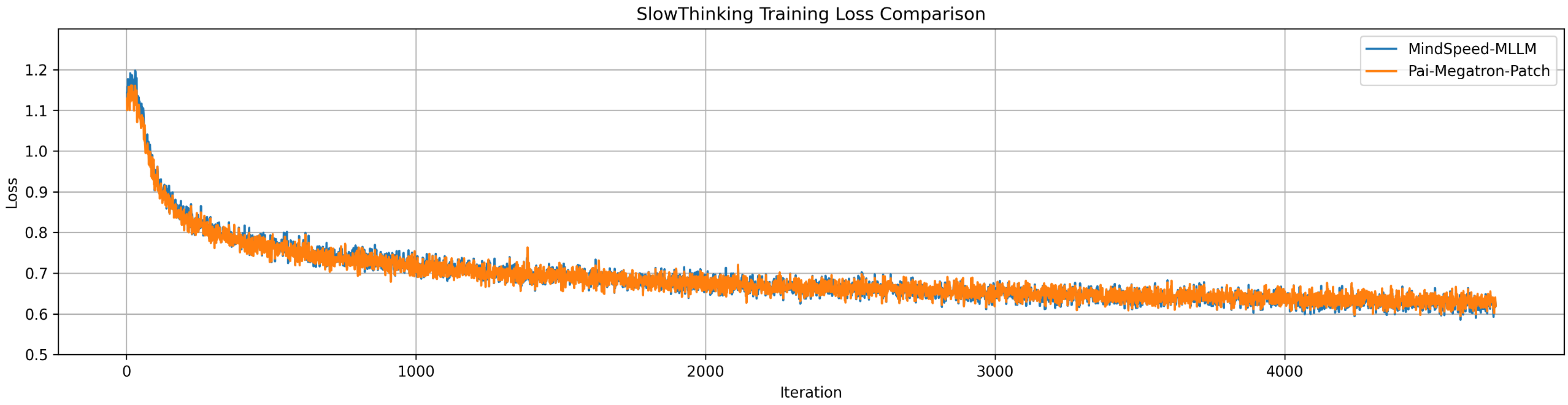}
    \vspace{-4mm}
	\caption{loss decline trend on the in-house slow thinking dataset.}
	\label{fig:st_loss}
\end{figure}

\begin{figure}[ht]
    \centering
    \includegraphics[width=1\linewidth]{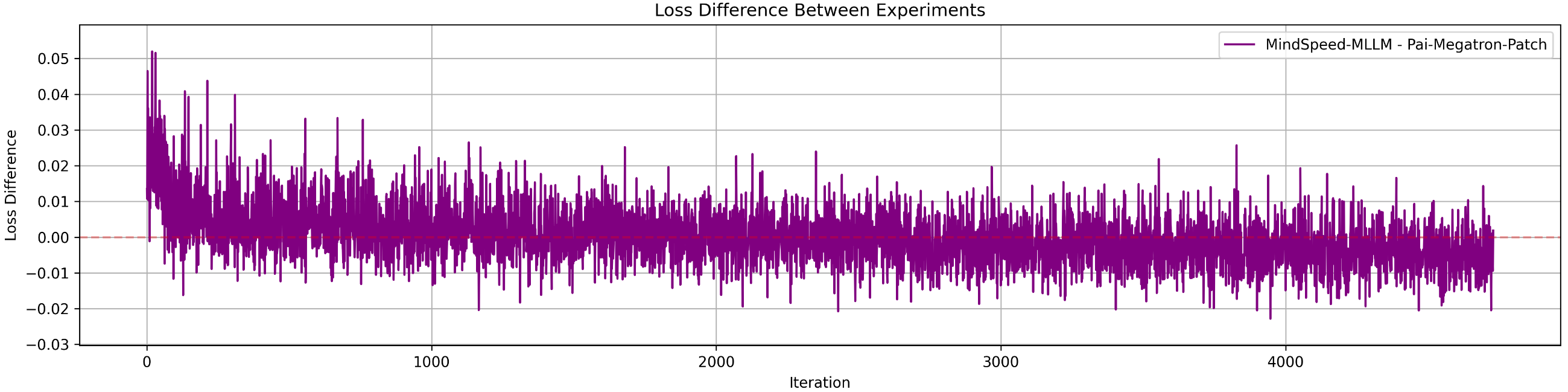}
    \vspace{-4mm}
	\caption{loss difference between MindSpeed-MLLM and the original MindSpeed-MM.}
	\label{fig:st_loss_compare}
\end{figure}

To further validate consistency, we adapted our custom data loader to MindSpeed-MM, ensuring full alignment of the data load logic, and conducted comparative experiments on the in-house Slow Thinking dataset.
As illustrated in Figures \ref{fig:st_loss} and \ref{fig:st_loss_compare}, after standardizing the data loading process, the loss values of MindSpeed-MLLM and MindSpeed-MM showed a near-perfect overlap, with the loss difference confined to within a percentile range.
This marginal discrepancy is considered acceptable, given the inherent different associated with the fused operators employed, as well as the randomness with computation and communication processes.

In the final benchmark evaluation, under the same training set, the models trained using Llama-Factory and MindSpeed-MLLM respectively achieved comparable results on the general visual and text benchmarks, with an error margin within ±1.5.

\section{Data Curation}\label{appendix:data}
The MindVL training corpus contains 447 billion diverse and high-quality tokens used for three training stages. The data is categorized according to target capabilities, and the curation process for each category is detailed in the following subsections.

\subsection{warm-up Data}\label{sec:warm-up}
The MindVL warm-up corpus contains 256 billion diverse, high-quality tokens, including Generic Image-Text Pairs, Optical Character Recognition (OCR), {Visual Grounding \& Counting, Science, Technology, Engineering, and Mathematics (STEM), and Graphical User Interface (GUI).

\begin{table}[tbp]
\centering
\caption{Warm-up Phase Dataset Composition}
\label{tab:warmup-dataset-composition}
\begin{tabular}{llcc}
\toprule
\textbf{Category} & \textbf{Dataset} & \textbf{Total Num} & \textbf{Percentage (\%)} \\
\midrule
Caption & Caption-CN-Rawcaption & 132,900,000 & 21.92 \\
Caption & Caption-EN-Recaption & 102,500,000 & 16.90 \\
Caption & Caption-EN-Rawcaption & 42,041,409 & 6.93 \\
Caption & CapsFusion-en & 35,840,000 & 5.91 \\
Caption & Caption-CN-Recaption & 25,300,000 & 4.17 \\
Caption & HiVision & 1,800,000 & 0.30 \\
\textbf{Caption Subtotal} & & \textbf{338,381,409} & \textbf{55.81} \\
\cmidrule(r){1-4}
Grounding & UMG-en & 42,240,000 & 6.97 \\
Grounding & Grounding & 35,000,000 & 5.77 \\
Grounding & UMG-zh & 14,080,000 & 2.32 \\
Grounding & GRIT-en & 8,448,000 & 1.39 \\
\textbf{Grounding Subtotal} & & \textbf{99,768,000} & \textbf{16.45} \\
\cmidrule(r){1-4}
OCR & WuKong-Text-zh & 33,280,000 & 5.49 \\
OCR & LAION-Text-en & 14,336,000 & 2.36 \\
OCR & Document-OCR & 17,000,000 & 2.80 \\
OCR & Scene-OCR & 17,000,000 & 2.80 \\
OCR & IIT-CDIP-en & 3,840,000 & 0.63 \\
OCR & Rendered-Text-en & 1,920,000 & 0.32 \\
OCR & PDF-en & 2,560,000 & 0.42 \\
OCR & PDF-zh & 2,560,000 & 0.42 \\
OCR & arXiv-en & 2,304,000 & 0.38 \\
OCR & DOCX-en & 960,000 & 0.16 \\
OCR & DOCX-zh & 960,000 & 0.16 \\
OCR & README & 1,280,000 & 0.21 \\
OCR & Scene-Text & 560,000 & 0.09 \\
\textbf{OCR Subtotal} & & \textbf{98,000,000} & \textbf{16.16} \\
\cmidrule(r){1-4}
Table\&Chart & Table-Data & 10,000,000 & 1.65 \\
Table\&Chart & Chart-SFT-en & 1,440,000 & 0.24 \\
Table\&Chart & PubTables-en & 120,000 & 0.02 \\
Table\&Chart & PubTables-zh & 120,000 & 0.02 \\
Table\&Chart & MMC-Align-en & 420,000 & 0.07 \\
Table\&Chart & MMC-Instruct-en & 420,000 & 0.07 \\
\textbf{Table\&Chart Subtotal} & & \textbf{12,520,000} & \textbf{2.06} \\
\cmidrule(r){1-4}
STEM & STEM-Caption-0729-en & 2,230,000 & 0.37 \\
STEM & STEM-Caption-0729-zh & 2,230,000 & 0.37 \\
STEM & STEM-Shuffle & 1,000,000 & 0.16 \\
\textbf{STEM Subtotal} & & \textbf{5,460,000} & \textbf{0.90} \\
\cmidrule(r){1-4}
GUI & Screen-UI-zh & 360,000 & 0.06 \\
GUI & Screen-UI-en & 120,000 & 0.02 \\
GUI & Screen-QA-zh & 120,000 & 0.02 \\
GUI & Screen-Ref-zh & 120,000 & 0.02 \\
GUI & Screen-Navi-zh & 120,000 & 0.02 \\
\textbf{GUI Subtotal} & & \textbf{840,000} & \textbf{0.14} \\
\midrule
\textbf{Total} & & \textbf{606,221,409} & \textbf{100.00} \\
\bottomrule
\end{tabular}
\end{table}

\subsubsection{Generic Image-Text Pairs}
Web-sourced image-text pairs, which include alt texts, captions, and surrounding contextual text, are now available at an unprecedented scale. With billions of examples, they showcase remarkable diversity across visual and textual concepts. However, such data is inherently noisy, often containing texts that are irrelevant or factually inaccurate relative to the corresponding images, and it frequently exhibits class imbalance.

To address these challenges, we have adopted a series of filtering measures, including image-based criteria, text-based criteria, aesthetic scoring, pornography and violence scoring, watermark scoring, CLIP scoring~\citep{radford2021learning}, and URL filtering. For data with inadequate image-text relevance, we use a model~\citep{Qwen2.5-VL,chen2024expanding} to recapture the images. 

Furthermore, to cover more visual concepts and preserve as many types of long-tail visual concepts as possible, we perform clustering on the images within the filtered image-text pairs. We then manually inspect the data in each cluster and map these clusters to the data's category labels. For clusters that represent overly broad concepts, we conduct re-clustering. Finally, we obtain the results of hierarchical clustering: a larger variance within a cluster indicates greater diversity. We perform data sampling based on variance—clusters with larger variance undergo more extensive sampling, while clusters with smaller variance are sampled less.

\subsubsection{Optical Character Recognition (OCR)}
OCR data consists of two types, document OCR data and scene OCR data, each following a structured processing workflow. For document OCR data, the process starts with acquiring and initially screening source PDF data, then processes PDFs page-by-page via a pipeline to generate image-text pairs and image-text interleaved formats. After extracting these data (with image markers removed from pair data and text-only MD files filtered from interleaved data), rule-based cleaning addresses issues like garbled characters or abnormal spaces. For scene OCR data, pre-processed tar packages are first sent to the Focus OCR model~\citep{liu2024focus} to generate OCR results; the same tar packages are then processed by the PaddleOCR model~\citep{cui2025paddleocr}, with results from both models intersected to form production-line OCR results. In addition to the aforementioned process, we have also calculated the coverage rate of words and characters in OCR text, and conducted targeted supplementation for data with relatively low coverage. In addition, we have also incorporated data such as small-language content, ancient books, and calligraphy works.

\subsubsection{Tabel \& Chart}
For raw tables, we perform layout detection and content recognition separately. Specifically, we use the RapidTable~\citep{RapidTable} and LORE~\citep{xing2023lore,long2025lore++} models for layout detection and layout judgment, and PaddleOCR for content recognition. Subsequently, we integrate the results and conduct post-processing for HTML formatting. Finally, we convert the table images into HTML format. Alongside the previously mentioned data, we also collect open data to enhance the ability
to interpret tables and charts. For tables, we use the PubTables-1M dataset~\citep{smock2022pubtables}, including both its original English version and a translated Chinese version, to gather table recognition data. For charts, we employ chart-to-table conversion and chart-based QA data from existing datasets, including MMC~\citep{liu2023mmc} and ChartSFT~\citep{meng2024chartassisstant}. 

\subsubsection{Visual Grounding}
First, we perform recaptioning on the crawled and open-source images. We then use the Florence-2-large model~\citep{xiao2024florence} to conduct caption to phrase grounding processing. After obtaining object bounding boxes and region captions, we employ SAM~\citep{kirillov2023segment} to generate point annotations. In addition, we utilize GrIT-20M~\citep{Kosmos2}, a synthetic caption dataset with additional location labels for major visual elements, to enhance the grounding capability. We incorporate referring and grounding data from UMG-41M~\citep{shi2024umg}.  Specifically, each region is randomly assigned to either a referring task or a grounding task. In the referring task, we provide the model with a bounding box to generate a caption for that specific region, while in the grounding task, we reverse this by using the caption to predict the corresponding bounding box.

\subsubsection{Science, Technology, Engineering, and Mathematics (STEM)}
To enhance the model’s disciplinary knowledge, we incorporated a diverse collection of data across various STEM domains, obtained through both crawling and manual annotation. We referred to the disciplinary classifications of MMMU~\citep{yue2024mmmu} and K12, extracted such data from a vast volume of PDFs, and also collected publicly available data. Additionally, we used a model to re-annotate the sourced disciplinary images. Ultimately, we obtained nearly 5 million data samples.

\subsubsection{Graphical User Interface (GUI)}
For GUI data processing, we first perform pre-annotation via GPT-4V~\citep{hurst2023gpt}, which involves tasks such as Referring\&Grounding (for icons, text, and coitems), ScreenQA, and Navigation. Next, human verification is conducted by human annotators, focusing on Referring-coitem, QA, and the answers for Navigation. After that, a sequence of data post-processing operations is carried out: first, merging with original annotations; second, using UI-Hawk~\citep{zhang2024ui} and GLM-4V~\citep{glm2024chatglm} to annotate referring-icon, and UI-Hawk to annotate navigation; third, employing GPT-4o~\citep{hurst2024gpt} to annotate Navigation actions; fourth, merging the results of these three models and performing consistency checks; finally, rewriting the answer style with Qwen2.5-7B-Instruct to ensure consistency and linguistic quality.

The data used during the warm-up phase is shown in the Table~\ref{tab:warmup-dataset-composition}. Note that the "Total num" in the Table~\ref{tab:warmup-dataset-composition} is an approximate value. 606,221,409 samples are about 256B tokens. Apart from CapsFusion, UMG, GRIT, PubTables, ChartSFT, and MMC, all other datasets are either self-constructed or have undergone further processing of open-source data. The 16B tokens of data used in MindVL-671B-A37B are uniformly sampled from all warm-up datasets.

\subsection{Multitask Training Data}\label{sec:appendix_multitask-training-data}
 Multitask training data consists of internleaved image-text (Table \ref{tab:interleaved}), visual instruction (Tabel \ref{tab:multitask}, about 80B tokens), web2code and text instruction (Tabel \ref{tab:language-statistics}), totally 179B tokens.

\subsubsection{Internleaved Image-Text Data}
For interleaved image-text data, we primarily crawl Chinese websites, English websites, news websites, and Wikipedia from the internet. We then use model-based methods to filter the crawled data. First, we apply CLIP scores to filter the data according to image-text relevance. Next, we task the model with scoring the image-text interleaved data across multiple dimensions, including question-and-answer quality, relevance between images and questions, complementarity between answers, images and questions, and balance of information density. The model is also required to provide justifications for these scores. Finally, we use these scores to filter the image-text interleaved data.

\begin{table}[htbp]
  \centering
  \caption{Data Statistics and Proportion of Interleaved Image-Text Data. ``GPT4o'' / ``QWEN'' denotes we use GPT4o / Qwen2.5-VL 72B filter the crawled data.}
  \resizebox{\linewidth}{!}{
  \begin{tabular}{lrrrr}
    \toprule
    \textbf{Data Category}          & \textbf{Single-image Count} & \textbf{Multi-image Count} & \textbf{Total Count} & \textbf{Proportion (\%)} \\
    \midrule
    HTML\-EN\-GPT4o                   & 2994750                     & 2994750                     & 5989500              & 12.24                    \\
    HTML\-EN\-QWEN             & 356942                      & 356942                      & 713884               & 1.46                     \\
    HTML\-ZH\-GPT4o                   & 763016                      & 763016                      & 1526032              & 3.12                     \\
    HTML\-ZH\-QWEN             & 8096192                     & 8096192                     & 16192384             & 33.09                    \\
    OBELICS\-GPT4o                    & 3339440                     & 3339440                     & 6678880              & 13.65                    \\
    News Webpages                  & 4167856                     & 4167856                     & 8335712              & 17.03                    \\
    Knowledge & —                           & —                           & 7500000              & 15.33                    \\
    Wikiweb2m   & —                           & —                           & 2000000              & 4.09                     \\
    \midrule
    \textbf{Total}                 & —                           & —                           & \textbf{48936392}    & \textbf{100.00}          \\
    \bottomrule
  \end{tabular}}
  \label{tab:interleaved}
\end{table}

For other types of multitask training data, we have collected a wide range of open source data, as presented in Table~\ref{tab:multitask}. Within this dataset, we used Qwen2.5VL-72B to reannotate certain portions, such as those whose names begin with "Qwen". In addition, we also sampled some data from the warm-up phase and incorporated it into the training process at this stage. 200M web2code data is also used in this stage. Furthermore, we utilized language data produced by DeepSeek R1 to preserve the MLLM's linguistic abilities as shown in Tabel~\ref{tab:language-statistics}, with the ratio of multimodal to language data being approximately 8:2 at this stage.

\begin{table}[htbp]
  \centering
  \caption{Data Statistics and Proportion of text instruction.}
  \small 
  \begin{tabular}{lrr}
    \toprule
    \textbf{Data Source} & \textbf{Number} & \textbf{Proportion (\%)} \\
    \midrule
    Competition Mathematics & 345606 & 16.06 \\
    Code & 113064 & 5.25 \\
    Math & 296500 & 13.78 \\ 
    General QA & 1316361 & 61.18 \\
    Tulu Instruction & 19118 & 0.89 \\
    Multi-turn Conversation & 61114 & 2.84 \\
    \midrule
    \textbf{Total} & \textbf{2151763} & \textbf{100.00} \\
    \bottomrule
  \end{tabular}
  \label{tab:language-statistics}
\end{table}

\begin{longtable}{c l r c}
  \caption{Data Statistics and Proportion of Visual Instruction}\label{tab:multitask} \\
  \toprule
  \textbf{No.} & \textbf{Data Set} & \textbf{Number} & \textbf{Percentage (\%)} \\
  \midrule
  \endfirsthead 
  \caption*{Data Statistics and Proportion of Visual Instruction (Continue)} \\
  \toprule
  \textbf{No.} & \textbf{Data Set} & \textbf{Number} & \textbf{Percentage (\%)}  \\
  \midrule
  \endhead 
  \bottomrule
  \endlastfoot 

  1 & Cauldron/ai2d & 2434 & 0.0064 \\
  2 & Cauldron/cocoqa & 46287 & 0.1214 \\
  3 & Cauldron/datikz & 47974 & 0.1258 \\
  4 & Cauldron/diagram-image-to-text & 300 & 0.0008 \\
  5 & Cauldron/docvqa & 10189 & 0.0267 \\
  6 & Cauldron/dvqa & 199937 & 0.5242 \\
  7 & Cauldron/figureqa & 100000 & 0.2622 \\
  8 & Cauldron/finqa & 5276 & 0.0138 \\
  9 & Cauldron/geomverse & 9303 & 0.0244 \\
  10 & Cauldron/hateful-memes & 8500 & 0.0223 \\
  11 & Cauldron/hitab & 2500 & 0.0066 \\
  12 & Cauldron/iam & 5663 & 0.0148 \\
  13 & Cauldron/iconqa & 27307 & 0.0716 \\
  14 & Cauldron/infographic-vqa & 2118 & 0.0056 \\
  15 & Cauldron/intergps & 1280 & 0.0034 \\
  16 & Cauldron/localized-narratives & 199998 & 0.5244 \\
  17 & Cauldron/mapqa & 37417 & 0.0981 \\
  18 & Cauldron/mimic-cgd & 70939 & 0.1860 \\
  19 & Cauldron/multihiertt & 7619 & 0.0200 \\
  20 & Cauldron/nlvr2 & 50426 & 0.1322 \\
  21 & Cauldron/plotqa & 157007 & 0.4117 \\
  22 & Cauldron/raven & 42000 & 0.1091 \\
  23 & Cauldron/rendered-text & 10000 & 0.0262 \\
  24 & Cauldron/robut-sqa & 8514 & 0.0223 \\
  25 & Cauldron/robut-wikisql & 74989 & 0.1966 \\
  26 & Cauldron/robut-wtq & 38246 & 0.1003 \\
  27 & Cauldron/scienceqa & 4976 & 0.0130 \\
  28 & Cauldron/screen2words & 15730 & 0.0412 \\
  29 & Cauldron/spot-the-diff & 8566 & 0.0225 \\
  30 & Cauldron/tabmwp & 22722 & 0.0596 \\
  31 & Cauldron/tallyqa & 98680 & 0.2587 \\
  32 & Cauldron/tat-qa & 2199 & 0.0058 \\
  33 & Cauldron/textcaps & 21953 & 0.0576 \\
  34 & Cauldron/textvqa & 21953 & 0.0576 \\
  35 & Cauldron/tqa & 1493 & 0.0039 \\
  36 & Cauldron/vistext & 9969 & 0.0261 \\
  37 & Cauldron/visual7w & 14366 & 0.0377 \\
  38 & Cauldron/visualmrc & 3027 & 0.0079 \\
  39 & Cauldron/vqarad & 313 & 0.0008 \\
  40 & Cauldron/vsr & 2157 & 0.0056 \\
  41 & Cauldron/websight & 10000 & 0.0262 \\
  42 & ChartQA/augmented & 15474 & 0.0406 \\
  43 & ChartQA/cap & 18207 & 0.0477 \\
  44 & ChartQA/cot & 18215 & 0.0478 \\
  45 & ChartQA/gemini-v & 16393 & 0.0430 \\
  46 & ChartQA/gemini-v-cot & 16393 & 0.0430 \\
  47 & ChartQA/human & 3699 & 0.0097 \\
  48 & ChartSFT/unichart & 120554 & 0.3161 \\
  49 & CoSyn-400K/chart & 116807 & 0.3063 \\
  50 & CoSyn-400K/chemical & 8935 & 0.0234 \\
  51 & CoSyn-400K/circuit & 10463 & 0.0274 \\
  52 & CoSyn-400K/diagram & 34956 & 0.0916 \\
  53 & CoSyn-400K/document & 71275 & 0.1869 \\
  54 & CoSyn-400K/graphic & 26961 & 0.0707 \\
  55 & CoSyn-400K/math & 66707 & 0.1749 \\
  56 & CoSyn-400K/music & 11962 & 0.0314 \\
  57 & CoSyn-400K/nutrition & 6924 & 0.0181 \\
  58 & CoSyn-400K/table & 46511 & 0.1220 \\
  59 & CoSyn-point & 19385 & 0.0508 \\
  60 & DocVQA/cap & 9892 & 0.0259 \\
  61 & DocVQA/cot & 11938 & 0.0313 \\
  62 & DocVQA/gemini-v & 11182 & 0.0293 \\
  63 & DocVQA/gemini-v-cot & 11182 & 0.0293 \\
  64 & DocVQA/human & 10194 & 0.0267 \\
  65 & InfoVQA/cap & 4290 & 0.0112 \\
  66 & InfoVQA/cot & 6060 & 0.0159 \\
  67 & InfoVQA/gemini-v & 5187 & 0.0136 \\
  68 & InfoVQA/gemini-v-cot & 5187 & 0.0136 \\
  69 & InfoVQA/human & 4406 & 0.0116 \\
  70 & M4-Instruct/3D-LLM-3-datasets & 49890 & 0.1308 \\
  71 & M4-Instruct/AESOP & 6915 & 0.0181 \\
  72 & M4-Instruct/ALFRED & 22565 & 0.0592 \\
  73 & M4-Instruct/Birds-to-Words & 14281 & 0.0374 \\
  74 & M4-Instruct/CLEVR-Change & 3885 & 0.0102 \\
  75 & M4-Instruct/FlintstonesSV & 22341 & 0.0586 \\
  76 & M4-Instruct/HQ-Edit & 50000 & 0.1311 \\
  77 & M4-Instruct/HQ-Edit-Diff & 7000 & 0.0184 \\
  78 & M4-Instruct/IEdit & 3456 & 0.0091 \\
  79 & M4-Instruct/MIT-States-PropertyCoherence & 1900 & 0.0050 \\
  80 & M4-Instruct/MIT-States-StateCoherence & 1900 & 0.0050 \\
  81 & M4-Instruct/MagicBrush & 14249 & 0.0374 \\
  82 & M4-Instruct/MagicBrush-Diff & 6698 & 0.0176 \\
  83 & M4-Instruct/PororoSV & 12299 & 0.0322 \\
  84 & M4-Instruct/RAVEN & 35000 & 0.0918 \\
  85 & M4-Instruct/RecipeQA-ImageCoherence & 8699 & 0.0228 \\
  86 & M4-Instruct/ScanQA & 25563 & 0.0670 \\
  87 & M4-Instruct/TQA & 8249 & 0.0216 \\
  88 & M4-Instruct/VISION & 9900 & 0.0260 \\
  89 & M4-Instruct/VIST & 26026 & 0.0682 \\
  90 & M4-Instruct/WebQA & 9338 & 0.0245 \\
  91 & M4-Instruct/coinstruct & 50000 & 0.1311 \\
  92 & M4-Instruct/contrastive-caption & 25240 & 0.0662 \\
  93 & M4-Instruct/dreamsim & 15941 & 0.0418 \\
  94 & M4-Instruct/iconqa & 34603 & 0.0907 \\
  95 & M4-Instruct/imagecode & 16594 & 0.0435 \\
  96 & M4-Instruct/multi-vqa & 4993 & 0.0131 \\
  97 & M4-Instruct/nextqa & 3870 & 0.0102 \\
  98 & M4-Instruct/nlvr2 & 86373 & 0.2265 \\
  99 & M4-Instruct/star & 3032 & 0.0079 \\
  100 & M4-Instruct/twitter-post & 5734 & 0.0150 \\
  101 & Markdown/arxiv-v2 & 200000 & 0.5244 \\
  102 & Markdown/docx-en & 200000 & 0.5244 \\
  103 & Markdown/docx-zh & 200000 & 0.5244 \\
  104 & Markdown/pubtables-c & 200000 & 0.5244 \\
  105 & Markdown/pubtables-e & 200000 & 0.5244 \\
  106 & Markdown/pubtables-o & 200000 & 0.5244 \\
  107 & Markdown/readme-v2 & 200000 & 0.5244 \\
  108 & Math-PUMA/Synthesis-train-answer & 120239 & 0.3153 \\
  109 & Math-PUMA/Synthesis-train-instruction & 120239 & 0.3153 \\
  110 & Math-PUMA/VarsityTutors-pure-text & 232959 & 0.6108 \\
  111 & Math-PUMA/VarsityTutors-with-image & 78935 & 0.2070 \\
  112 & Math-PUMA/VisualWebInstruct-train-answer & 263599 & 0.6911 \\
  113 & OCR/iit-cdip-v2-en & 200000 & 0.5244 \\
  114 & OCR/laion-text-en & 50000 & 0.1311 \\
  115 & OCR/laion-text-v2-en & 200000 & 0.5244 \\
  116 & OCR/pdf-en & 200000 & 0.5244 \\
  117 & OCR/pdf-zh & 200000 & 0.5244 \\
  118 & OCR/wukong-text-v2-zh & 200000 & 0.5244 \\
  119 & OCR/wukong-text-zh & 50000 & 0.1311 \\
  120 & OneVision/ai2d-gpt4v & 4833 & 0.0127 \\
  121 & OneVision/ai2d-internvl & 12372 & 0.0324 \\
  122 & OneVision/clevr-math-mathv360k & 5249 & 0.0138 \\
  123 & OneVision/figureqa-mathv360k & 17556 & 0.0460 \\
  124 & OneVision/geo3k & 2060 & 0.0054 \\
  125 & OneVision/geometry3k-mathv360k & 9693 & 0.0254 \\
  126 & OneVision/geoqa-plus-mathv360k & 17131 & 0.0449 \\
  127 & OneVision/geos-mathv360k & 467 & 0.0012 \\
  128 & OneVision/iconqa-mathv360k & 22558 & 0.0591 \\
  129 & OneVision/iiit5k & 1959 & 0.0051 \\
  130 & OneVision/image-textualization-filtered & 99542 & 0.2610 \\
  131 & OneVision/infographic-gpt4v & 1951 & 0.0051 \\
  132 & OneVision/k12-printing & 256605 & 0.6728 \\
  133 & OneVision/lrv-chart & 1745 & 0.0046 \\
  134 & OneVision/lrv-normal-filtered & 10456 & 0.0274 \\
  135 & OneVision/mapqa-mathv360k & 5194 & 0.0136 \\
  136 & OneVision/orand-car-a & 1968 & 0.0052 \\
  137 & OneVision/pmc-vqa-mathv360k & 35917 & 0.0942 \\
  138 & OneVision/scienceqa-nona-context & 19177 & 0.0503 \\
  139 & OneVision/super-clevr-mathv360k & 8611 & 0.0226 \\
  140 & OneVision/tabmwp-mathv360k & 22421 & 0.0588 \\
  141 & OneVision/unigeo-mathv360k & 11918 & 0.0312 \\
  142 & OneVision/vision-flan-filtered & 186029 & 0.4877 \\
  143 & OneVision/vizwiz-mathv360k & 6573 & 0.0172 \\
  144 & PixMo/ask-model-anything & 161737 & 0.4241 \\
  145 & PixMo/ask-model-anything-zh & 140343 & 0.3679 \\
  146 & PixMo/cap & 717042 & 1.8800 \\
  147 & PixMo/cap-qa & 271714 & 0.7124 \\
  148 & PixMo/cap-qa-zh & 194097 & 0.5090 \\
  149 & PixMo/cap-zh & 508134 & 1.3323 \\
  150 & PixMo/count & 36916 & 0.0968 \\
  151 & PixMo/docs-charts & 116814 & 0.3063 \\
  152 & PixMo/docs-diagrams & 16551 & 0.0434 \\
  153 & PixMo/docs-other & 71282 & 0.1869 \\
  154 & PixMo/docs-tables & 46518 & 0.1220 \\
  155 & PixMo/point-explanations & 79551 & 0.2086 \\
  156 & PixMo/points & 2376222 & 6.2301 \\
  157 & QwenCaps/PixMo/cap & 712401 & 1.8680 \\
  158 & QwenCaps/PixMo/cap-zh & 642166 & 1.6837 \\
  159 & QwenCaps/ai2d & 9778 & 0.0256 \\
  160 & QwenCaps/allava-laion & 480858 & 1.2608 \\
  161 & QwenCaps/chart2text & 16605 & 0.0435 \\
  162 & QwenCaps/chart-to-text & 44096 & 0.1156 \\
  163 & QwenCaps/chartqa & 18317 & 0.0480 \\
  164 & QwenCaps/coco & 163119 & 0.4277 \\
  165 & QwenCaps/docvqa & 11166 & 0.0293 \\
  166 & QwenCaps/flickr30k & 31772 & 0.0833 \\
  167 & QwenCaps/funsd & 149 & 0.0004 \\
  168 & QwenCaps/infovqa & 4890 & 0.0128 \\
  169 & QwenCaps/laion-gpt4v & 10962 & 0.0287 \\
  170 & QwenCaps/llava & 557086 & 1.4606 \\
  171 & QwenCaps/llavar-s1 & 47231 & 0.1238 \\
  172 & QwenCaps/ocr-vqa & 208190 & 0.5458 \\
  173 & QwenCaps/opencqa & 7724 & 0.0202 \\
  174 & QwenCaps/poie & 2999 & 0.0079 \\
  175 & QwenCaps/sam & 33533 & 0.0879 \\
  176 & QwenCaps/scienceqa & 4985 & 0.0131 \\
  177 & QwenCaps/screen2words & 15669 & 0.0411 \\
  178 & QwenCaps/sroie & 608 & 0.0016 \\
  179 & QwenCaps/textvqa & 24990 & 0.0655 \\
  180 & QwenCaps/vg & 105345 & 0.2762 \\
  181 & QwenCaps/vision-flan & 181189 & 0.4751 \\
  182 & QwenCaps/vistext & 8816 & 0.0231 \\
  183 & QwenCaps/xfund & 1043 & 0.0027 \\
  184 & QwenQA/Cauldron/cocoqa & 27022 & 0.0708 \\
  185 & QwenQA/Cauldron/diagram-image-to-text & 297 & 0.0008 \\
  186 & QwenQA/Cauldron/hateful-memes & 3729 & 0.0098 \\
  187 & QwenQA/Cauldron/mapqa & 22950 & 0.0602 \\
  188 & QwenQA/Cauldron/tabmwp & 22410 & 0.0588 \\
  189 & QwenQA/Cauldron/tallyqa & 83671 & 0.2194 \\
  190 & QwenQA/Cauldron/textvqa & 16059 & 0.0421 \\
  191 & QwenQA/Cauldron/tqa & 861 & 0.0023 \\
  192 & QwenQA/Cauldron/visual7w & 13006 & 0.0341 \\
  193 & QwenQA/Cauldron/vsr & 1650 & 0.0043 \\
  194 & QwenQA/ChartQA/augmented & 12469 & 0.0327 \\
  195 & QwenQA/ChartQA/human & 3290 & 0.0086 \\
  196 & QwenQA/DocVQA/human & 8924 & 0.0234 \\
  197 & QwenQA/InfoVQA/human & 4092 & 0.0107 \\
  198 & QwenQA/PixMo/cap-qa & 247256 & 0.6483 \\
  199 & QwenQA/PixMo/cap-qa-zh & 218868 & 0.5738 \\
  200 & QwenQA/ai2d & 10593 & 0.0278 \\
  201 & QwenQA/allava-laion & 484257 & 1.2697 \\
  202 & QwenQA/allava-vflan & 179020 & 0.4694 \\
  203 & QwenQA/aokvqa & 39358 & 0.1032 \\
  204 & QwenQA/gqa & 69198 & 0.1814 \\
  205 & QwenQA/okvqa & 3442 & 0.0090 \\
  206 & QwenQA/raw/ChartQA/augmented & 15474 & 0.0406 \\
  207 & QwenQA/raw/ChartQA/human & 3699 & 0.0097 \\
  208 & QwenQA/raw/DocVQA/human & 10194 & 0.0267 \\
  209 & QwenQA/raw/InfoVQA/human & 4394 & 0.0115 \\
  210 & QwenQA/raw/ai2d & 12413 & 0.0325 \\
  211 & QwenQA/raw/aokvqa & 76762 & 0.2013 \\
  212 & QwenQA/raw/gqa & 72140 & 0.1891 \\
  213 & QwenQA/raw/okvqa & 8996 & 0.0236 \\
  214 & QwenQA/raw/scienceqa & 12726 & 0.0334 \\
  215 & QwenQA/raw/stvqa & 18921 & 0.0496 \\
  216 & QwenQA/raw/tabfact & 13127 & 0.0344 \\
  217 & QwenQA/raw/vqav2 & 82783 & 0.2171 \\
  218 & QwenQA/raw/wtq & 1679 & 0.0044 \\
  219 & QwenQA/scienceqa & 11071 & 0.0290 \\
  220 & QwenQA/stvqa & 14381 & 0.0377 \\
  221 & QwenQA/tabfact & 12648 & 0.0332 \\
  222 & QwenQA/vqav2 & 78718 & 0.2064 \\
  223 & QwenQA/wtq & 1588 & 0.0042 \\
  224 & R1-Vision/PixMo/cap-qa & 260144 & 0.6821 \\
  225 & R1-Vision/PixMo/cap-qa-zh & 218962 & 0.5741 \\
  226 & Screen/grd & 212472 & 0.5571 \\
  227 & Screen/grd-small & 212472 & 0.5571 \\
  228 & Screen/nav & 286364 & 0.7508 \\
  229 & Screen/nav-small & 286364 & 0.7508 \\
  230 & Screen/ref & 402742 & 1.0560 \\
  231 & Screen/ref-small & 402742 & 1.0560 \\
  232 & Screen/sqa & 417579 & 1.0948 \\
  233 & Screen/sqa-small & 417579 & 1.0948 \\
  234 & Screen/ui-en & 50000 & 0.1311 \\
  235 & Screen/ui-en-small & 50000 & 0.1311 \\
  236 & Screen/ui-zh & 50000 & 0.1311 \\
  237 & Screen/ui-zh-small & 50000 & 0.1311 \\
  238 & XFUND/zh & 149 & 0.0004 \\
  239 & XFUND/zh-qa & 671 & 0.0018 \\
  240 & Allava-caption-laion & 505588 & 1.3256 \\
  241 & Allava-caption-vflan & 182864 & 0.4794 \\
  242 & Allava-instruct-laion & 505588 & 1.3256 \\
  243 & Allava-instruct-vflan & 183839 & 0.4820 \\
  244 & Aokvqa & 76762 & 0.2013 \\
    245 & Chartsft & 808768 & 2.1205 \\
    246 & Cipc-docvqa & 5243 & 0.0137 \\
    247 & Coavqa-bench & 36019 & 0.0944 \\
    248 & Coavqa-chat & 18004 & 0.0472 \\
    249 & Coco-rec & 130723 & 0.3427 \\
    250 & Coco-reg & 130723 & 0.3427 \\
    251 & Crohme & 9821 & 0.0257 \\
    252 & Datikz-v2 & 94532 & 0.2479 \\
    253 & Dcc & 97306 & 0.2551 \\
    254 & Docmatix & 1249967 & 3.2773 \\
    255 & E2e-ocr & 259511 & 0.6804 \\
    256 & E2e-ocr-0324 & 359996 & 0.9440 \\
    257 & Eikie-qa & 183639 & 0.4815 \\
    258 & Estvqa & 17047 & 0.0447 \\
    259 & Estvqa-ca & 17047 & 0.0447 \\
    260 & Estvqa-grounding & 17047 & 0.0447 \\
    261 & Flickr & 148915 & 0.3904 \\
    262 & Funsd & 149 & 0.0004 \\
    263 & Funsd-qa & 431 & 0.0011 \\
    264 & Geo170k-alignment & 60252 & 0.1580 \\
    265 & Geo170k-qa-tuning & 117205 & 0.3073 \\
    266 & Geo-synth-caption-1108 & 320000 & 0.8390 \\
    267 & Geo-synth-qa-1108 & 320004 & 0.8390 \\
    268 & Gqa & 72140 & 0.1891 \\
    269 & Hme & 74502 & 0.1953 \\
    270 & Kvqa & 24602 & 0.0645 \\
    271 & Laion-gpt4v & 12356 & 0.0324 \\
    272 & Llava-s2 & 157712 & 0.4135 \\
    273 & Llarar-s2 & 19800 & 0.0519 \\
    274 & Lrv-instruction & 180722 & 0.4738 \\
    275 & Lvis-instruct4v & 222711 & 0.5840 \\
    276 & M2e & 99956 & 0.2621 \\
    277 & Mavis-900k & 458042 & 1.2009 \\
    278 & Mavis-caps & 599748 & 1.5725 \\
    279 & Mlhme & 30000 & 0.0786 \\
    280 & Mmc-non-arxiv-2 & 21103 & 0.0553 \\
    281 & Mmc-non-arxiv-3 & 88917 & 0.2331 \\
    282 & Noahcaps-en & 14956 & 0.0392 \\
    283 & Noahcaps-zh & 44394 & 0.1164 \\
    284 & Objects365 & 1737995 & 4.5568 \\
    285 & Ocrvqa & 207572 & 0.5442 \\
    286 & Okvqa & 8996 & 0.0236 \\
    287 & Petal-wiki & 2396596 & 6.2836 \\
    288 & Poie & 2250 & 0.0059 \\
    289 & Pointqa-general & 82254 & 0.2157 \\
    290 & Pointqa-local & 10403 & 0.0273 \\
    291 & Pointqa-looktwice & 22838 & 0.0599 \\
    292 & Sharegpt4o-image & 57289 & 0.1502 \\
    293 & Sharegpt4o-video & 2111 & 0.0055 \\
    294 & Sharegpt4v & 102025 & 0.2675 \\
    295 & Shikra & 5814 & 0.0152 \\
    296 & Slidecaps & 34810 & 0.0913 \\
    297 & Slidevqa & 10341 & 0.0271 \\
    298 & Sroie & 500 & 0.0013 \\
    299 & Stvqa & 18921 & 0.0496 \\
    300 & Svit & 136349 & 0.3575 \\
    301 & Tabfact & 13182 & 0.0346 \\
    302 & Tech4all & 8256 & 0.0216 \\
    303 & Textocr-gpt4v & 25114 & 0.0658 \\
    304 & Tiku/math-4o-zh & 386025 & 1.0121 \\
    305 & Tiku/multidisciplinary-en & 2220078 & 5.8207 \\
    306 & Unimer & 987289 & 2.5885 \\
    307 & VCR & 95036 & 0.2492 \\
    308 & Vflan-4o-zh & 35899 & 0.0941 \\
    309 & VG-rem & 611465 & 1.6032 \\
    310 & Viquae & 1190 & 0.0031 \\
    311 & VQA-cdip-anneal & 753921 & 1.9767 \\
    312 & VQAv2 & 82783 & 0.2171 \\
    313 & WTQ & 1679 & 0.0044 \\

  \midrule
   & \textbf{Total} & \textbf{38140772} & \textbf{100.0000} \\
\end{longtable}

\subsection{Supervised Fine-tuning Data}\label{sec:sft_data}
During the initial phase of SFT data construction, our objective is to equip the model to handle a broad spectrum of application scenarios. To this end, we develop a model capability taxonomy based on the classification of traditional visual tasks and practical application requirements of vision-language models. Guided by this taxonomy, we sample from open-source multi-task training data to supplement high-quality instruction data. We use a pretrained model to classify the open-source data and extract the required categories. For data where image and question quality are high but answer quality is poor, we re-annotate the answers using a model and perform manual verification. Additionally, based on category guidelines, we generate targeted instruction data for certain categories to further enhance the model’s capabilities. Furthermore, we utilized language data produced by DeepSeek R1 to preserve the MLLM's linguistic abilities, with the ratio of multimodal to language data being approximately 1:1 at this stage.

\begin{longtable}{r l r r}
  \caption{Detailed Statistics of SFT Data} \\
  \toprule
  \textbf{No.} & \textbf{Dataset Name} & \textbf{Number} & \textbf{Percentage (\%)} \\
  \midrule
  \endfirsthead
  
  \caption*{Detailed Statistics of SFT Data (Continued).} \\
  \toprule
  \textbf{No.} & \textbf{Dataset Name} & \textbf{Number} & \textbf{Percentage (\%)} \\
  \midrule
  \endhead
  
  \bottomrule
  \endlastfoot

  1 & Competition Mathematics & 150000 & 2.4269 \\
  2 & Code & 50000 & 0.8089 \\
  3 & Math & 200000 & 3.2359 \\
  4 & General QA & 300000 & 4.8539 \\
  5 & Tulu Instruction & 19118 & 0.3093 \\
  6 & Multi-turn Conversation & 61114 & 0.9888 \\
  7 & MegaScience & 1250000 & 20.2244 \\
  8 & OpenHermes2-5 & 200000 & 3.2359 \\
  9 & QwenCaps/PixMo/cap & 712401 & 11.5263 \\
  10 & QwenCaps/PixMo/cap-zh & 10000 & 0.1618 \\
  11 & QwenCaps/Allava-Laion & 10000 & 0.1618 \\
  12 & QwenCaps/COCO & 10000 & 0.1618 \\
  13 & QwenCaps/Flickr30k & 10000 & 0.1618 \\
  14 & QwenCaps/Laion-GPT4V & 10000 & 0.1618 \\
  15 & QwenCaps/LLaVA & 10000 & 0.1618 \\
  16 & QwenCaps/LLaVaR-S1 & 10000 & 0.1618 \\
  17 & QwenCaps/OCR-VQA & 10000 & 0.1618 \\
  18 & QwenCaps/SAM & 10000 & 0.1618 \\
  19 & QwenCaps/VG & 10000 & 0.1618 \\
  20 & QwenCaps/Vision-FLAN & 10000 & 0.1618 \\
  21 & DCC & 10000 & 0.1618 \\
  22 & NoahCaps-EN & 10000 & 0.1618 \\
  23 & NoahCaps-ZH & 10000 & 0.1618 \\
  24 & ShareGPT4O-Image & 10000 & 0.1618 \\
  25 & ShareGPT4O-Video & 10000 & 0.1618 \\
  26 & Tech4All & 10000 & 0.1618 \\
  27 & OneVision/CLEVR-Math-MathV360K & 10000 & 0.1618 \\
  28 & OneVision/Super-CLEVR-MathV360K & 10000 & 0.1618 \\
  29 & PixMo/Ask-Model-Anything & 161737 & 2.6168 \\
  30 & PixMo/Ask-Model-Anything-ZH & 10000 & 0.1618 \\
  31 & PixMo/Count & 10000 & 0.1618 \\
  32 & QwenQA/Cauldron/COCOQA & 27022 & 0.4372 \\
  33 & QwenQA/Cauldron/TallyQA & 83671 & 1.3538 \\
  34 & QwenQA/Cauldron/Visual7W & 13006 & 0.2104 \\
  35 & QwenQA/Cauldron/VSR & 1650 & 0.0267 \\
  36 & QwenQA/PixMo/Cap-QA & 247256 & 3.9990 \\
  37 & QwenQA/PixMo/Cap-QA-ZH & 10000 & 0.1618 \\
  38 & QwenQA/Allava-Laion & 484257 & 7.8350 \\
  39 & QwenQA/Allava-VFLAN & 179020 & 2.8965 \\
  40 & QwenQA/AOKVQA & 39358 & 0.6368 \\
  41 & QwenQA/GQA & 69198 & 1.1196 \\
  42 & QwenQA/OKVQA & 3442 & 0.0557 \\
  43 & QwenQA/VQAv2 & 78718 & 1.2736 \\
  44 & QwenQA/Raw/AOKVQA & 10000 & 0.1618 \\
  45 & QwenQA/Raw/GQA & 10000 & 0.1618 \\
  46 & QwenQA/Raw/OKVQA & 10000 & 0.1618 \\
  47 & QwenQA/Raw/VQAv2 & 10000 & 0.1618 \\
  48 & VFLAN-4O-ZH & 10000 & 0.1618 \\
  49 & Cauldron/IconQA & 10000 & 0.1618 \\
  50 & Cauldron/Mimic-CGD & 10000 & 0.1618 \\
  51 & Cauldron/NLVR2 & 10000 & 0.1618 \\
  52 & Cauldron/Raven & 10000 & 0.1618 \\
  53 & Cauldron/Spot-the-Diff & 10000 & 0.1618 \\
  54 & PixMo/Points & 10000 & 0.1618 \\
  55 & COCO-Rec & 130723 & 2.1150 \\
  56 & COCO-Reg & 130723 & 2.1150 \\
  57 & Flickr & 10000 & 0.1618 \\
  58 & Objects365 & 10000 & 0.1618 \\
  59 & Cauldron/IAM & 10000 & 0.1618 \\
  60 & CoSyn-400K/Chart & 10000 & 0.1618 \\
  61 & CoSyn-400K/Chemical & 10000 & 0.1618 \\
  62 & CoSyn-400K/Circuit & 10000 & 0.1618 \\
  63 & CoSyn-400K/Diagram & 10000 & 0.1618 \\
  64 & CoSyn-400K/Document & 10000 & 0.1618 \\
  65 & CoSyn-400K/Graphic & 10000 & 0.1618 \\
  66 & CoSyn-400K/Music & 10000 & 0.1618 \\
  67 & CoSyn-400K/Nutrition & 10000 & 0.1618 \\
  68 & CoSyn-400K/Table & 10000 & 0.1618 \\
  69 & OneVision/IIIT5K & 10000 & 0.1618 \\
  70 & OneVision/K12-Printing & 10000 & 0.1618 \\
  71 & OneVision/LRV-Chart & 10000 & 0.1618 \\
  72 & OneVision/Orand-Car-A & 10000 & 0.1618 \\
  73 & QwenCaps/ChartQA & 18317 & 0.2964 \\
  74 & QwenCaps/Chart2Text & 16605 & 0.2687 \\
  75 & QwenCaps/Chart-to-Text & 44096 & 0.7135 \\
  76 & QwenCaps/DocVQA & 11166 & 0.1807 \\
  77 & QwenCaps/FUNSD & 149 & 0.0024 \\
  78 & QwenCaps/InfoVQA & 4890 & 0.0791 \\
  79 & QwenCaps/OpenCQA & 7724 & 0.1250 \\
  80 & QwenCaps/POIE & 2999 & 0.0485 \\
  81 & QwenCaps/SROIE & 608 & 0.0098 \\
  82 & QwenCaps/TextVQA & 24990 & 0.4043 \\
  83 & QwenCaps/Vistext & 8816 & 0.1426 \\
  84 & QwenCaps/XFUND & 1043 & 0.0169 \\
  85 & QwenQA/Cauldron/Diagram-Image-to-Text & 297 & 0.0048 \\
  86 & QwenQA/Cauldron/Hateful-Memes & 3729 & 0.0603 \\
  87 & QwenQA/Cauldron/TabMWP & 22410 & 0.3626 \\
  88 & QwenQA/Cauldron/TextVQA & 16059 & 0.2598 \\
  89 & QwenQA/ChartQA/Human & 3290 & 0.0532 \\
  90 & QwenQA/ChartQA/Augmented & 12469 & 0.2017 \\
  91 & QwenQA/DocVQA/Human & 8924 & 0.1444 \\
  92 & QwenQA/InfoVQA/Human & 4092 & 0.0662 \\
  93 & QwenQA/STVQA & 14381 & 0.2327 \\
  94 & QwenQA/TabFact & 12648 & 0.2046 \\
  95 & QwenQA/WTQ & 1588 & 0.0257 \\
  96 & QwenQA/Raw/STVQA & 10000 & 0.1618 \\
  97 & QwenQA/Raw/TabFact & 10000 & 0.1618 \\
  98 & QwenQA/Raw/WTQ & 10000 & 0.1618 \\
  99 & XFUND/ZH-QA & 10000 & 0.1618 \\
  100 & CIPC-DocVQA & 10000 & 0.1618 \\
  101 & CoAVQA-Chat & 10000 & 0.1618 \\
  102 & CoAVQA-Bench & 10000 & 0.1618 \\
  103 & E2E-OCR-0324 & 359996 & 5.8245 \\
  104 & ESTVQA & 10000 & 0.1618 \\
  105 & ESTVQA-Grounding & 10000 & 0.1618 \\
  106 & FUNSD-QA & 10000 & 0.1618 \\
  107 & POIE & 10000 & 0.1618 \\
  108 & SlideCaps & 10000 & 0.1618 \\
  109 & SlideVQA & 10000 & 0.1618 \\
  110 & SROIE & 10000 & 0.1618 \\
  111 & VQA-CDIP-SFT & 10000 & 0.1618 \\
  112 & QwenCaps/Screen2Words & 10000 & 0.1618 \\
  113 & Cauldron/VisualMRC & 10000 & 0.1618 \\
  114 & Cauldron/WebSight & 10000 & 0.1618 \\
  115 & CoSyn-Point & 10000 & 0.1618 \\
  116 & Screen/Nav & 10000 & 0.1618 \\
  117 & Screen/Ref & 10000 & 0.1618 \\
  118 & Screen/SQA & 10000 & 0.1618 \\
  119 & Cauldron/GeomVerse & 10000 & 0.1618 \\
  120 & CoSyn-400K/Math & 66707 & 1.0793 \\
  121 & OneVision/Geometry3K-MathV360K & 10000 & 0.1618 \\
  122 & OneVision/GeoQA-Plus-MathV360K & 10000 & 0.1618 \\
  123 & OneVision/Geos-MathV360K & 10000 & 0.1618 \\
  124 & OneVision/IconQA-MathV360K & 10000 & 0.1618 \\
  125 & OneVision/PMC-VQA-MathV360K & 10000 & 0.1618 \\
  126 & OneVision/TabMWP-MathV360K & 10000 & 0.1618 \\
  127 & OneVision/Unigeo-MathV360K & 10000 & 0.1618 \\
  128 & OneVision/Geo3K & 10000 & 0.1618 \\
  129 & OneVision/Mavis-Math-Metagen & 10000 & 0.1618 \\
  130 & OneVision/Mavis-Math-Rule-Geo & 10000 & 0.1618 \\
  131 & CROHME & 10000 & 0.1618 \\
  132 & Geo170K-Alignment & 10000 & 0.1618 \\
  133 & Geo170K-QA-Tuning & 10000 & 0.1618 \\
  134 & HME & 10000 & 0.1618 \\
  135 & M2E & 10000 & 0.1618 \\
  136 & MLHME & 10000 & 0.1618 \\
  137 & Unimer & 10000 & 0.1618 \\
  138 & QwenCaps/AI2D & 9778 & 0.1582 \\
  139 & QwenCaps/ScienceQA & 4985 & 0.0807 \\
  140 & QwenQA/Cauldron/MapQA & 22950 & 0.3713 \\
  141 & QwenQA/Cauldron/TQA & 861 & 0.0139 \\
  142 & QwenQA/AI2D & 10593 & 0.1714 \\
  143 & QwenQA/ScienceQA & 11071 & 0.1791 \\
  \midrule
    & Total & \textbf{6180645} & \textbf{100.0000} \\
\end{longtable}

\section{Experiments}
\subsection{Results of Test-Time Resolution Search}\label{sec:Test-Time Resolution Search}
In this Section, we present detail evaluation results with test-time image resolution search strategy. 
On MME dataset, as shown in Figure~\ref{fig:MME-img-search}, models trained with different resolution lengths yield consistent evaluation results when min-pixels is less than 16. In addition, down-sampling images on MME results in a significant improvement in model performance.
On OCRBench dataset, as shown in Figure~\ref{fig:OCRBench-img-search}, when min-pixels is set to 16 or 32, the model performs better than when it is set to 4.
The reason is that too small input images are out of distribution in OCR task for our model.
On the other hand, when min-pixels is set to 64, the model performance degrades. The reason is that the image blurring effect caused by up-sampling images has a negative impact on the model’s visual perception.
On DocVQA and InfoVQA datasets, as shown in Figure~\ref{fig:DocVQA-img-search} and Figure~\ref{fig:InfoVQA-img-search}, min-pixels variable has no detectable impact on the evaluation results since most of the images in these two datasets have a resolution higher than min-pixels.
Furthermore, the optimal inference max-pixels on these two datasets falls within the range of 2560–3072.

\begin{figure}[h] 
    \centering
        \subfigure{
        \includegraphics[width=0.49\textwidth]{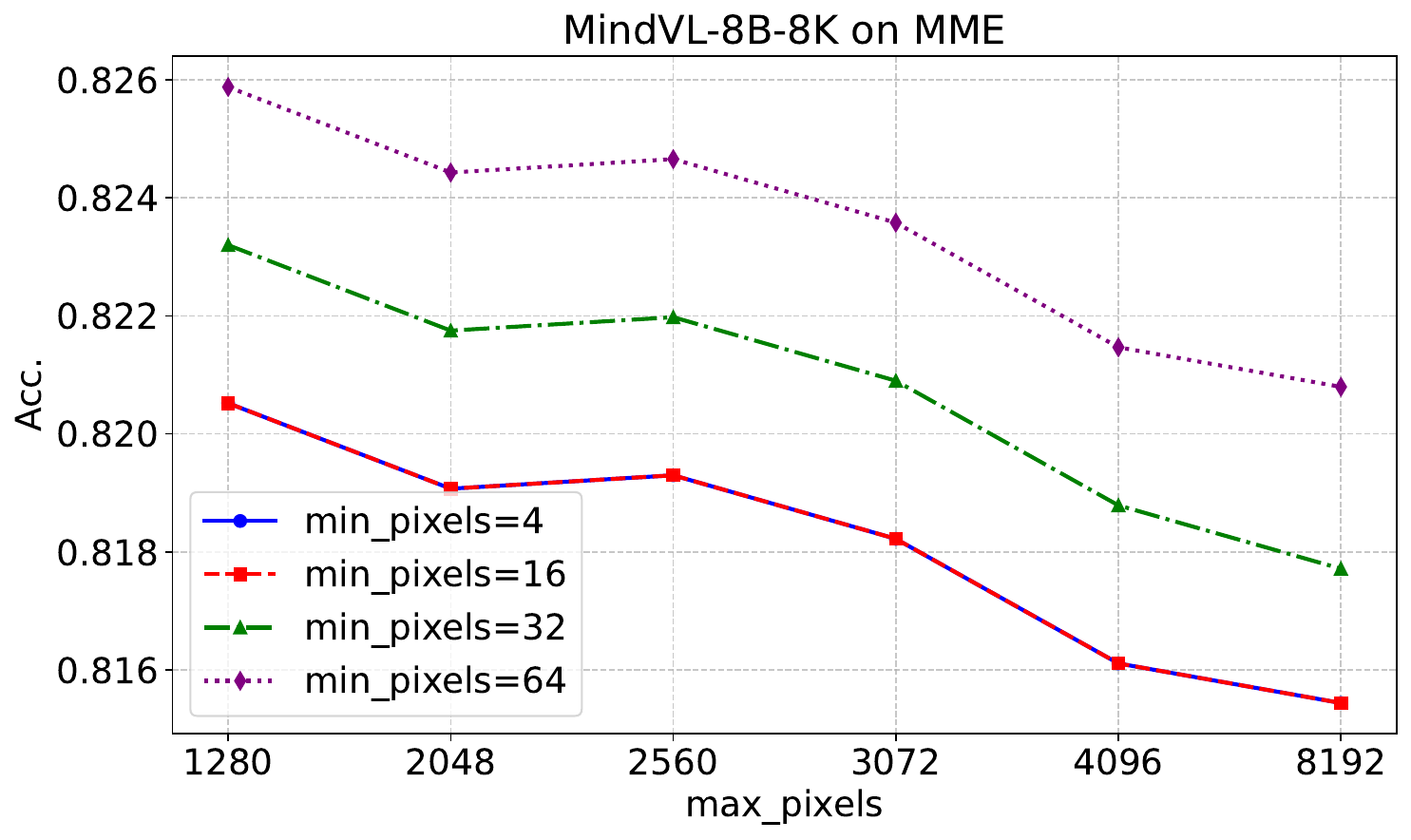}

        \includegraphics[width=0.49\textwidth]{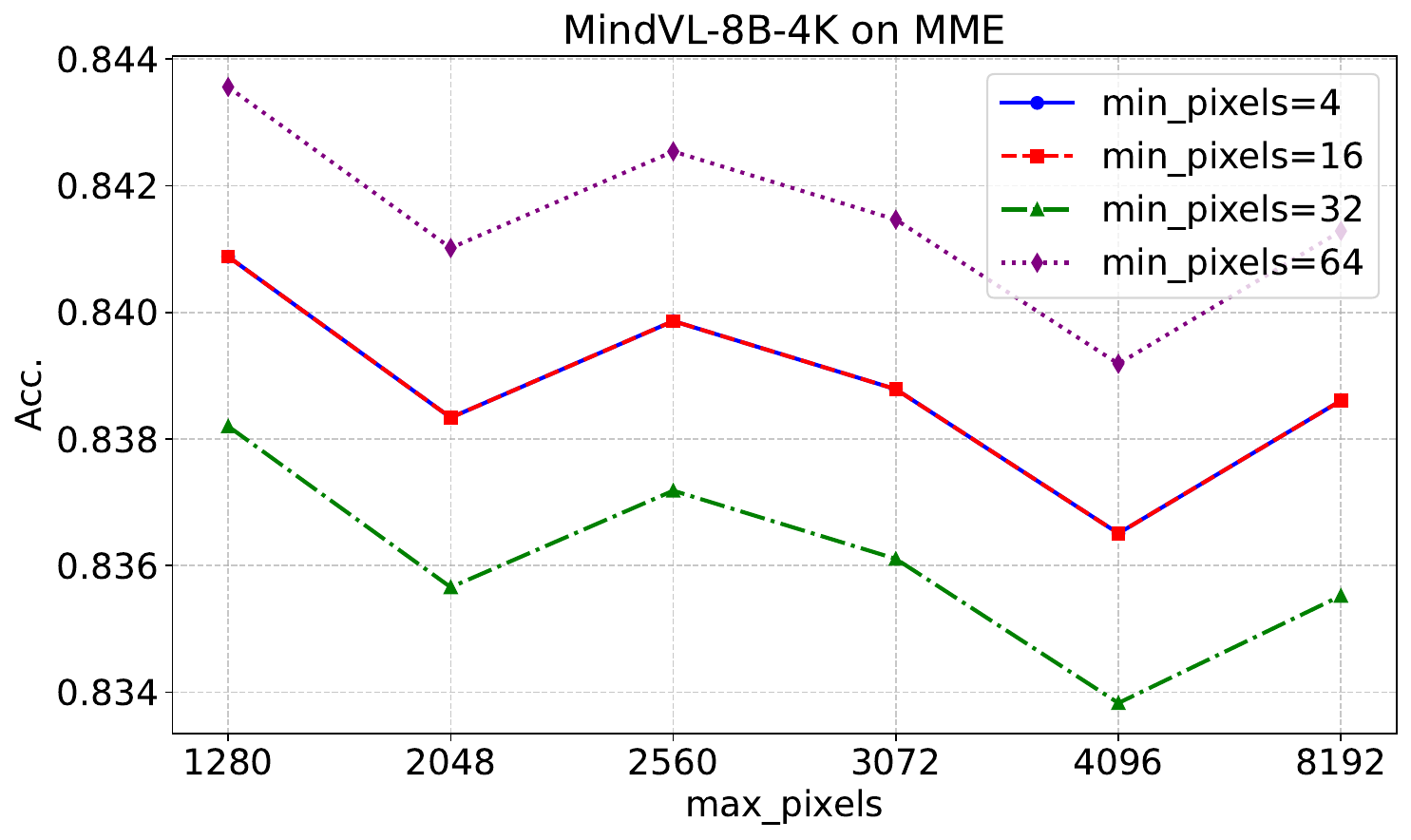}}

        \subfigure{
        \includegraphics[width=0.49\textwidth]{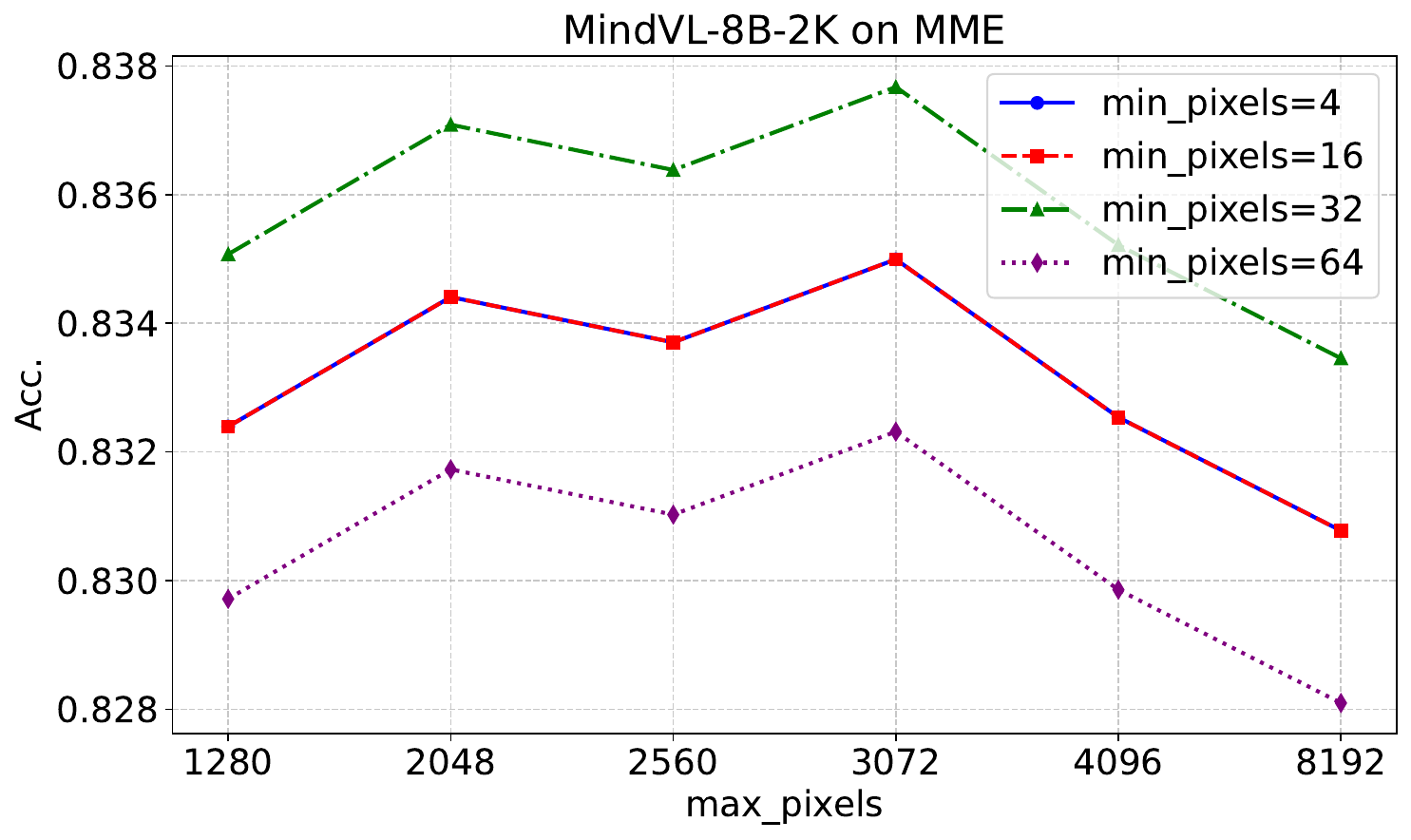}

        \includegraphics[width=0.49\textwidth]{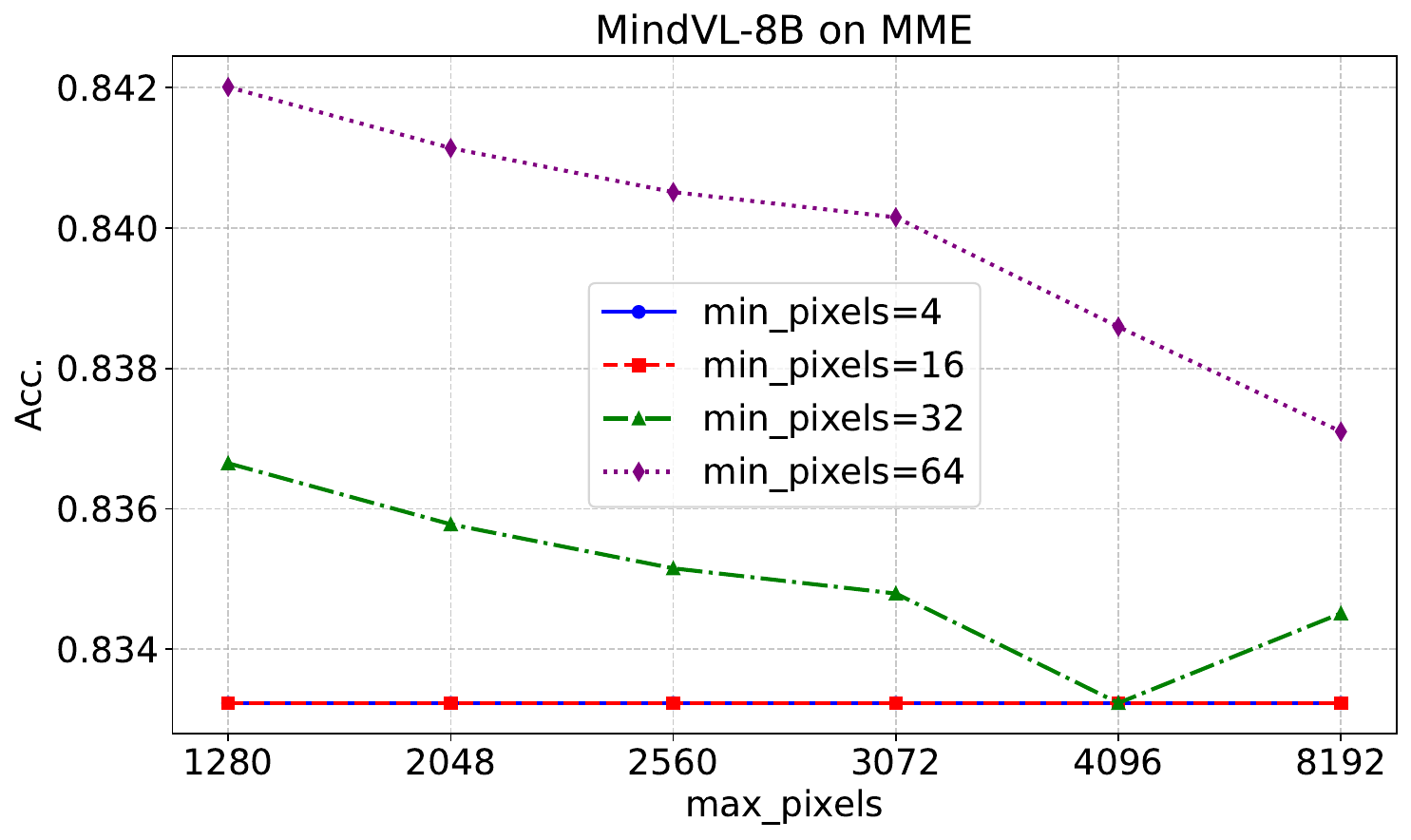}}
        \caption{Evaluation results on MME of MindVL with different input image resolutions.}
  \label{fig:MME-img-search}%
\end{figure}

\begin{figure}[h] 
    \centering
        \subfigure{
        \includegraphics[width=0.49\textwidth]{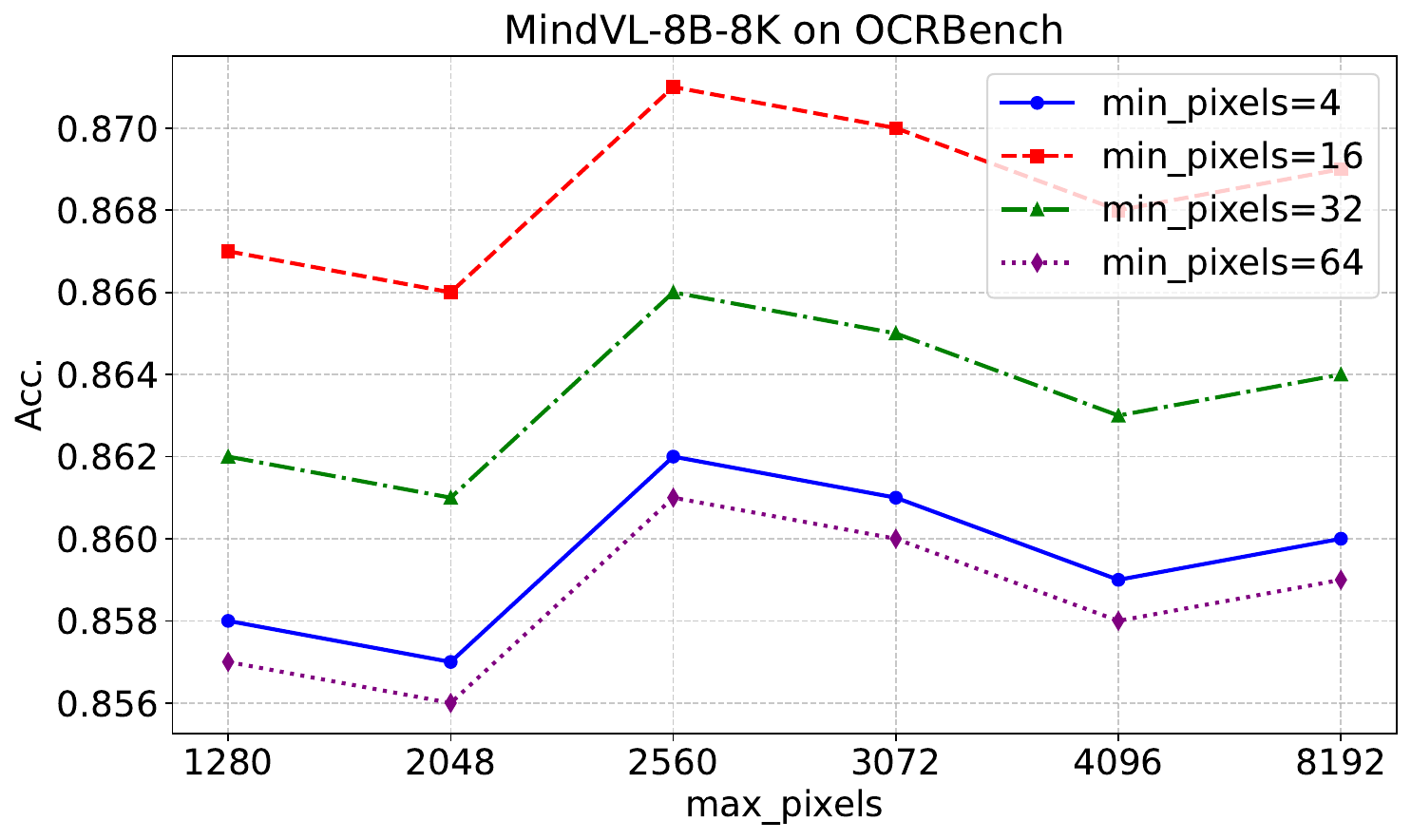}

        \includegraphics[width=0.49\textwidth]{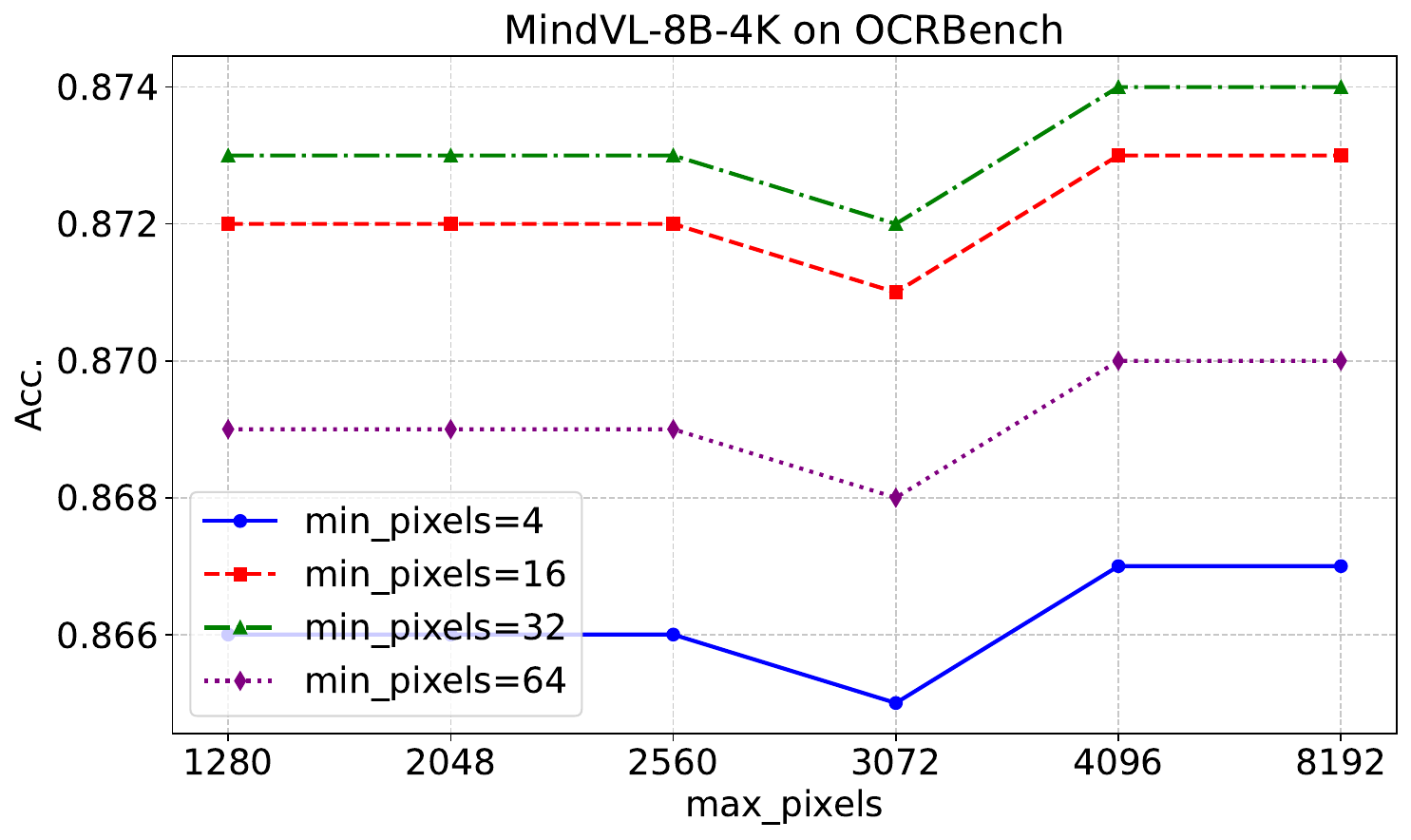}}

        \subfigure{
        \includegraphics[width=0.49\textwidth]{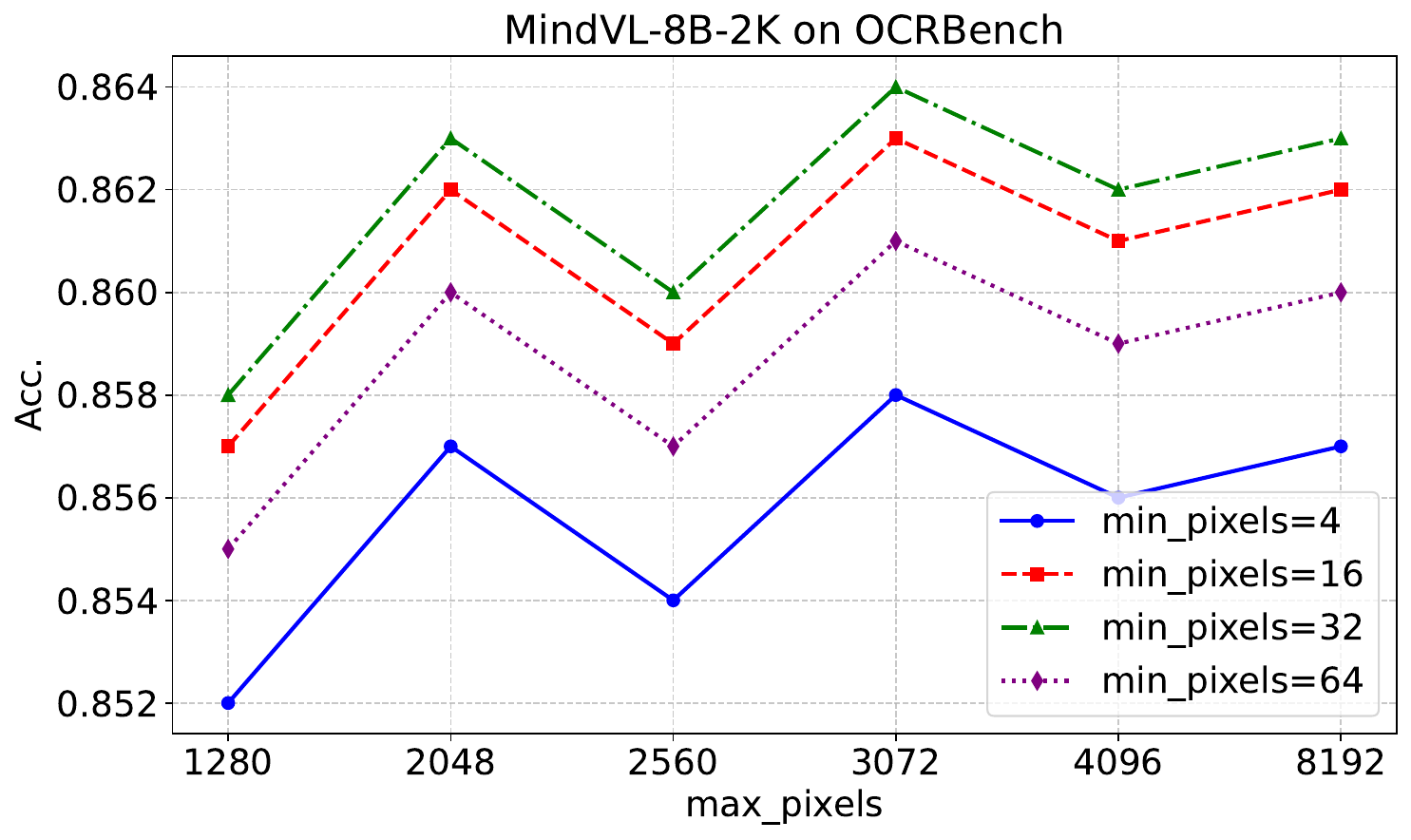}

        \includegraphics[width=0.49\textwidth]{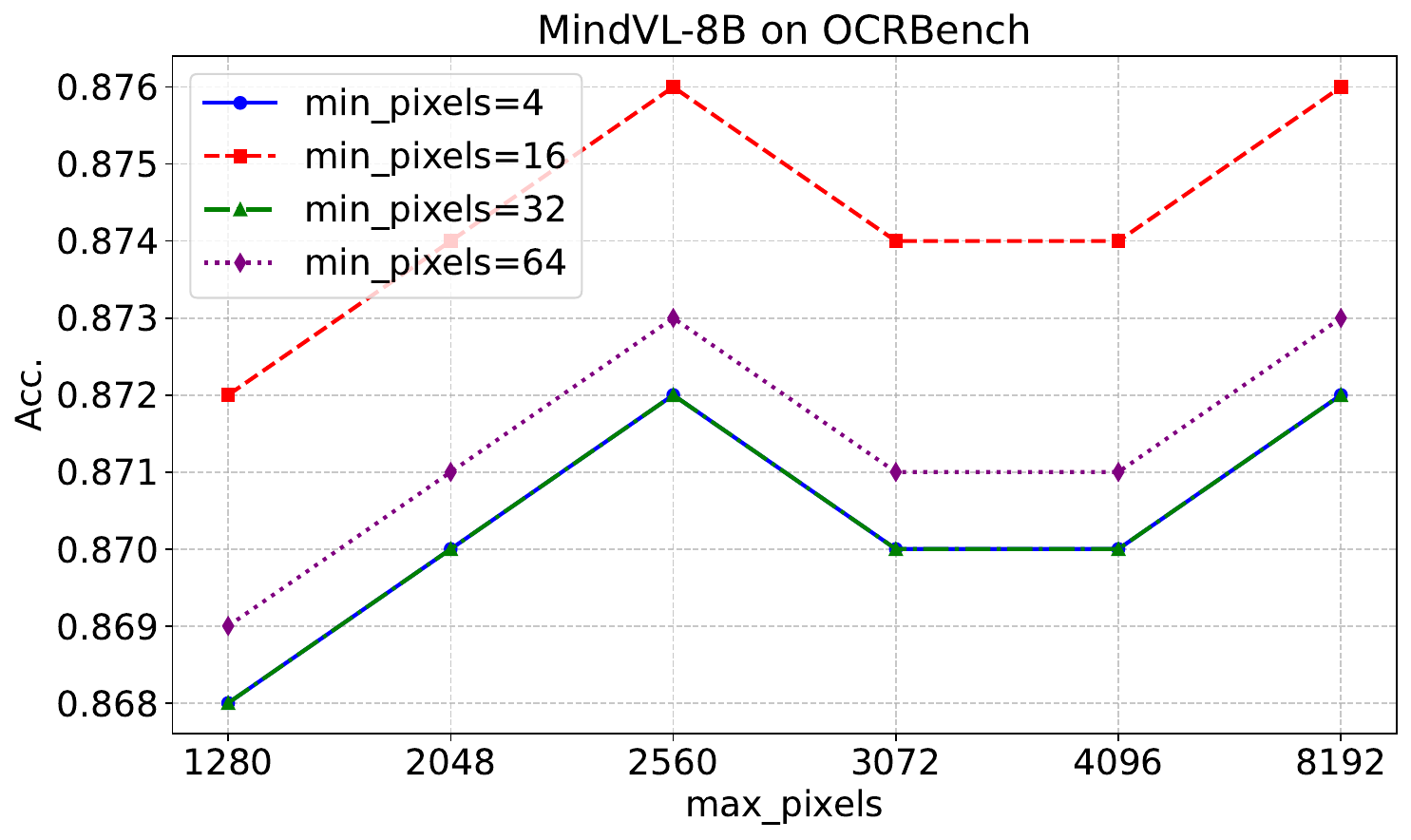}}
        \caption{Evaluation results on OCRBench of MindVL with different input image resolutions.}
  \label{fig:OCRBench-img-search}%

\end{figure}

\begin{figure}[h] 
    \centering
        \subfigure{
        \includegraphics[width=0.49\textwidth]{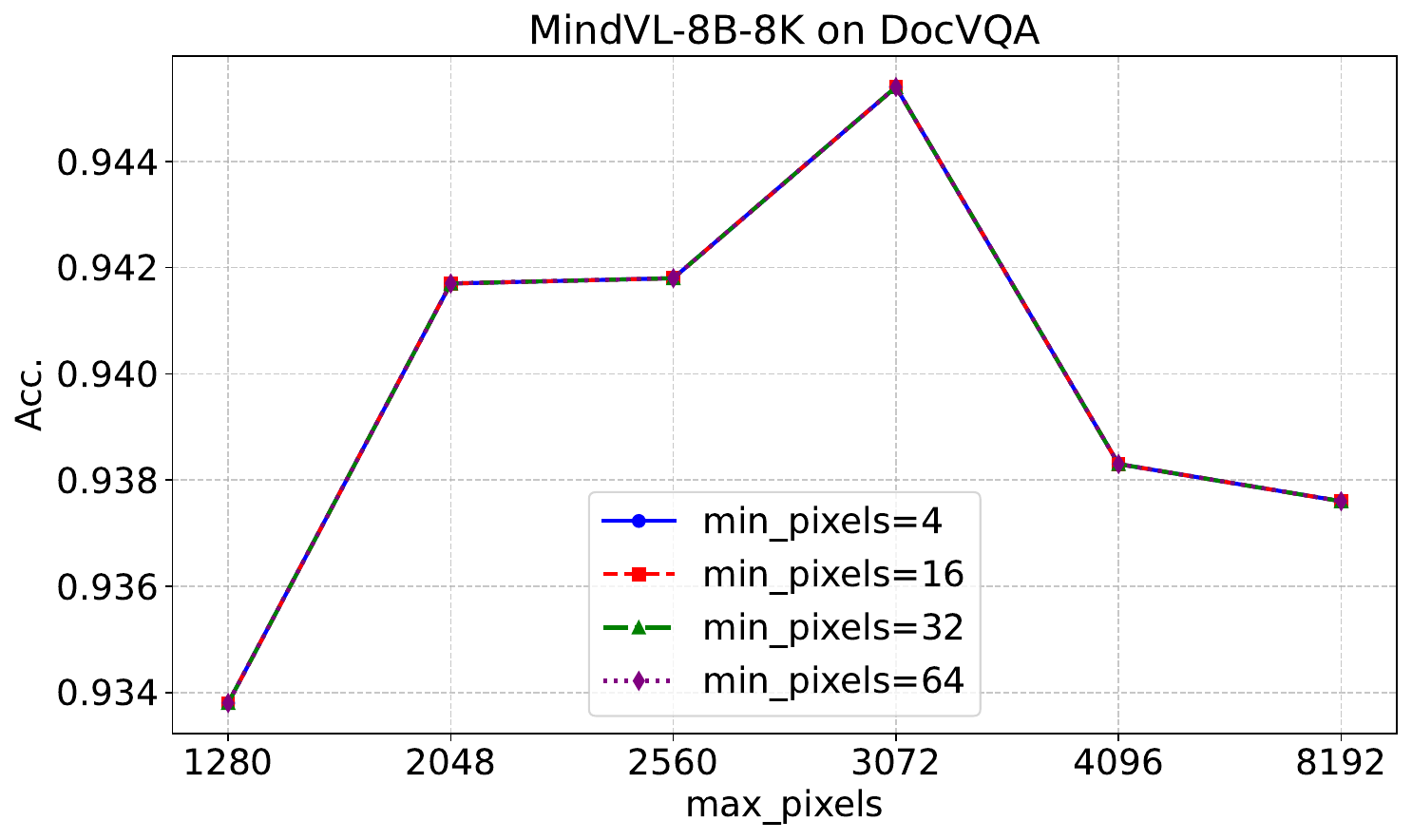}

        \includegraphics[width=0.49\textwidth]{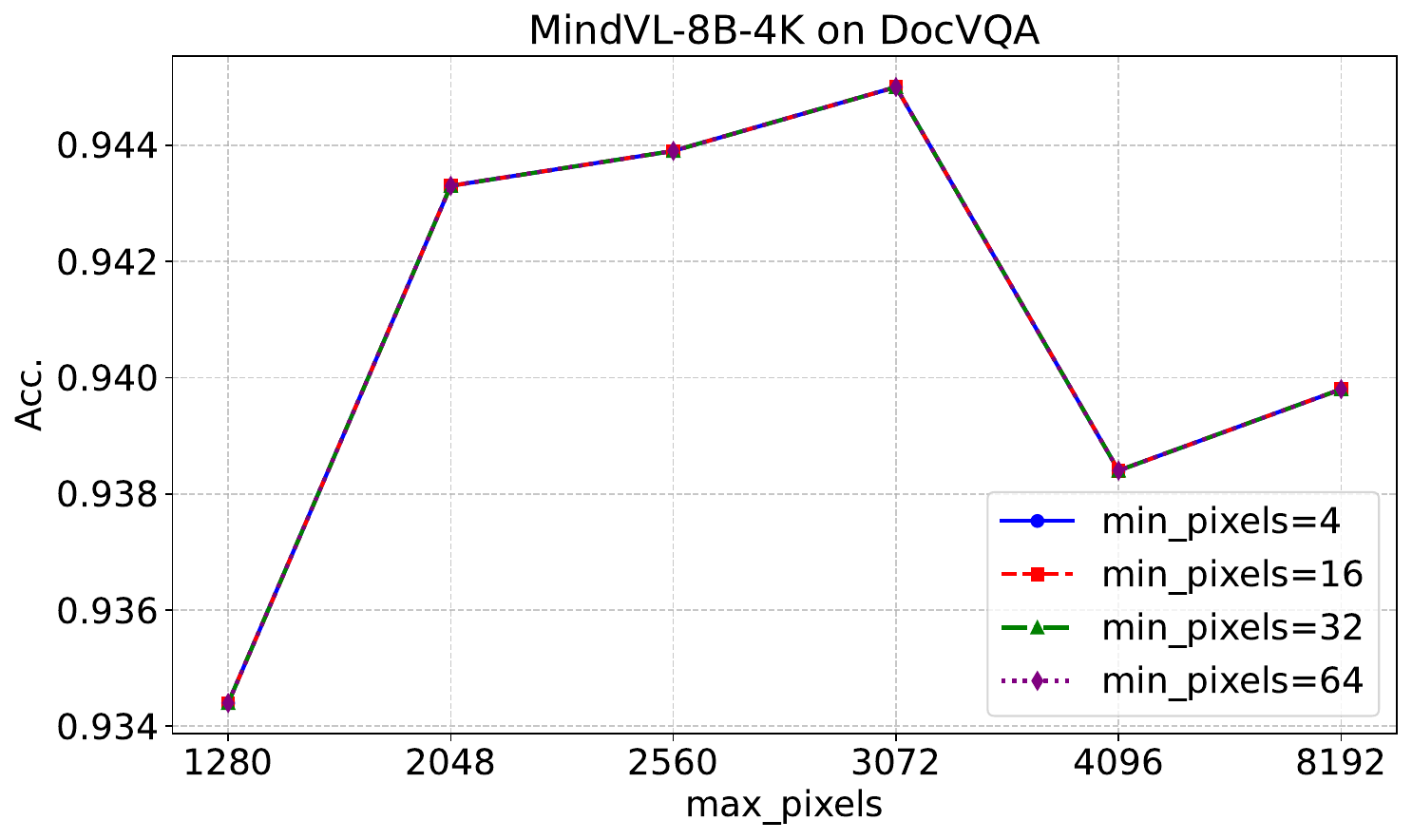}}

        \subfigure{
        \includegraphics[width=0.49\textwidth]{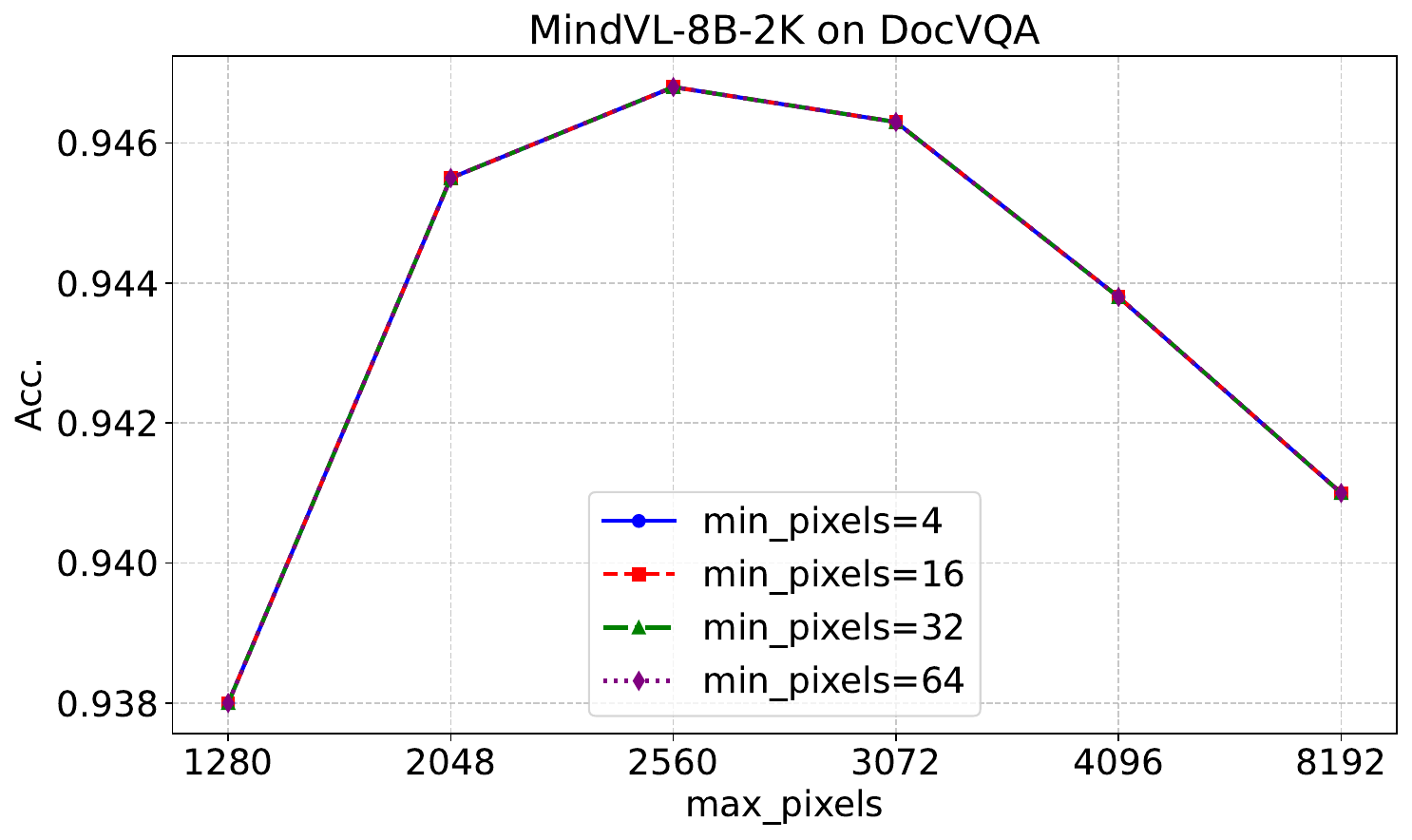}

        \includegraphics[width=0.49\textwidth]{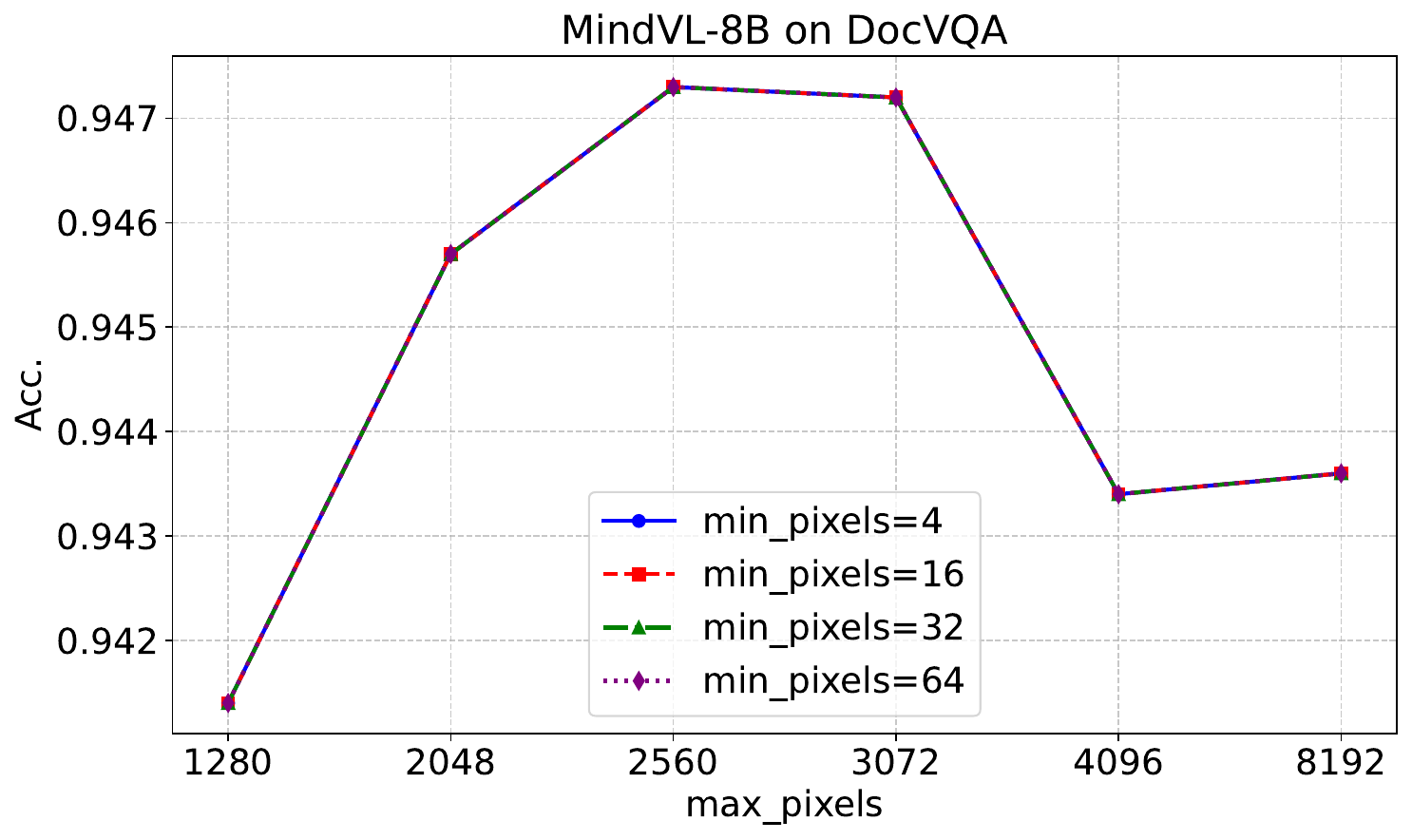}}
        \caption{Evaluation results on DocVQA of MindVL with different input image resolutions.}
  \label{fig:DocVQA-img-search}%

\end{figure}

\begin{figure}[h] 
    \centering
        \subfigure{
        \includegraphics[width=0.49\textwidth]{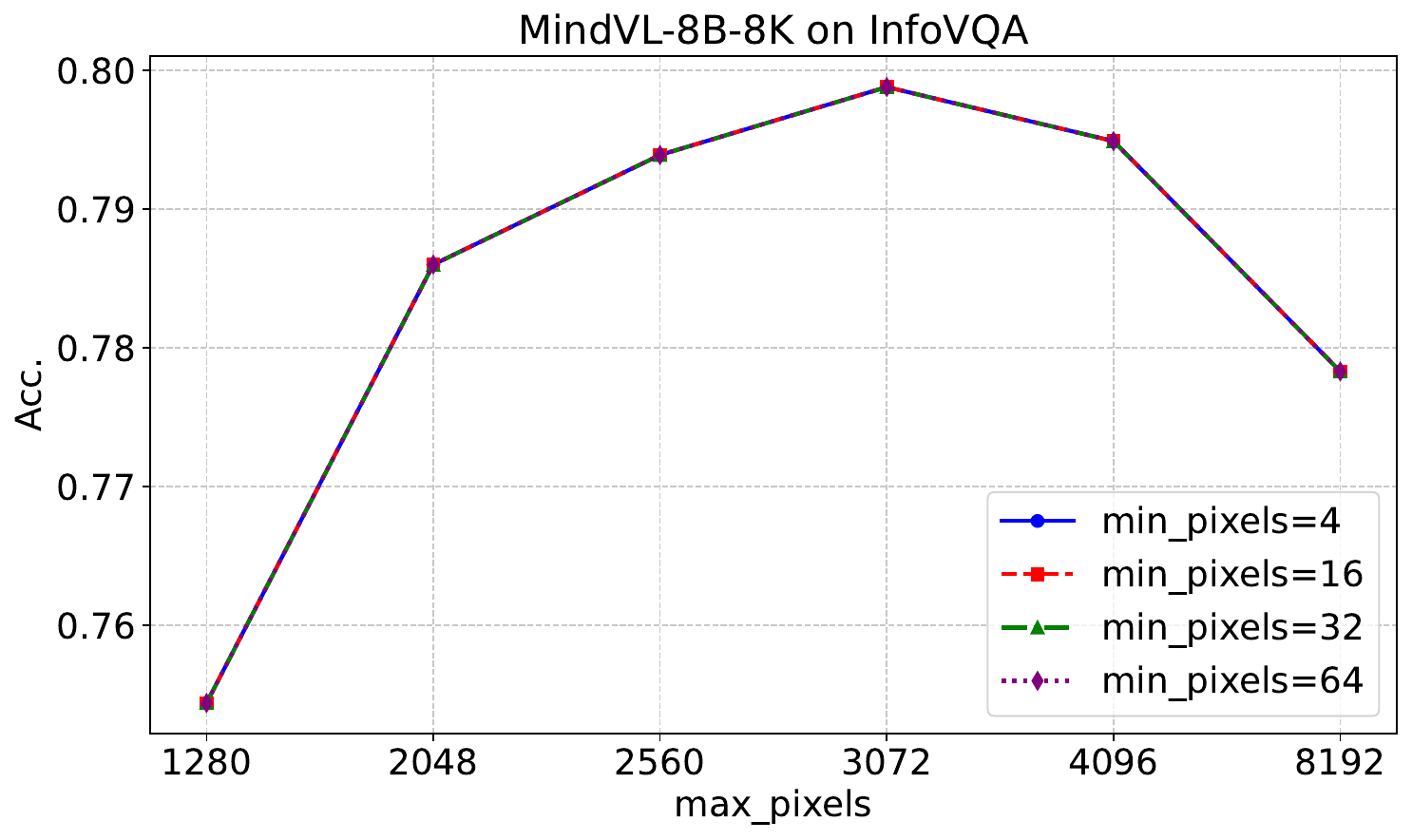}

        \includegraphics[width=0.49\textwidth]{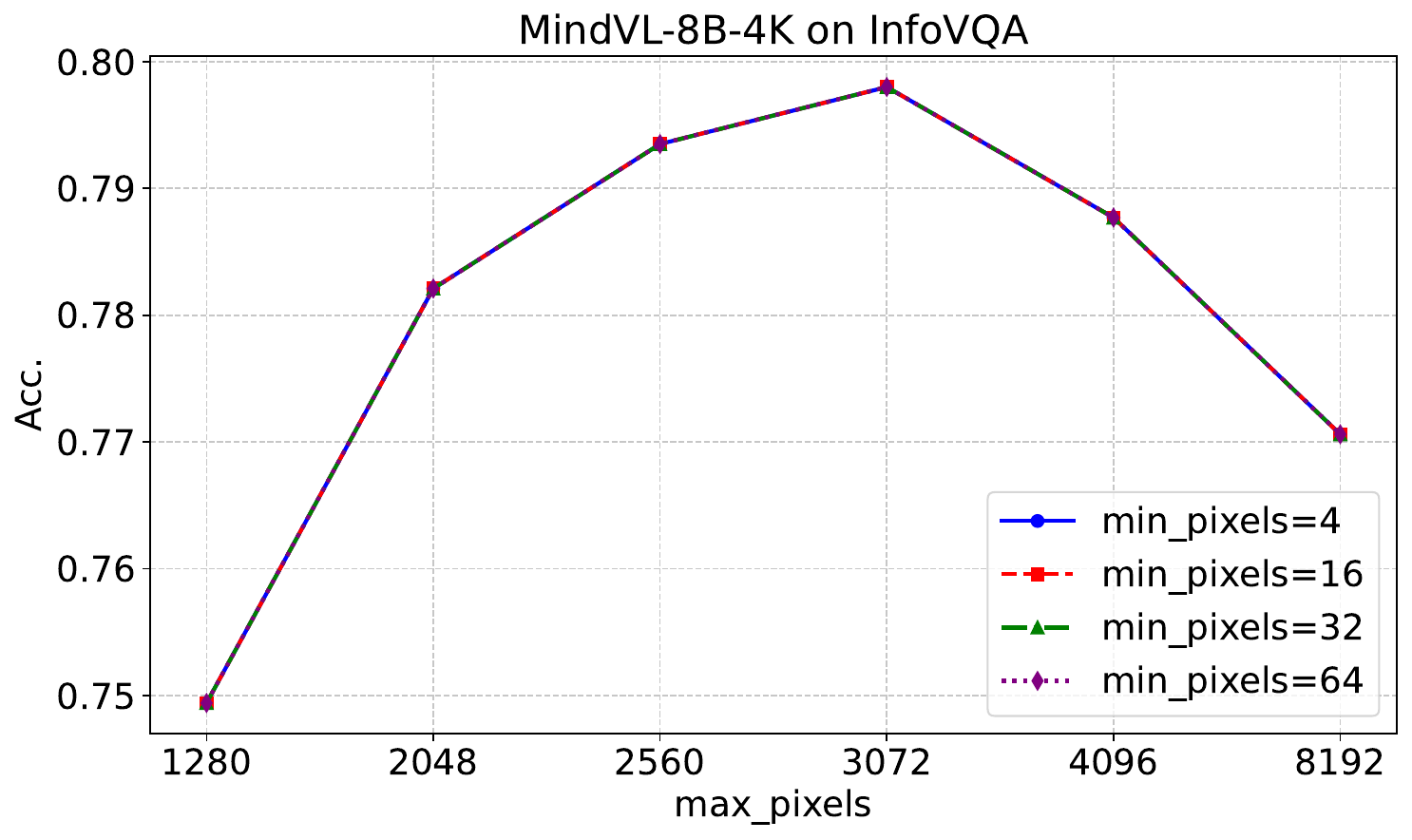}}

        \subfigure{
        \includegraphics[width=0.49\textwidth]{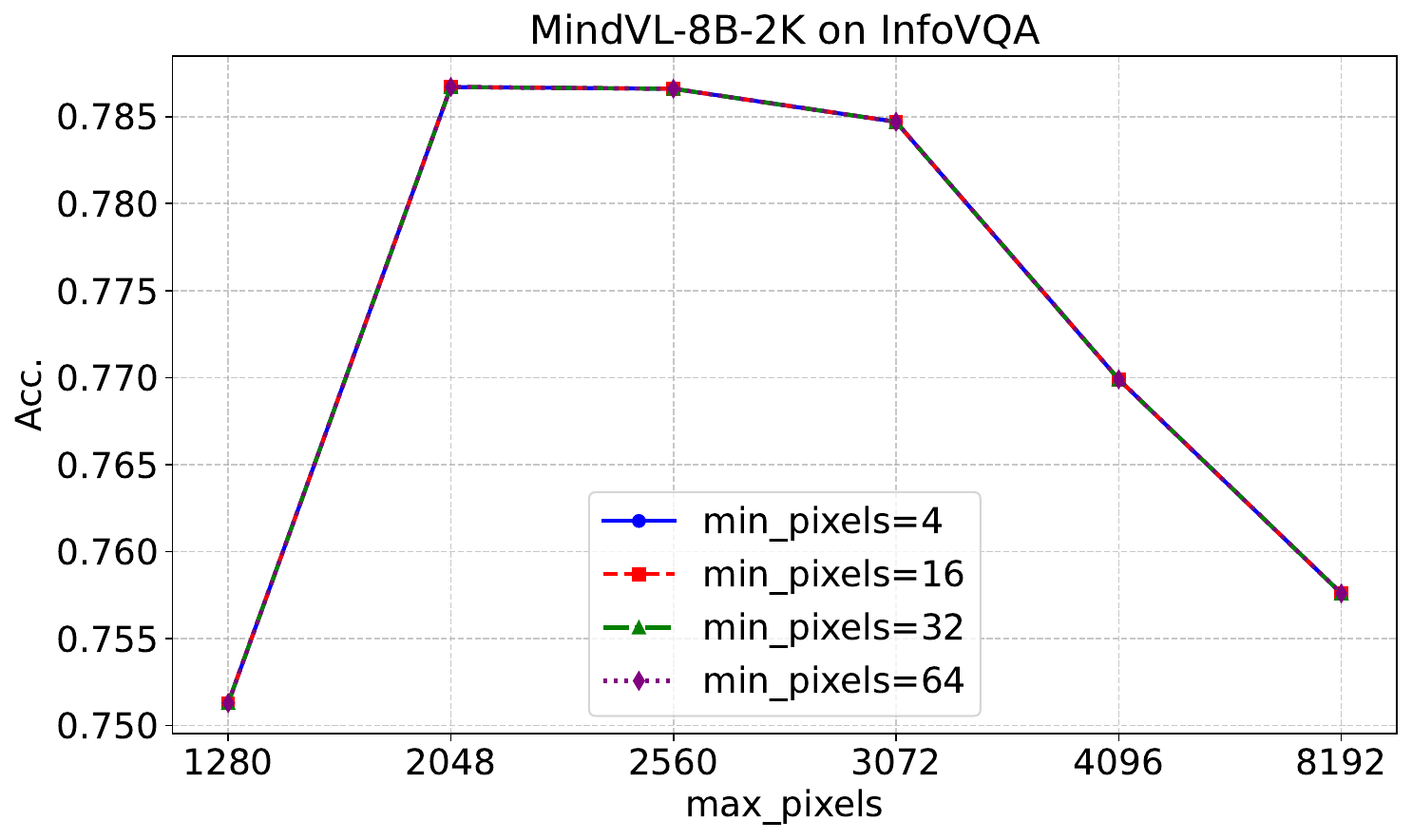}

        \includegraphics[width=0.49\textwidth]{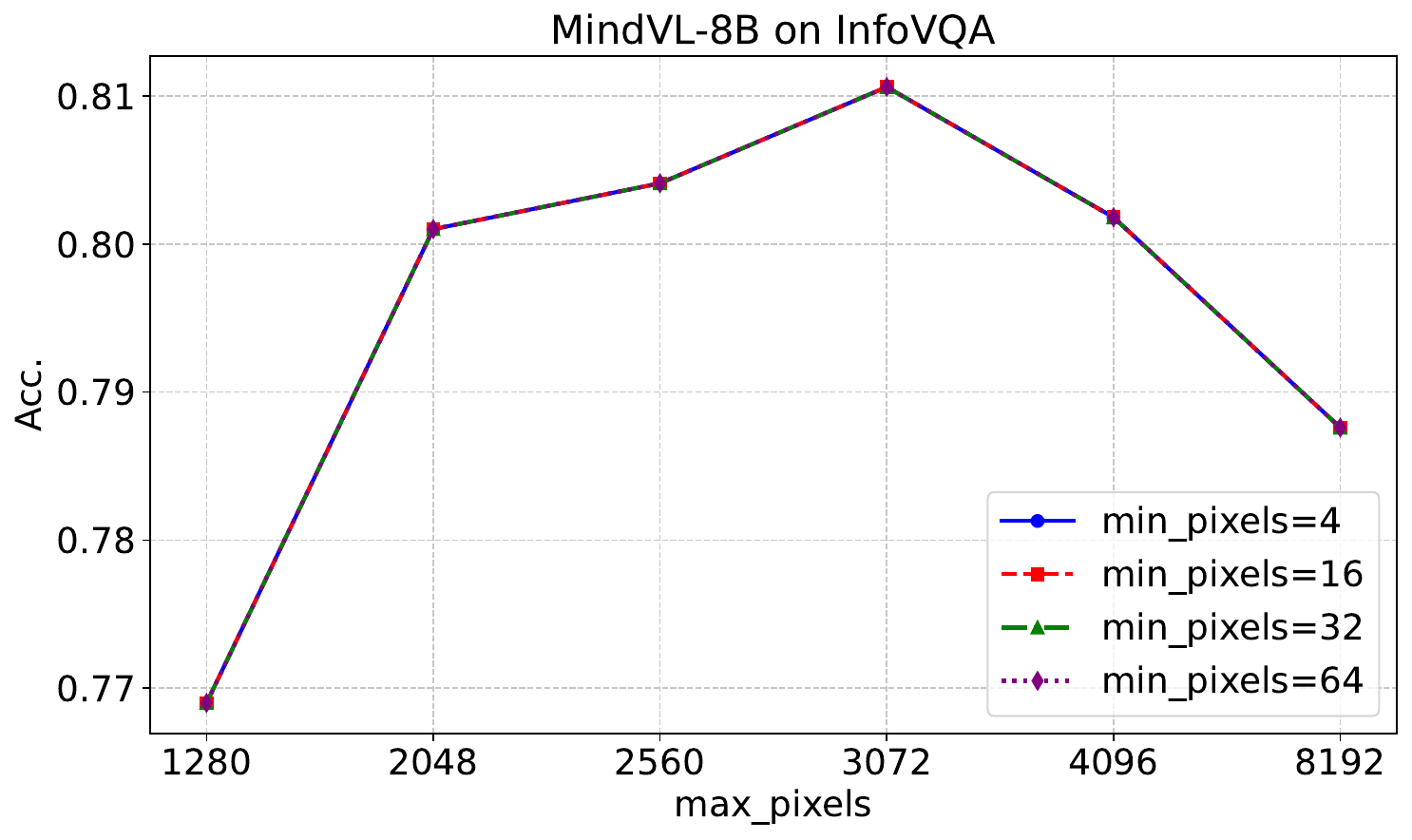}}
        \caption{Evaluation results on InfoVQA of MindVL with different input image resolutions.}
  \label{fig:InfoVQA-img-search}%

\end{figure}

\subsection{Integration of Multitask Data}
We also present the experimental results of the model trained with different data recipes. As shown in Table~\ref{tab:ablation_study_of_multitask_data},
MindVL-8B-V1Data model is trained with introduced data of Video, Captions, OCR, GUI, Math, Knowledge, Grounding and Genreal VQA introduced in Section~\ref{sec:multitask_training_data}. For MindVL-8B-V2Data model, Internleaved image-text data are added for training.
For MindVL-8B-V3Data model, STEM data and additional Captions, OCR, GUI, Grounding and Genreal VQA data are added for training.
By comparing the results of MindVL-8B-V1Data and MindVL-8B-V2Data, we can observe that adding Internleaved image-text data for training can significantly improve performance over all benchmarks.
It is because that interleaved image-text data not only contains complex logical relationships between text and images, but also covers a wide range of domain knowledge—both of which are highly beneficial for enhancing the multimodal model's cognitive capabilities.
Moreover, by adding more refined and diverse Captions, OCR, GUI, Grounding and Genreal VQA data to train the model MindVL-8B-V3Data, the performance on multiple benchmark datasets has been further improved, such as MME, MMBench and InfoVQA.

\begin{table}[htbp]
  \centering
  \setlength{\tabcolsep}{4pt}
  \footnotesize
  \caption{Ablation study results of multitask data.}
    \renewcommand{\arraystretch}{1.2} 
    \begin{tabular}{l|cccccc|c}
    \toprule
    \textbf{Model} & \textbf{MME} & \textbf{MMBench} & \textbf{OCRBench} & \textbf{DocVQA} & \textbf{ChartQA} & \textbf{InfoVQA} & \textbf{Overall} \\
    \midrule
   MindVL-8B-V1Data & 79.8 & 81.5 & 83.3 & 92.1 & 83.7 & 73.2 & 82.3 \\
   MindVL-8B-V2Data & 82.4 & 83.4 & 88.2 & 94.7 & 87.2 & 79.9 & 86.0 \\
   MindVL-8B-V3Data & 84.1 & 84.3 & 87.6 & 94.7 & 87.2 & 81.1 & 86.5 \\
    \bottomrule
    \end{tabular}%
  \label{tab:ablation_study_of_multitask_data}%
\end{table}%

\subsection{Ablation Study on Training Phases}


\begin{table}[htbp]
  \centering
  \footnotesize
  \setlength{\tabcolsep}{4pt}
  \caption{Ablation study results on training phases.}
    \renewcommand{\arraystretch}{1.2} 
    \begin{tabular}{l|cccccc|c}
    \toprule
    \textbf{Model} & \textbf{MME} & \textbf{MMBench} & \textbf{OCRBench} & \textbf{DocVQA} & \textbf{ChartQA} & \textbf{InfoVQA} & \textbf{Overall} \\
    \midrule
   MindVL-8B-4Phase & 80.8 & 82.3 & 83.6 & 92.7 & 86.4 & 76.9 & 83.8 \\
   MindVL-8B-2Phase* & 79.4 & 81.8 & 84.9 & 93.5 & 85.2 & 75.8 & 83.4 \\
   MindVL-8B-2Phase & 81.6 & 82.4 & 87.0 & 94.2 & 86.8 & 79.6 & 85.3 \\
   MindVL-8B-3Phase & 84.1 & 84.3 & 87.6 & 94.7 & 87.2 & 81.1 & 86.5 \\
    \bottomrule
    \end{tabular}%
  \label{tab:training_phases}%
\end{table}%

As mentioned in Section~\ref{sec:train_recipe}, MindVL-8B undergoes a three-phase training pipeline: warm-up, multi-task learning, and SFT.
We denote the model trained on such recipe as MindVL-8B-3Phase.
Table~\ref{tab:training_phases} shows the results of ablation study on training phases.
MindVL-8B-4Phase model is trained on a four-phase training pipeline: warm-up, pretrain, multi-task learning, and SFT.
In the pretrain phase, both ViT and MLP adaptor are trained.
Our ultimate training recipe has only three-phase: warm-up, multi-task learning and SFT, achieves best performance by merging training data in warm-up and pretrain phases of MindVL-8B-4Phase.
By changing the four-phase training recipe to a three-phase training recipe, we have not only improved training efficiency but also enhanced the model performances.
Such results indicate that training a MLP adaptor can also align ViT and LLMs effectively.

We also conducted experiments with  two-phase training recipes, MindVL-8B-2Phase* undergoes warm-up and SFT, and MindVL-8B-2Phase undergoes warm-up and multi-task learning with the training data in corresponding phase.
The results show that MindVL-8B-2Phase outperformance MindVL-8B-2Phase*, which indicate that high-quality multi-task data is crucial for enhancing the capabilities of multimodal models.

\section{Examples}
We list some samples whose answers are predicted by our MindVL-671B-37B.



\newpage
\begin{figure}[!t] 
    \centering
        \includegraphics[width=0.99\textwidth]{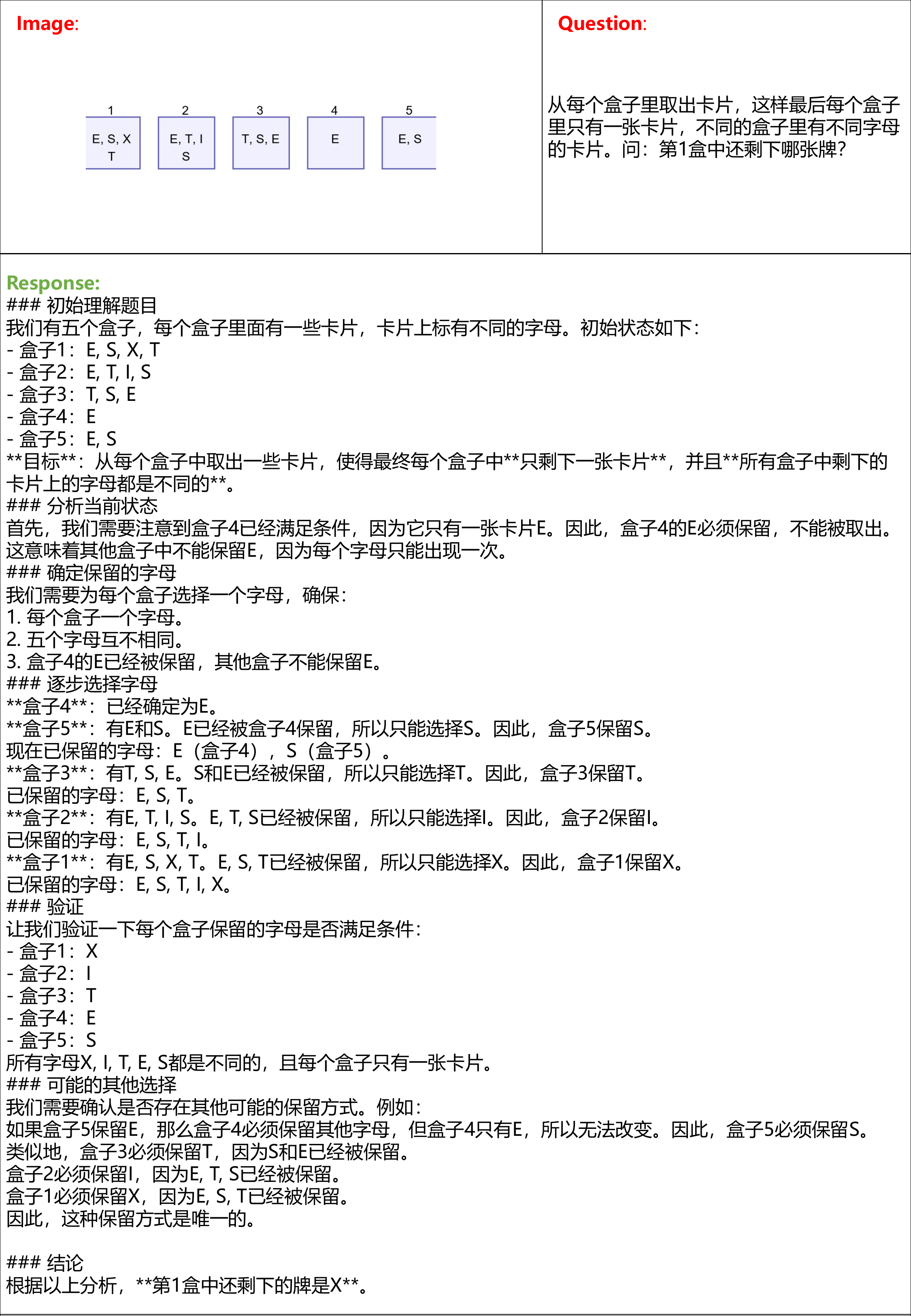}

\end{figure}

\newpage
\begin{figure}[!t] 
    \centering
        \includegraphics[width=0.99\textwidth]{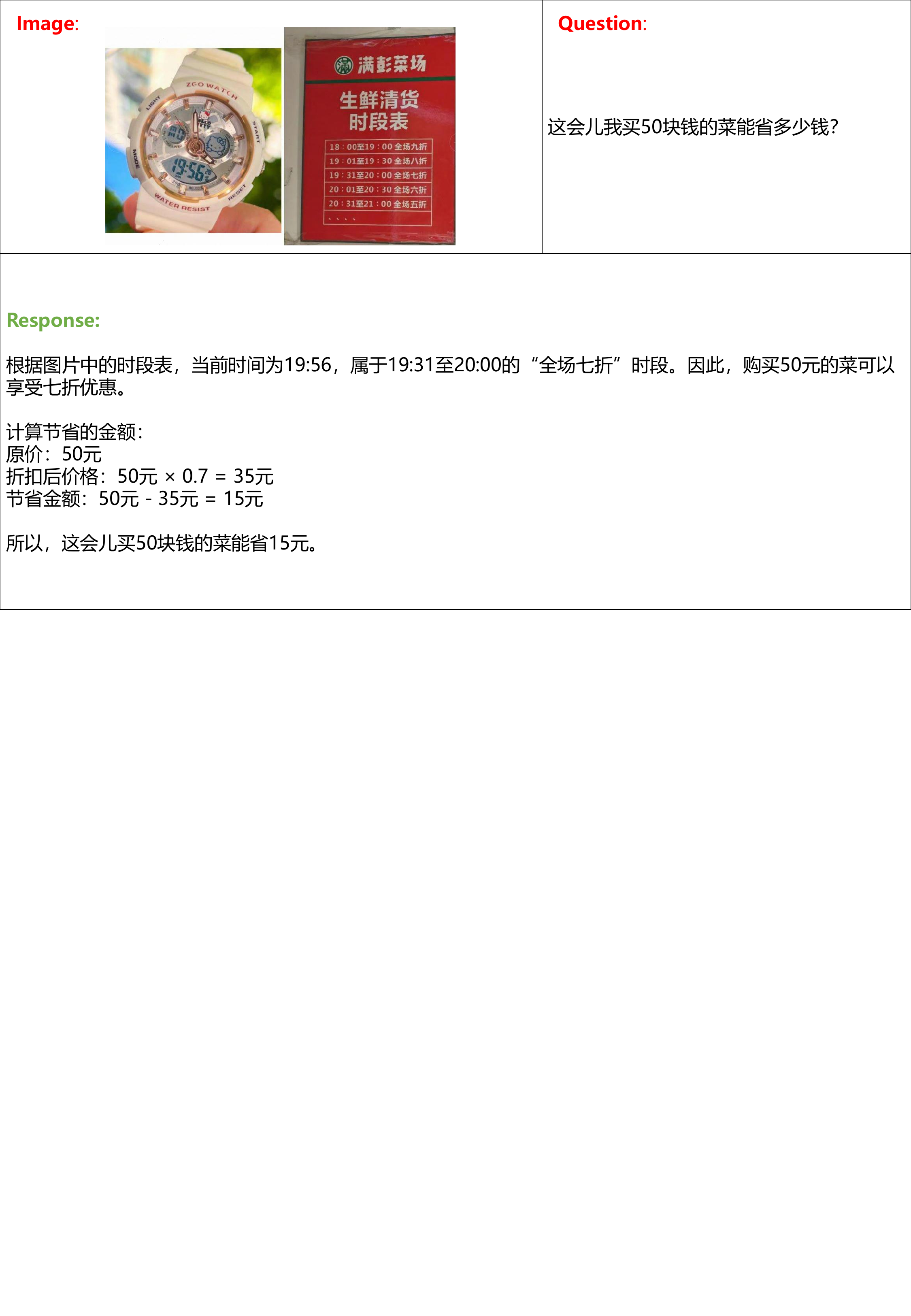}

\end{figure}
\newpage
\begin{figure}[!t] 
    \centering
        \includegraphics[width=0.99\textwidth]{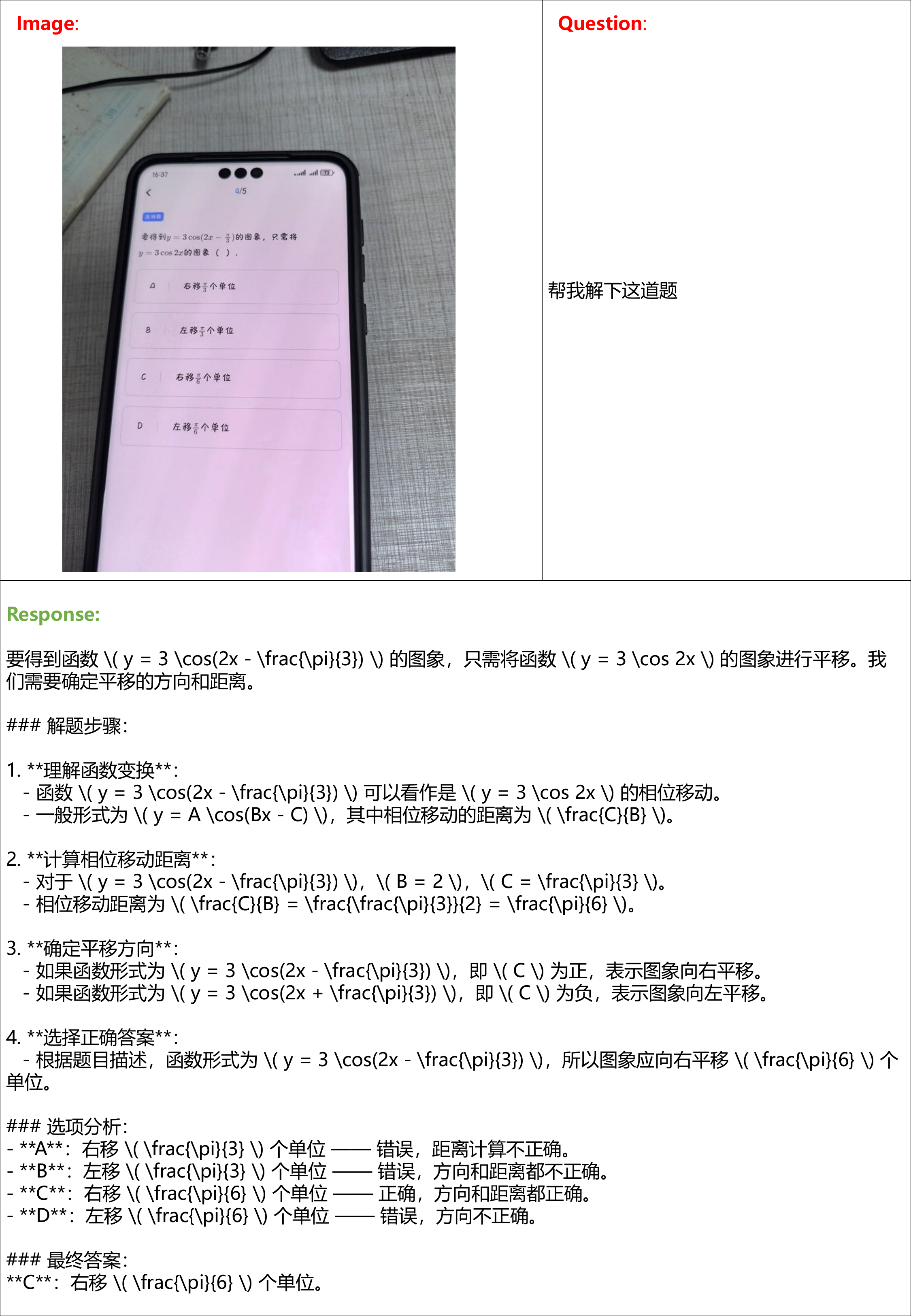}

\end{figure}
\newpage
\begin{figure}[!t] 
    \centering
        \includegraphics[width=0.99\textwidth]{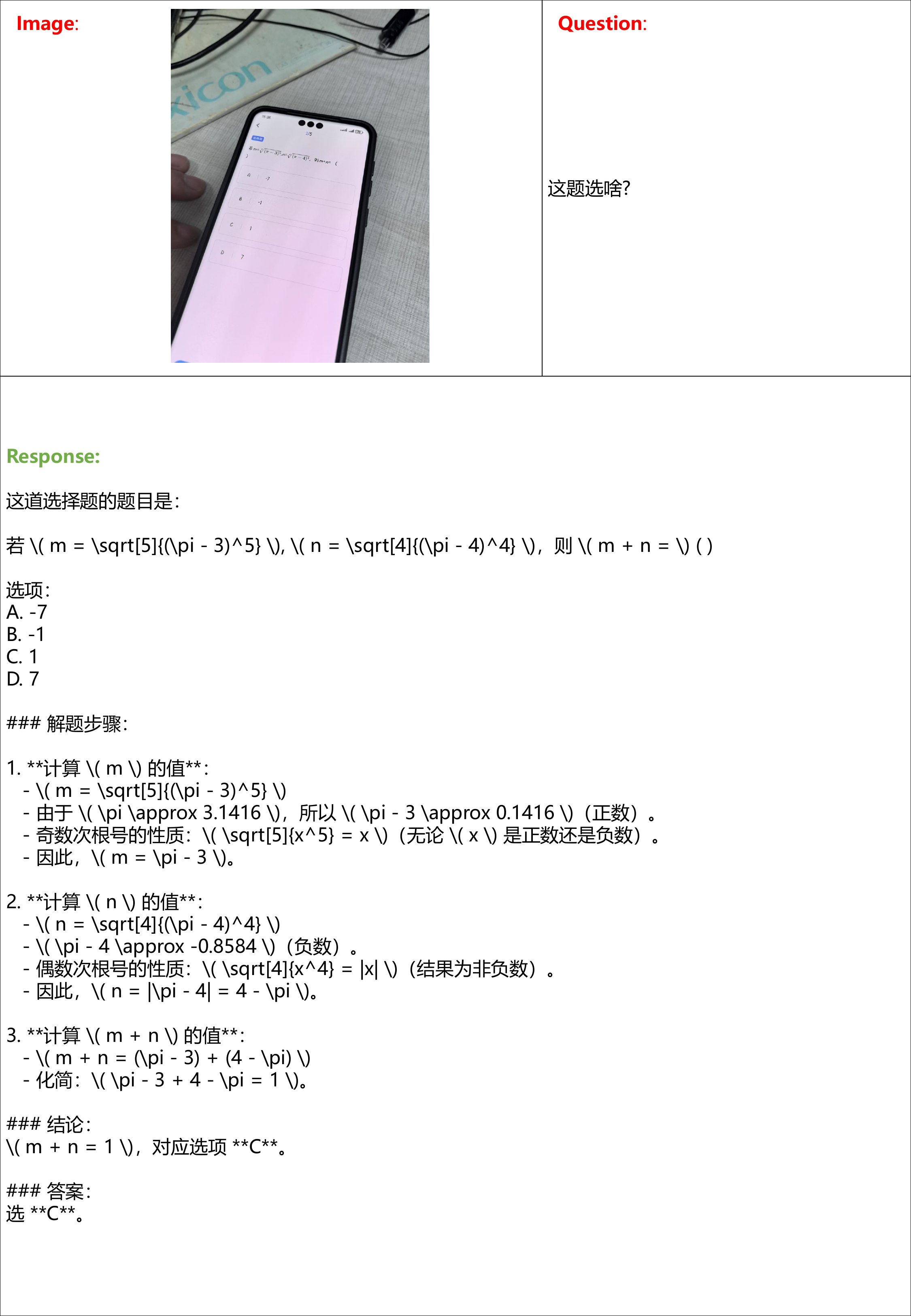}

\end{figure}
\newpage
\begin{figure}[!t] 
    \centering
        \includegraphics[width=0.99\textwidth]{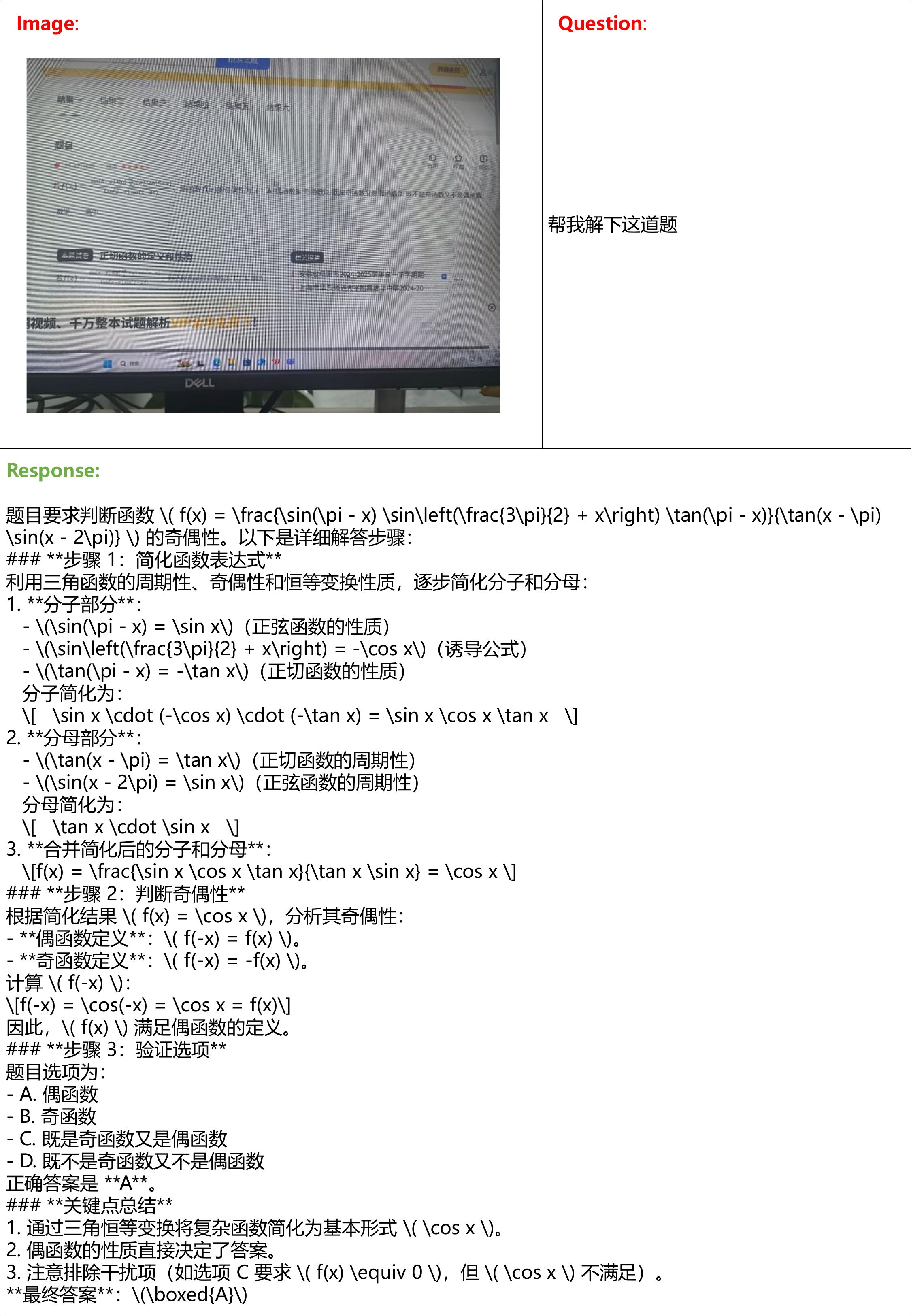}

\end{figure}

\end{document}